\documentclass[11pt,a4paper]{article}
\usepackage[hyperref]{naaclhlt2018}
\usepackage{times}
\usepackage{multirow}
\usepackage{latexsym}
\usepackage{graphicx} 
\usepackage{amssymb}
\usepackage{amsmath}
\usepackage{hyperref}
\usepackage{subfigure}
\usepackage{fancyhdr}

\usepackage{url}

\aclfinalcopy 

\title{DR-BiLSTM: Dependent Reading Bidirectional LSTM for \\ Natural Language Inference}

\author{
	Reza Ghaeini$^1$\thanks{\ \ This work was conducted as part of an internship program at Philips Research.} , Sadid A. Hasan$^2$, Vivek Datla$^2$, Joey Liu$^2$, Kathy Lee$^2$, Ashequl Qadir$^2$, \\ 
	\textbf{Yuan Ling$^2$, Aaditya Prakash$^2$, Xiaoli Z. Fern$^1$ and Oladimeji Farri$^2$} \\
	$^1$Oregon State University, Corvallis, OR, USA \\
	$^2$Artificial Intelligence Laboratory, Philips Research North America, Cambridge, MA, USA \\
	\{ghaeinim,xfern\}@eecs.oregonstate.edu \\
	\{sadid.hasan,vivek.datla,joey.liu,kathy.lee\_1,ashequl.qadir\}@philips.com \\
	\{yuan.ling,aaditya.prakash,dimeji.farri\}@philips.com
}

\date{}

\begin{document}
	\maketitle
	\begin{abstract}		
		We present a novel deep learning architecture to address the natural language inference (NLI) task. Existing approaches mostly rely on simple reading mechanisms for independent encoding of the premise and hypothesis. Instead, we propose a novel \emph{dependent reading} bidirectional LSTM network (DR-BiLSTM) to efficiently model the relationship between a premise and a hypothesis during encoding and inference. We also introduce a sophisticated ensemble strategy to combine our proposed models, which noticeably improves final predictions. Finally, we demonstrate how the results can be improved further with an additional preprocessing step. Our evaluation shows that DR-BiLSTM obtains the best single model and ensemble model results achieving the new state-of-the-art scores on the Stanford NLI dataset.
		
	\end{abstract}
	
	\section{Introduction}
	Natural Language Inference (NLI; a.k.a. Recognizing Textual Entailment, or RTE) is an important and challenging task for natural language understanding \cite{nli}. The goal of NLI is to identify the logical relationship (\emph{entailment}, \emph{neutral}, or \emph{contradiction}) between a premise and a corresponding hypothesis. Table~\ref{tab:sample} shows few example relationships from the Stanford Natural Language Inference (SNLI) dataset \cite{snli}.

	\begin{table}[t]
		\small
		\begin{center}
			\begin{tabular}{|c|c|c|}
				\hline 
				\multirow{3}{*}{\textbf{P$^a$}} & A senior is waiting at the & \multirow{3}{*}{Relationship} \\
				& window of a restaurant &  \\
				& that serves sandwiches. &  \\ \hline
				\multirow{6}{*}{\textbf{H$^b$}} & A person waits to be   & \multirow{2}{*}{Entailment} \\
				& served his food. &  \\ \cline{2-3}
				& A man is looking to order & \multirow{2}{*}{Neutral} \\ 
				& a grilled cheese sandwich. & \\ \cline{2-3}
				& A man is waiting in line & \multirow{2}{*}{Contradiction} \\
				& for the bus. &  \\ \hline
				\multicolumn{3}{|l|}{$^a$\textbf{P}, Premise.}\\
				\multicolumn{3}{|l|}{$^b$\textbf{H}, Hypothesis.}\\ \hline
			\end{tabular}
		\end{center}
		\caption{\label{tab:sample} Examples from the SNLI dataset.}
	\end{table}

	Recently, NLI has received a lot of attention from the researchers, especially due to the availability of large annotated datasets like SNLI \cite{snli}. Various deep learning models have been proposed that achieve successful results for this task \cite{gong2017,ibm2017,him2017,nse2017,google2016, kai_2016,re-read}. Most of these existing NLI models use attention mechanism to jointly interpret and align the premise and hypothesis. Such models use simple reading mechanisms to encode the premise and hypothesis independently. However, such a complex task require explicit modeling of dependency relationships between the premise and the hypothesis during the encoding and inference processes to prevent the network from the loss of relevant, contextual information. In this paper, we refer to such strategies as \emph{dependent reading}. 
	
	There are some alternative reading mechanisms available in the literature \cite{re-read,Rocktaschel2015} that consider dependency aspects of the premise-hypothesis relationships. However, these mechanisms have two major limitations:
	
	\begin{itemize}
		\item So far, they have only explored dependency aspects during the encoding stage, while ignoring its benefit during inference.
		\item Such models only consider encoding a hypothesis depending on the premise, disregarding the dependency aspects in the opposite direction.
	\end{itemize}
	
	We propose a dependent reading bidirectional LSTM (DR-BiLSTM) model to address these limitations. Given a premise $u$ and a hypothesis $v$, our model first encodes them considering dependency on each other ($u|v$ and $v|u$). Next, the model employs a soft attention mechanism to extract relevant information from these encodings. The augmented sentence representations are then passed to the inference stage, which uses a similar dependent reading strategy in both directions, i.e. $u \rightarrow v$ and $v \rightarrow u$. Finally, a decision is made through a multi-layer perceptron (MLP) based on the aggregated information. 
	
	Our experiments on the SNLI dataset show that DR-BiLSTM achieves the best single model and ensemble model performance obtaining improvements of a considerable margin of $0.4\%$ and $0.3\%$ over the previous state-of-the-art single and ensemble models, respectively. 
	
	Furthermore, we demonstrate the importance of a simple preprocessing step performed on the SNLI dataset. Evaluation results show that such preprocessing allows our single model to achieve the same accuracy as the state-of-the-art ensemble model and improves our ensemble model to outperform the state-of-the-art ensemble model by a remarkable margin of $0.7\%$. Finally, we perform an extensive analysis to clarify the strengths and weaknesses of our models. 
	
	\section{Related Work}
	Early studies use small datasets while leveraging lexical and syntactic features for NLI \cite{nli}. The recent availability of large-scale annotated datasets \cite{snli,multinli} has enabled researchers to develop various deep learning-based architectures for NLI. 
	
	\citet{google2016} propose an attention-based model \cite{nmt} that decomposes the NLI task into sub-problems to solve them in parallel. They further show the benefit of adding intra-sentence attention to input representations. \citet{him2017} explore sequential inference models based on chain LSTMs with attentional input encoding and demonstrate the effectiveness of syntactic information. We also use similar attention mechanisms. However, our model is distinct from these models as they do not benefit from dependent reading strategies.
	
	\citet{Rocktaschel2015} use a word-by-word neural attention mechanism while \citet{re-read} propose re-read LSTM units by considering the dependency of a hypothesis on the information of its premise ($v|u$) to achieve promising results. However, these models suffer from weak inferencing methods by disregarding the dependency aspects from the opposite direction ($u|v$). Intuitively, when a human judges a premise-hypothesis relationship, s/he might consider back-and-forth reading of both sentences before coming to a conclusion. Therefore, it is essential to encode the premise-hypothesis dependency relations from both directions to optimize the understanding of their relationship.    
	
	\citet{ibm2017} propose a bilateral multi-perspective matching (BiMPM) model, which resembles the concept of matching a premise and hypothesis from both directions. Their matching strategy is essentially similar to our attention mechanism that utilizes relevant information from the other sentence for each word sequence. They use similar methods as \citet{him2017} for encoding and inference, without any dependent reading mechanism. 
	
	Although NLI is well studied in the literature, the potential of dependent reading and interaction between a premise and hypothesis is not rigorously explored. In this paper, we address this gap by proposing a novel deep learning model (DR-BiLSTM). Experimental results demonstrate the effectiveness of our model.
	
	\begin{figure}[ht]
		\centering
		\includegraphics[width=0.48\textwidth ]{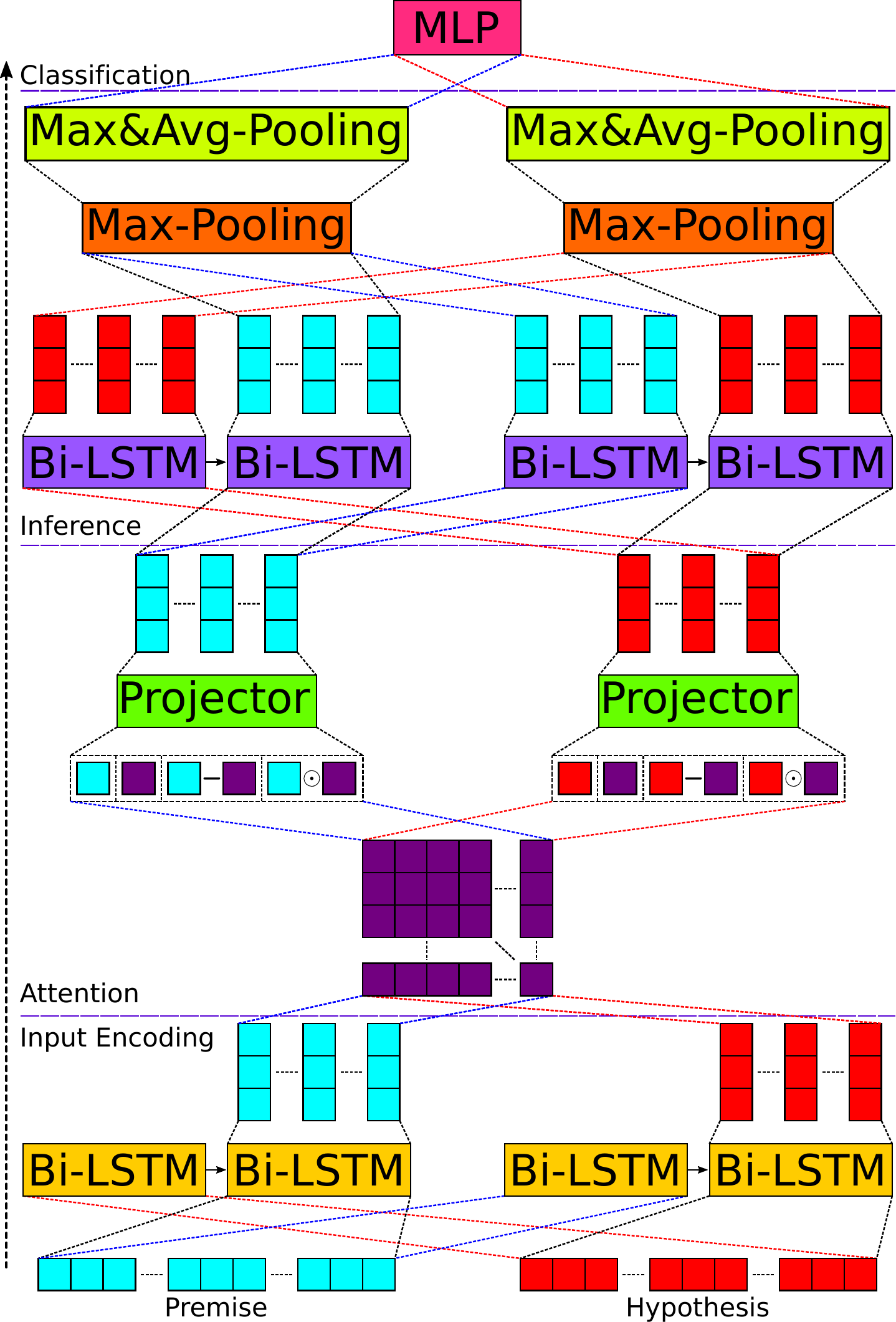}
		\caption{A high-level view of DR-BiLSTM. The data (premise $u$ and hypothesis $v$, depicted with cyan and red tensors respectively) flows from bottom to top. Relevant tensors are shown with the same color and elements with the same colors share parameters. \label{fig:model}}
	\end{figure}
	
	\section{Model}
	Our proposed model (DR-BiLSTM) is composed of the following major components: input encoding, attention, inference, and classification. Figure~\ref{fig:model} demonstrates a high-level view of our proposed NLI framework. 
	
	Let $u=[u_1, \cdots, u_n]$ and $v=[v_1, \cdots, v_m]$ be the given premise with length $n$ and hypothesis with length $m$ respectively, where $u_i, v_j \in \mathbb{R}^r$ is an word embedding of $r$-dimensional vector. The task is to predict a label $y$ that indicates the logical relationship between premise $u$ and hypothesis $v$.
	
	\subsection{Input Encoding}
	\label{sec:enc}
	
	RNNs are the natural solution for variable length sequence modeling, consequently, we utilize a bidirectional LSTM (BiLSTM) \cite{lstm} for encoding the given sentences. For ease of presentation, we only describe how we encode $u$ depending on $v$. The same procedure is utilized for the reverse direction ($v|u$). 
	
	To dependently encode $u$, we first process $v$ using the BiLSTM. Then we read $u$ through the BiLSTM that is initialized with previous reading final states (memory cell and hidden state). Here we represent a word (e.g. $u_i$) and its context depending on the other sentence (e.g. $v$).  Equations \ref{eq:enc:pout} and \ref{eq:enc:hout} formally represent this component.
	
	\begin{equation}
	\begin{split}
	\bar{v}, s_v &= \textit{BiLSTM}(v, 0) \\
	\hat{u}, - &= \textit{BiLSTM}(u, s_v)
	\end{split}
	\label{eq:enc:pout}
	\end{equation}
	
	\begin{equation}
	\begin{split}
	\bar{u}, s_u &= \textit{BiLSTM}(u, 0) \\
	\hat{v}, - &= \textit{BiLSTM}(v, s_u)
	\end{split}
	\label{eq:enc:hout}
	\end{equation}
	
	\noindent where $\{\bar{u} \in \mathbb{R}^{n \times 2d}, \hat{u} \in \mathbb{R}^{n \times 2d}, s_u \}$ and $\{\bar{v} \in \mathbb{R}^{m \times 2d}, \hat{v} \in \mathbb{R}^{m \times 2d}, s_v \}$ are the independent reading sequences, dependent reading sequences, and BiLSTM final state of independent reading of $u$ and $v$ respectively. Note that, ``$-$'' in these equations means that we do not care about the associated variable and its value. BiLSTM inputs are the word embedding sequences and initial state vectors. $\hat{u}$ and $\hat{v}$ are passed to the next layer as the output of the input encoding component. 
	
	The proposed encoding mechanism yields a richer representation for both premise and hypothesis by taking the history of each other into account. Using a max or average pooling over the independent and dependent readings does not further improve our model. This was expected since dependent reading produces more promising and relevant encodings.
	
	\subsection{Attention}
	\label{sec:att}
	
	We employ a soft alignment method to associate the relevant sub-components between the given premise and hypothesis. In deep learning models, such purpose is often achieved with a soft attention mechanism. Here we compute the unnormalized attention weights as the similarity of hidden states of the premise and hypothesis with Equation~\ref{eq:energy} (energy function).

	\begin{equation}
	e_{ij} = \hat{u}_i \hat{v}_j^T,  \quad  i \in [1,n], j \in [1,m]
	\label{eq:energy}
	\end{equation}
	
	\noindent where $\hat{u}_i$ and $\hat{v}_j$ are the dependent reading hidden representations of $u$ and $v$ respectively which are computed earlier in Equations \ref{eq:enc:pout} and \ref{eq:enc:hout}. Next, for each word in either premise or hypothesis, the relevant semantics in the other sentence is extracted and composed according to $e_{ij}$. Equations \ref{eq:att:p} and \ref{eq:att:h} provide formal and specific details of this procedure.
	
	\begin{equation}
	\tilde{u}_i = \sum_{j=1}^{m} \frac{\exp(e_{ij})}{\sum_{k=1}^{m} \exp(e_{ik})} \hat{v}_j, \quad i \in [1,n]
	\label{eq:att:p}
	\end{equation}
	\begin{equation}
	\tilde{v}_j = \sum_{i=1}^{n} \frac{\exp(e_{ij})}{\sum_{k=1}^{n} \exp(e_{kj})} \hat{u}_i, \quad j \in [1,m]
	\label{eq:att:h}
	\end{equation}
	
	\noindent where $\tilde{u}_i$ represents the extracted relevant information of $\hat{v}$ by attending to $\hat{u}_i$ while $\tilde{v}_j$ represents the extracted relevant information of $\hat{u}$ by attending to $\hat{v}_j$. 
	
	To further enrich the collected attentional information, a trivial next step would be to pass the concatenation of the tuples $(\hat{u}_i, \tilde{u}_i)$ or $(\hat{v}_j, \tilde{v}_j)$ which provides a linear relationship between them. However, the model would suffer from the absence of \emph{similarity} and \emph{closeness} measures. Therefore, we calculate the difference and element-wise product for the tuples $(\hat{u}_i, \tilde{u}_i)$ and $(\hat{v}_j, \tilde{v}_j)$ that represent the similarity and closeness information respectively \cite{him2017,ama}.
	
	The difference and element-wise product are then concatenated with the computed vectors, $(\hat{u}_i, \tilde{u}_i)$ or $(\hat{v}_j, \tilde{v}_j)$, respectively. Finally, a feedforward neural layer with ReLU activation function projects the concatenated vectors from $8d$-dimensional vector space into a $d$-dimensional vector space (Equations \ref{eq:prj:p} and \ref{eq:prj:h}). This helps the model to capture deeper dependencies between the sentences besides lowering the complexity of vector representations.
	
	\begin{equation}
	\begin{split}
	a_i &= [\hat{u}_i, \tilde{u}_i, \hat{u}_i -\tilde{u}_i, \hat{u}_i \odot \tilde{u}_i] \\
	p_i &=  \textit{ReLU}(W_p a_i + b_p)
	\end{split}
	\label{eq:prj:p}
	\end{equation}
	\begin{equation}
	\begin{split}
	b_j &= [\hat{v}_j, \tilde{v}_j, \hat{v}_j -\tilde{v}_j, \hat{v}_j \odot \tilde{v}_j] \\
	q_j &=  \textit{ReLU}(W_p b_j + b_p)
	\end{split}
	\label{eq:prj:h}
	\end{equation}
	
	\noindent Here $\odot$ stands for element-wise product while $W_p \in \mathbb{R}^{8d\times d}$ and $b_p \in \mathbb{R}^{d}$ are the trainable weights and biases of the projector layer respectively.
	
	\subsection{Inference}
	During this phase, we use another BiLSTM to aggregate the two sequences of computed matching vectors, $p$ and $q$ from the attention stage (Section~\ref{sec:att}). This aggregation is performed in a sequential manner to avoid losing effect of latent variables that might rely on the sequence of matching vectors. 
	
	Instead of aggregating the sequences of matching vectors individually, we propose a similar dependent reading approach for the inference stage. We employ a BiLSTM reading process (Equations~\ref{eq:inf:pout} and \ref{eq:inf:hout}) similar to the input encoding step discussed in Section~\ref{sec:enc}. But rather than passing just the dependent reading information to the next step, we feed both independent reading ($\bar{p}$ and $\bar{q}$) and dependent reading ($\hat{p}$ and $\hat{q}$) to a max pooling layer, which selects maximum values from each sequence of independent and dependent readings ($\bar{p}_i$ and $\hat{p}_i$) as shown in Equations~\ref{eq:mpool:p} and \ref{eq:mpool:h}. The main intuition behind this architecture is to maximize the inferencing ability of the model by considering both independent and dependent readings.

	\begin{equation}
	\begin{split}
	\bar{q}, s_q &= \textit{BiLSTM}(q, 0) \\
	\hat{p}, - &= \textit{BiLSTM}(p, s_q)
	\end{split}
	\label{eq:inf:pout}
	\end{equation}
	\begin{equation}
	\begin{split}
	\bar{p}, s_p &= \textit{BiLSTM}(p, 0) \\
	\hat{q}, - &= \textit{BiLSTM}(q, s_p)
	\end{split}
	\label{eq:inf:hout}
	\end{equation}
	\begin{equation}
	\tilde{p} = \textit{MaxPooling}(\bar{p}, \hat{p})
	\label{eq:mpool:p}
	\end{equation}
	\begin{equation}
	\tilde{q} = \textit{MaxPooling}(\bar{q}, \hat{q})
	\label{eq:mpool:h}
	\end{equation}

	\noindent Here $\{\bar{p} \in \mathbb{R}^{n \times 2d}, \hat{p} \in \mathbb{R}^{n \times 2d}, s_p\}$ and $\{\bar{q} \in \mathbb{R}^{m \times 2d}, \hat{q} \in \mathbb{R}^{m \times 2d}, s_q\}$ are the independent reading sequences, dependent reading sequences, and BiLSTM final state of independent reading of $p$ and $q$ respectively. BiLSTM inputs are the word embedding sequences and initial state vectors. 
	
	Finally, we convert $\tilde{p} \in \mathbb{R}^{n\times 2d}$ and $\tilde{q} \in \mathbb{R}^{m \times 2d}$ to fixed-length vectors with pooling, $U \in \mathbb{R}^{4d}$ and $V \in \mathbb{R}^{4d}$. As shown in Equations~\ref{eq:fixlpool:p} and \ref{eq:fixlpool:h}, we employ both max and average pooling and describe the overall inference relationship with concatenation of their outputs.
	
	\begin{equation}
	U = [\textit{MaxPooling}(\tilde{p}), \textit{AvgPooling}(\tilde{p})]
	\label{eq:fixlpool:p}
	\end{equation}
	\begin{equation}
	V = [\textit{MaxPooling}(\tilde{q}), \textit{AvgPooling}(\tilde{q})]
	\label{eq:fixlpool:h}
	\end{equation}
	
	\subsection{Classification}
	Here, we feed the concatenation of $U$ and $V$ ($[U,V]$) into a multilayer perceptron (MLP) classifier that includes a hidden layer with $\emph{tanh}$ activation and $\emph{softmax}$ output layer. The model is trained in an end-to-end manner.
	
	\begin{equation}
	\textit{Output} = \textit{MLP}([U,V])
	\label{eq:mlp}
	\end{equation}

\section{Experiments and Evaluation}
	
	\subsection{Dataset} \label{sec:data:snli}
	The Stanford Natural Language Inference (SNLI) dataset contains $570K$ human annotated sentence pairs. The premises are drawn from the Flickr30k~\cite{flickr} corpus, and then the hypotheses are manually composed for each relationship class (\emph{entailment}, \emph{neutral}, \emph{contradiction}, and \emph{-}). The ``\emph{-}'' class indicates that there is no consensus decision among the annotators, consequently, we remove them during the training and evaluation following the literature. We use the same data split as provided in \citet{snli} to report comparable results with other models.	
	
	\subsection{Experimental Setup} \label{sec:exp_s}
	We use pre-trained $300$-$D$ Glove $840B$ vectors ~\cite{glove} to initialize our word embedding vectors. All hidden states of BiLSTMs during input encoding and inference have $450$ dimensions ($r=300$ and $d=450$). The weights are learned by minimizing the log-loss on the training data via the Adam optimizer~\cite{adam}. The initial learning rate is 0.0004. To avoid overfitting, we use dropout~\cite{dropout} with the rate of 0.4 for regularization, which is applied to all feedforward connections. During training, the word embeddings are updated to learn effective representations for the NLI task. We use a fairly small batch size of 32 to provide more exploration power to the model. Our observation indicates that using larger batch sizes hurts the performance of our model. 
	
	\begin{figure}[ht]
		\centering
		\includegraphics[width=0.49\textwidth ]{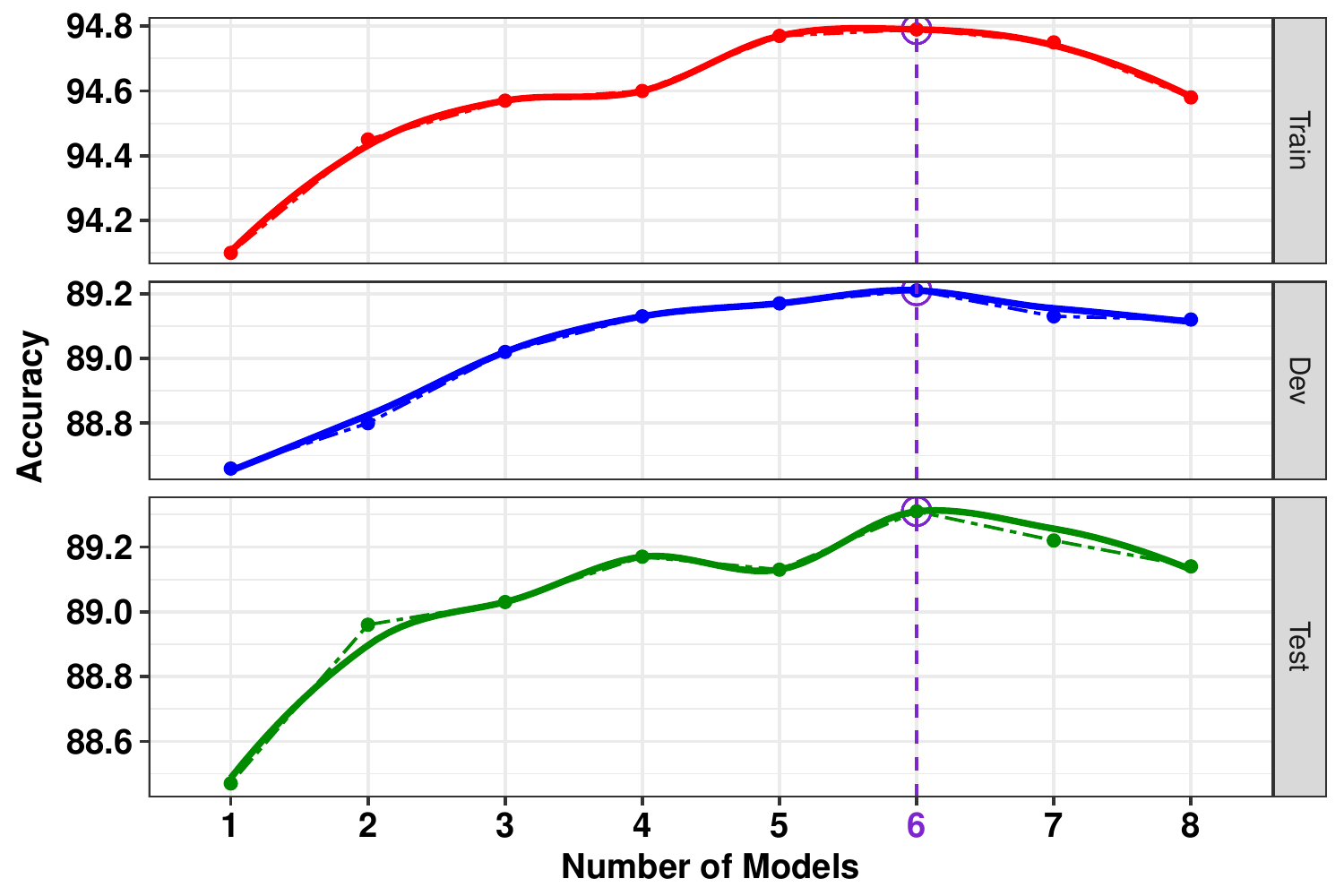}
		\caption{Performance of $n$ ensemble models reported for training (red, top), development (blue, middle), and test (green, bottom) sets of SNLI. For $n$ number of models, the best performance on the development set is used as the criteria to determine the final ensemble. The best performance on development set ($89.22\%$) is observed using 6 models and is henceforth considered as our final DR-BiLSTM (Ensemble) model. \label{fig:ensemble}}
	\end{figure}
	
	\subsection{Ensemble Strategy} \label{sec:es}
	Ensemble methods use multiple models to obtain better predictive performance. Previous works typically utilize trivial ensemble strategies by either using majority votes or averaging the probability distributions over the same model with different initialization seeds \cite{ibm2017,gong2017}. 
	
	By contrast, we use weighted averaging of the probability distributions where the weight of each model is learned through its performance on the SNLI development set. Furthermore, the differences between our models in the ensemble originate from: 1) variations in the number of dependent readings (i.e. 1 and 3 rounds of dependent reading), 2) projection layer activation (\emph{tanh} and \emph{ReLU} in Equations~\ref{eq:prj:p} and \ref{eq:prj:h}), and 3) different initialization seeds. 
	
	The main intuition behind this design is that the effectiveness of a model may depend on the complexity of a premise-hypothesis instance. For a simple instance, a simple model could perform better than a complex one, while a complex instance may need further consideration toward disambiguation. Consequently, using models with different rounds of dependent readings in the encoding stage should be beneficial. 
	
	Figure~\ref{fig:ensemble} demonstrates the observed performance of our ensemble method with different number of models. The performance of the models are reported based on the best obtained accuracy on the development set. We also study the effectiveness of other ensemble strategies e.g. majority voting, and averaging the probability distributions. But, our ensemble strategy performs the best among them (see Section~\ref{app:ensemble:sec} in the Appendix for additional details).
	
	\subsection{Preprocessing}
	We perform a trivial preprocessing step on SNLI to recover some out-of-vocabulary words found in the development set and test set. Note that our vocabulary contains all words that are seen in the training set, so there is no out-of-vocabulary word in it. The SNLI dataset is not immune to human errors, specifically, misspelled words. We noticed that misspelling is the main reason for some of the observed out-of-vocabulary words. Consequently, we simply fix the unseen misspelled words using Microsoft spell-checker (other approaches like edit distance can also be used). Moreover, while dealing with an unseen word during evaluation, we try to: 1) replace it with its lower case, or 2) split the word when it contains a ``-'' (e.g. ``marsh-like'') or starts with ``un'' (e.g. ``unloading''). If we still could not find the word in our vocabulary, we consider it as an \emph{unknown} word. In the next subsection, we demonstrate the importance and impact of such trivial preprocessing (see Section~\ref{app:preproc:sec} in the Appendix for additional details).
	
	\subsection{Results}	
	
	\noindent Table~\ref{tab:snli:result} shows the accuracy of the models on training and test sets of SNLI. The first row represents a baseline classifier presented by \citet{snli} that utilizes handcrafted features. All other listed models are deep-learning based. The gap between the traditional model and deep learning models demonstrates the effectiveness of deep learning methods for this task. We also report the estimated human performance on the SNLI dataset, which is the average accuracy of five annotators in comparison to the gold labels \cite{gong2017}. It is noteworthy that recent deep learning models surpass the human performance in the NLI task.
	
	\begin{table}[ht]
		\begin{center}
			\small
			\begin{tabular}{lcc}
				\hline
				\multirow{2}{*}{\textbf{Model}}& \multicolumn{2}{c}{\bf{Accuracy}} \\ \cline{2-3}
				& \textbf{Train}&\textbf{Test}\\ 
				\hline \hline
				\cite{snli} (Feature)&99.7\%&78.2\%\\ 
				\hline \hline
				\cite{snli}&83.9\%&80.6\%\\ 
				\cite{Vendrov2015}&98.8\%&81.4\%\\
				\cite{Mou2016}&83.3\%&82.1\%\\
				\cite{Bowman2016}&89.2\%&83.2\%\\
				\cite{Liu2016}&84.5\%&84.2\%\\
				\cite{nse2017}&86.2\%&84.6\%\\ 
				\hline\hline
				\cite{Rocktaschel2015}&85.3\%&83.5\%\\
				\cite{Wang2016}&92.0\%&86.1\%\\
				\cite{fusion2016}&88.5\%&86.3\%\\
				\cite{google2016}&90.5\%&86.8\%\\
				\cite{nti2017}&88.5\%&87.3\%\\
				\cite{re-read}&90.7\%&87.5\%\\
				\cite{ibm2017} (Single)&90.9\%&87.5\%\\
				\cite{him2017} (Single)&92.6\%&88.0\% \\
				\cite{gong2017} (Single)&91.2\%&88.0\% \\
				\hline \hline
				\cite{him2017} (Ensemble)&93.5\%&88.6\%\\
				\cite{ibm2017} (Ensemble)&93.2\%&88.8\%\\
				\cite{gong2017} (Ensemble)&92.3\%&88.9\%\\
				\hline \hline
				Human Performance (Estimated)&97.2\%&87.7\%\\
				\hline\hline
				DR-BiLSTM (Single)&94.1\%&\textbf{88.5\%} \\
				DR-BiLSTM (Single)$+$Process&94.1\%&\textbf{88.9\%} \\
				\hline
				DR-BiLSTM (Ensemble)&94.8\%&\textbf{89.3\%} \\
				DR-BiLSTM (Ensem.)$+$Process&94.8\%&\textbf{89.6\%} \\
				\hline
			\end{tabular}
		\end{center}
		\caption{\label{tab:snli:result} Accuracies of the models on the training set and test set of SNLI. DR-BiLSTM (Ensemble) achieves the accuracy of $89.3\%$, the best result observed on SNLI, while DR-BiLSTM (Single) obtains the accuracy of $88.5\%$, which considerably outperforms the previous non-ensemble models. Also, utilizing a trivial preprocessing step yields to further improvements of $0.4\%$ and $0.3\%$ for single and ensemble DR-BiLSTM models respectively.}
	\end{table}
	
	As shown in Table~\ref{tab:snli:result}, previous deep learning models (rows 2-19) can be divided into three categories: 1) sentence encoding based models (rows 2-7), 2) single inter-sentence attention-based models (rows 8-16), and 3) ensemble inter-sentence attention-based models (rows 17-19). We can see that inter-sentence attention-based models perform better than sentence encoding based models, which supports our intuition. Natural language inference requires a deep interaction between the premise and hypothesis. Inter-sentence attention-based approaches can provide such interaction while sentence encoding based models fail to do so. 

	To further enhance the modeling of interaction between the premise and hypothesis for efficient disambiguation of their relationship, we introduce the dependent reading strategy in our proposed DR-BiLSTM model. The results demonstrate the effectiveness of our model. DR-BiLSTM (Single) achieves $88.5\%$ accuracy on the test set which is noticeably the best reported result among the existing single models for this task. Note that the difference between DR-BiLSTM and \citet{him2017} is statistically significant with a p-value of $<0.001$ over the \emph{Chi-square} test\footnote{Chi-square test ($\chi^2$ test) is used to determine if there is a significant difference between two categorical variables (i.e. models' outputs).}.
	
	To further improve the performance of NLI systems, researchers have built ensemble models. Previously, ensemble systems obtained the best performance on SNLI with a huge margin. Table~\ref{tab:snli:result} shows that our proposed single model achieves competitive results compared to these reported ensemble models. Our ensemble model considerably outperforms the current state-of-the-art by obtaining $89.3\%$ accuracy.
	
	Up until this point, we discussed the performance of our models where we have not considered preprocessing for recovering the out-of-vocabulary words. In Table~\ref{tab:snli:result}, ``DR-BiLSTM (Single) $+$ Process'', and ``DR-BiLSTM (Ensem.) $+$ Process'' represent the performance of our models on the preprocessed dataset. We can see that our preprocessing mechanism leads to further improvements of $0.4\%$ and $0.3\%$ on the SNLI test set for our single and ensemble models respectively. In fact, our single model (``DR-BiLSTM (Single) $+$ Process'') obtains the state-of-the-art performance over both reported single and ensemble models by performing a simple preprocessing step. Furthermore, ``DR-BiLSTM (Ensem.) $+$ Process'' outperforms the existing state-of-the-art remarkably ($0.7\%$ improvement). For more comparison and analyses, we use ``DR-BiLSTM (Single)'' and ``DR-BiLSTM (Ensemble)'' as our single and ensemble models in the rest of the paper.
	
	\subsection{Ablation and Configuration Study}
	We conducted an ablation study on our model to examine the importance and effect of each major component. Then, we study the impact of BiLSTM dimensionality on the performance of the development set and training set of SNLI. We investigate all settings on the development set of the SNLI dataset. 
	
	\begin{table}[ht]
		\begin{center}
			\small
			\resizebox{1\linewidth}{!}{
				\begin{tabular}{lcr}
					\hline
					\bf{Model} & \bf{Dev Acc$^a$} & \bf{p-value} \\
					\hline \hline
					DR-BiLSTM &\textbf{88.69\%} & -\\ \hline
					DR-BiLSTM - hidden MLP &88.45\% & $<$0.001\\ \hline
					DR-BiLSTM - average pooling &88.50\% & $<$0.001\\ 
					DR-BiLSTM - max pooling &88.39\% & $<$0.001\\ \hline
					DR-BiLSTM - elem. prd$^b$ &88.51\% & $<$0.001\\
					DR-BiLSTM - difference &88.24\% & $<$0.001\\
					DR-BiLSTM - diff$^c$ \& elem. prd &87.96\% & $<$0.001\\ \hline
					DR-BiLSTM - inference pooling &88.46\%& $<$0.001\\
					DR-BiLSTM - dep. infer$^d$ &88.43\%& $<$0.001\\
					DR-BiLSTM - dep. enc$^e$ &88.26\%& $<$0.001\\
					DR-BiLSTM - dep. enc \& infer &88.20\%& $<$0.001\\
					\hline \hline
					\multicolumn{3}{l}{$^a$Dev Acc, Development Accuracy.}\\
					\multicolumn{3}{l}{$^b$elem. prd, element-wise product.}\\
					\multicolumn{3}{l}{$^c$diff, difference.}\\
					\multicolumn{3}{l}{$^d$dep. infer, dependent reading inference.}\\
					\multicolumn{3}{l}{$^e$dep. enc, dependent reading encoding.}\\\hline
				\end{tabular}
			}
		\end{center}
		\caption{\label{tab:ablation} Ablation study results. Performance of different configurations of the proposed model on the development set of SNLI along with their p-values in comparison to DR-BiLSTM (Single).}
	\end{table}
	
	Table~\ref{tab:ablation} shows the ablation study results on the development set of SNLI along with the statistical significance test results in comparison to the proposed model, DR-BiLSTM. We can see that all modifications lead to a new model and their differences are statistically significant with a p-value of $<0.001$ over \emph{Chi square} test.
	
	Table~\ref{tab:ablation} shows that removing any part from our model hurts the development set accuracy which indicates the effectiveness of these components. Among all components, three of them have noticeable influences: max pooling, difference in the attention stage, and dependent reading. 
	
	Most importantly, the last four study cases in Table~\ref{tab:ablation} (rows 8-11) verify the main intuitions behind our proposed model. They illustrate the importance of our proposed dependent reading strategy which leads to significant improvement, specifically in the encoding stage. We are convinced that the importance of dependent reading in the encoding stage originates from its ability to focus on more important and relevant aspects of the sentences due to its prior knowledge of the other sentence during the encoding procedure.
	
	\begin{figure}[ht]
		\centering
		\includegraphics[width=0.49\textwidth ]{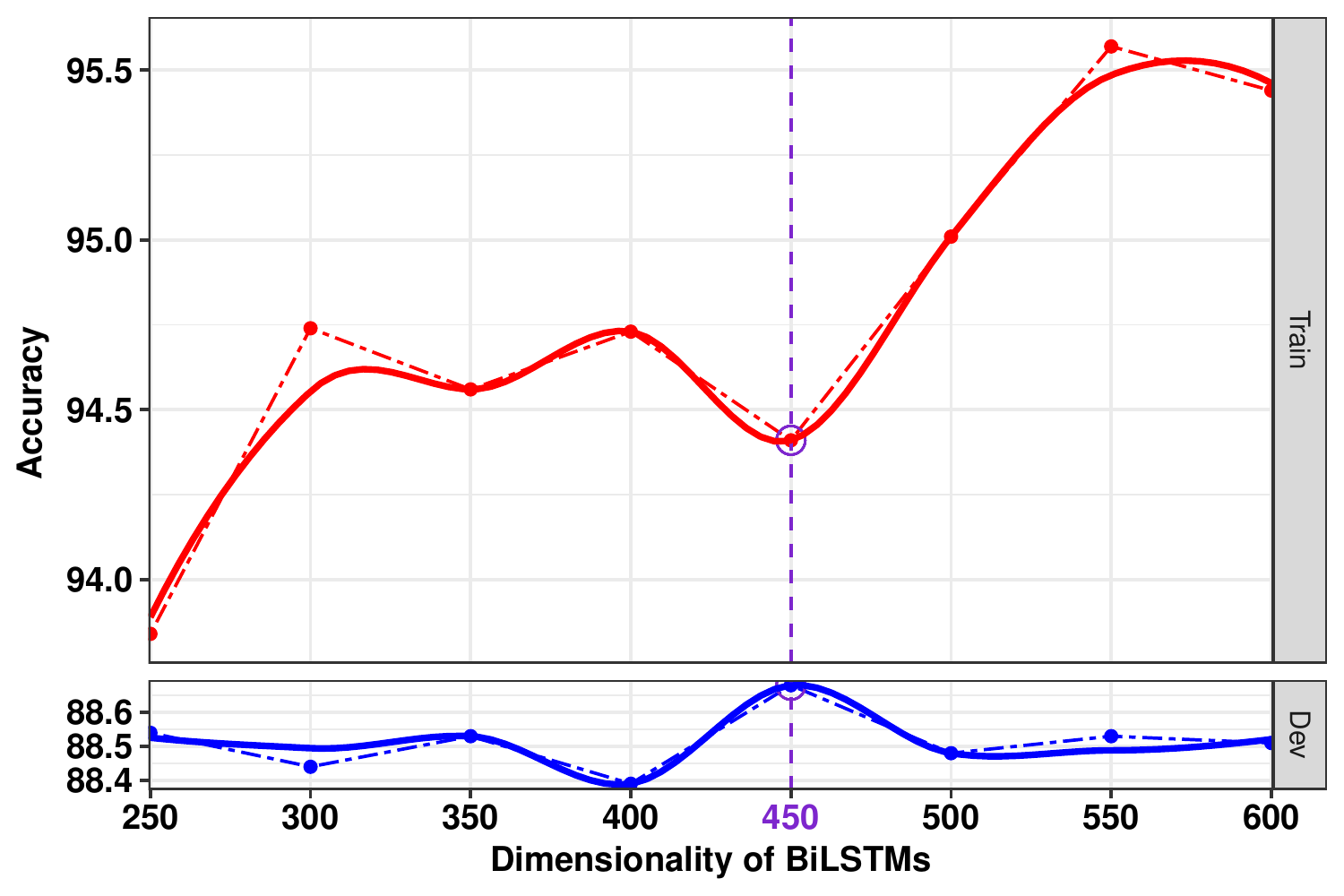}
		\caption{Impact of BiLSTM dimensionality in the proposed model on the training set (red, top) and development set (blue, bottom) accuracies of the SNLI dataset. \label{fig:dim:abl}}
	\end{figure}
	
	Figure~\ref{fig:dim:abl} shows the behavior of the proposed model accuracy on the training set and development set of SNLI. Since the models are selected based on the best observed development set accuracy during the training procedure, the training accuracy curve (red, top) is not strictly increasing. Figure~\ref{fig:dim:abl} demonstrates that we achieve the best performance with 450-dimensional BiLSTMs. In other words, using BiLSTMs with lower dimensionality causes the model to suffer from the lack of space for capturing proper information and dependencies. On the other hand, using higher dimensionality leads to overfitting which hurts the performance on the development set. Hence, we use 450-dimensional BiLSTM in our proposed model.
	
	\subsection{Analysis} \label{sec:err}
	We first investigate the performance of our models categorically. Then, we show a visualization of the energy function in the attention stage (Equation~\ref{eq:energy}) for an instance from the SNLI test set. 
	
	To qualitatively evaluate the performance of our models, we design a set of annotation tags that can be extracted automatically. This design is inspired by the reported annotation tags in \citet{multinli}. The specifications of our annotation tags are as follows:

		\begin{itemize}
		\item \textbf{High Overlap:} premise and hypothesis sentences share more than $70\%$ tokens.
		\item \textbf{Regular Overlap:} sentences share between $30\%$ and $70\%$ tokens.
		\item \textbf{Low Overlap:} sentences share less than $30\%$ tokens.
		\item \textbf{Long Sentence:} either sentence is longer than 20 tokens.
		\item \textbf{Regular Sentence:} premise or hypothesis length is between 5 and 20 tokens.
		\item \textbf{Short Sentence:} either sentence is shorter than 5 tokens.
		\item \textbf{Negation:} negation is present in a sentence.
		\item \textbf{Quantifier:} either of the sentences contains one of the following quantifiers: much, enough, more, most, less, least, no, none, some, any, many, few, several, almost, nearly.
		\item \textbf{Belief:} either of the sentences contains one of the following belief verbs: know, believe, understand, doubt, think, suppose, recognize, forget, remember, imagine, mean, agree, disagree, deny, promise.
	\end{itemize}

	\begin{table}[t]
		\centering
		\small
		\begin{tabular}{lcccr}
			\hline
			Annotation Tag & Freq$^a$ & ESIM & DR(S)$^b$ & DR(E)$^c$ \\
			\hline \hline
			Entailment & 34.3\% & 90.0\% & 89.8\% & 90.9\% \\
			Neutral & 32.8\% & 83.7\% & 85.1\% & 85.6\% \\
			Contradiction & 32.9\% & 90.0\% & 90.5\% & 91.4\% \\ \hline
			High Overlap & 24.3\% & 91.2\% & 90.7\% & 92.1\% \\
			Reg. Overlap & 33.7\% & 87.1\% & 87.9\% & 88.8\% \\
			Low Overlap & 45.4\% & 87.0\% & 87.8\% & 88.4\% \\ \hline
			Long Sentence & 6.4\%  & 92.2\% & 91.3\% & 91.9\% \\
			Reg. Sentence & 74.9\% & 87.8\% & 88.4\% & 89.2\% \\
			Short Sentence & 19.9\% & 87.6\% & 88.1\% & 89.3\% \\ \hline
			Negation & 2.1\% & 82.2\% & 85.7\% & 87.1\% \\
			Quantifier & 8.7\% & 85.5\% & 87.4\% & 87.6\% \\
			Belief & 0.2\% & 78.6\% & 78.6\% & 78.6\% \\
			\hline \hline
			\multicolumn{5}{l}{$^a$Freq, Frequency.} \\
			\multicolumn{5}{l}{$^b$DR(S), DR-BiLSTM (Single).} \\
			\multicolumn{5}{l}{$^c$DR(E), DR-BiLSTM (Ensemble).} \\ \hline
		\end{tabular}
		\caption{\label{tab:snli:deep:analz} Categorical performance analyses (accuracy) of ESIM~\cite{him2017}, DR-BiLSTM (DR(S)) and Ensemble DR-BiLSTM (DR(E)) on the SNLI test set.}
	\end{table}	
	
\begin{figure}[t]
		\centering
		\includegraphics[width=0.49\textwidth ]{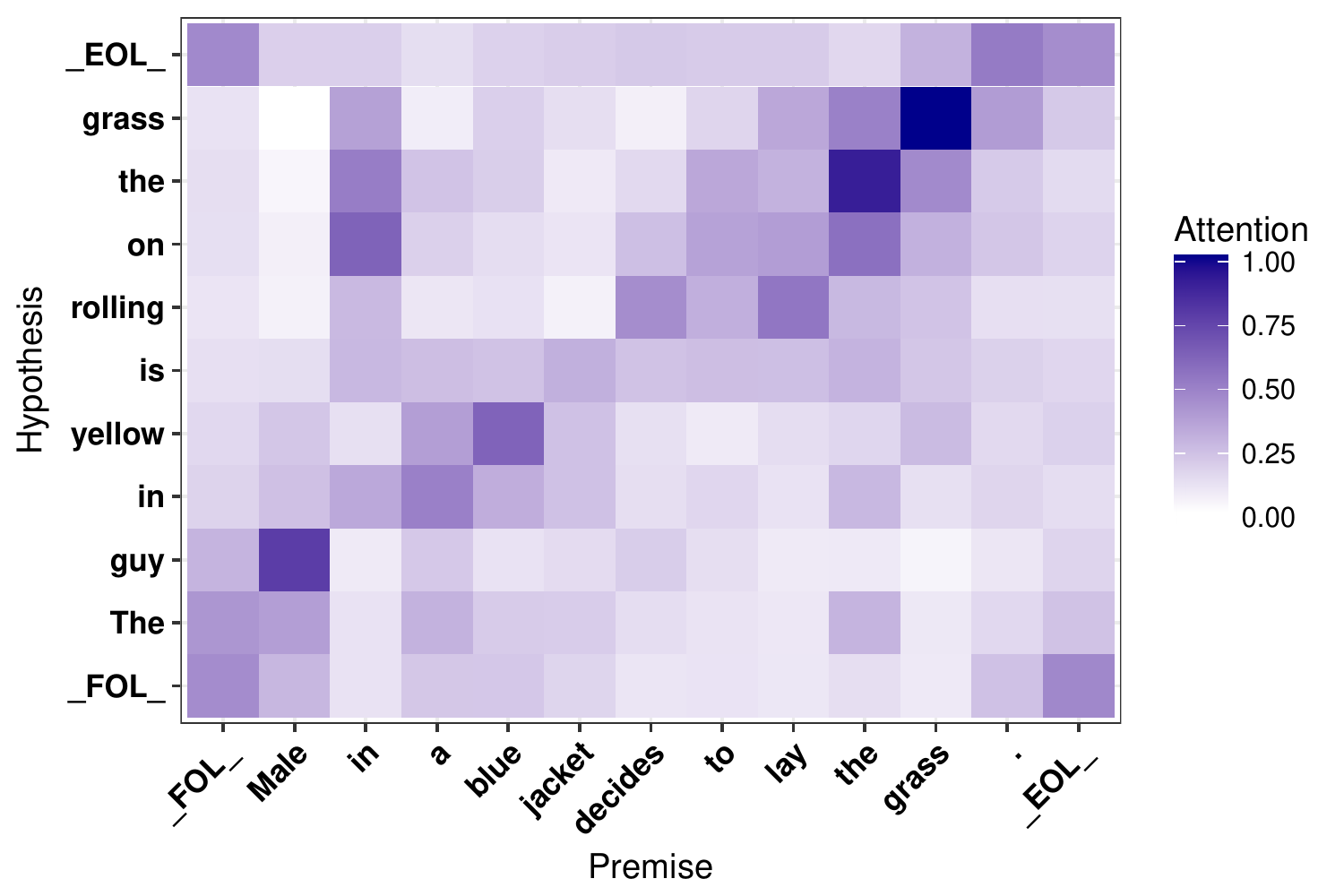}
		\caption{Normalized attention weights for a sample from the SNLI test set. Darker color illustrates higher attention.\label{fig:vis:att}}
	\end{figure}

	Table~\ref{fig:vis:att} shows the frequency of aforementioned annotation tags in the SNLI test set along with the performance (accuracy) of ESIM~\cite{him2017}, DR-BiLSTM (Single), and DR-BiLSTM (Ensemble). Table~\ref{fig:vis:att} can be divided into four major categories: 1) gold label data, 2) word overlap, 3) sentence length, and 4) occurrence of special words. We can see that DR-BiLSTM (Ensemble) performs the best in all categories which matches our expectation. Moreover, DR-BiLSTM (Single) performs noticeably better than ESIM in most of the categories except ``Entailment'', ``High Overlap'', and ``Long Sentence'', for which our model is not far behind (gaps of $0.2\%$, $0.5\%$, and $0.9\%$, respectively). It is noteworthy that DR-BiLSTM (Single) performs better than ESIM in more frequent categories. Specifically, the performance of our model in ``Neutral'', ``Negation'', and ``Quantifier'' categories (improvements of $1.4\%$, $3.5\%$, and $1.9\%$, respectively) indicates the superiority of our model in understanding and disambiguating complex samples. Our investigations indicate that ESIM generates somewhat uniform attention for most of the word pairs while our model could effectively attend to specific parts of the given sentences and provide more meaningful attention. In other words, the dependent reading strategy enables our model to achieve meaningful representations, which leads to better attention to obtain further gains on such categories like Negation and Quantifier sentences (see Section~\ref{app:attcat:sec} in the Appendix for additional details). 
	
	Finally, we show a visualization of the normalized attention weights (energy function, Equation~\ref{eq:energy}) of our model in Figure~\ref{fig:vis:att}. We show a sentence pair, where the premise is ``\emph{Male in a blue jacket decides to lay the grass.}'', and the hypothesis is ``\emph{The guy in yellow is rolling on the grass.}'', and its logical relationship is \emph{contradiction}. Figure~\ref{fig:vis:att} indicates the model's ability in attending to critical pairs of words like $<$Male, guy$>$, $<$decides, rolling$>$, and $<$lay, rolling$>$. Finally, high attention between \{decides, lay\} and \{rolling\}, and \{Male\} and \{guy\} leads the model to correctly classify the sentence pair as \emph{contradiction} (for more samples with attention visualizations, see Section~\ref{app:att:sec} of the Appendix).
	
	\section{Conclusion}
	We propose a novel natural language inference model (DR-BiLSTM) that benefits from a dependent reading strategy and achieves the state-of-the-art results on the SNLI dataset. We also introduce a sophisticated ensemble strategy and illustrate its effectiveness through experimentation. Moreover, we demonstrate the importance of a simple preprocessing step on the performance of our proposed models. Evaluation results show that the preprocessing step allows our DR-BiLSTM (single) model to outperform all previous single and ensemble methods. Similar superior performance is also observed for our DR-BiLSTM (ensemble) model. We show that our ensemble model outperforms the existing state-of-the-art by a considerable margin of $0.7\%$. Finally, we perform an extensive analysis to demonstrate the strength and weakness of the proposed model, which would pave the way for further improvements in this domain.

	\bibliography{naaclhlt2018}

\begin{thebibliography}{}
\expandafter\ifx\csname natexlab\endcsname\relax\def\natexlab#1{#1}\fi

\bibitem[{Bahdanau et~al.(2014)Bahdanau, Cho, and Bengio}]{nmt}
Dzmitry Bahdanau, Kyunghyun Cho, and Yoshua Bengio. 2014.
\newblock \href{http://arxiv.org/abs/1409.0473}{Neural machine translation by
  jointly learning to align and translate}.
\newblock {\em CoRR\/} abs/1409.0473.
\newblock
  \href{http://arxiv.org/abs/1409.0473}{http://arxiv.org/abs/1409.0473}.

\bibitem[{Bowman et~al.(2015)Bowman, Angeli, Potts, and Manning}]{snli}
Samuel~R. Bowman, Gabor Angeli, Christopher Potts, and Christopher~D. Manning.
  2015.
\newblock \href{http://aclweb.org/anthology/D/D15/D15-1075.pdf}{A large
  annotated corpus for learning natural language inference}.
\newblock In {\em Proceedings of the 2015 Conference on Empirical Methods in
  Natural Language Processing, {EMNLP} 2015, Lisbon, Portugal, September 17-21,
  2015\/}. pages 632--642.
\newblock
  \href{http://aclweb.org/anthology/D/D15/D15-1075.pdf}{http://aclweb.org/anthology/D/D15/D15-1075.pdf}.

\bibitem[{Bowman et~al.(2016)Bowman, Gauthier, Rastogi, Gupta, Manning, and
  Potts}]{Bowman2016}
Samuel~R. Bowman, Jon Gauthier, Abhinav Rastogi, Raghav Gupta, Christopher~D.
  Manning, and Christopher Potts. 2016.
\newblock \href{http://aclweb.org/anthology/P/P16/P16-1139.pdf}{A fast unified
  model for parsing and sentence understanding}.
\newblock In {\em Proceedings of the 54th Annual Meeting of the Association for
  Computational Linguistics, {ACL} 2016, August 7-12, 2016, Berlin, Germany,
  Volume 1: Long Papers\/}.
\newblock
  \href{http://aclweb.org/anthology/P/P16/P16-1139.pdf}{http://aclweb.org/anthology/P/P16/P16-1139.pdf}.

\bibitem[{Chen et~al.(2017)Chen, Zhu, Ling, Wei, Jiang, and Inkpen}]{him2017}
Qian Chen, Xiaodan Zhu, Zhen{-}Hua Ling, Si~Wei, Hui Jiang, and Diana Inkpen.
  2017.
\newblock \href{https://doi.org/10.18653/v1/P17-1152}{Enhanced {LSTM} for
  natural language inference}.
\newblock In {\em Proceedings of the 55th Annual Meeting of the Association for
  Computational Linguistics, {ACL} 2017, Vancouver, Canada, July 30 - August 4,
  Volume 1: Long Papers\/}. pages 1657--1668.
\newblock
  \href{https://doi.org/10.18653/v1/P17-1152}{https://doi.org/10.18653/v1/P17-1152}.

\bibitem[{Gong et~al.(2017)Gong, Luo, and Zhang}]{gong2017}
Yichen Gong, Heng Luo, and Jian Zhang. 2017.
\newblock \href{http://arxiv.org/abs/1709.04348}{Natural language inference
  over interaction space}.
\newblock {\em CoRR\/} abs/1709.04348.
\newblock
  \href{http://arxiv.org/abs/1709.04348}{http://arxiv.org/abs/1709.04348}.

\bibitem[{Hochreiter and Schmidhuber(1997)}]{lstm}
Sepp Hochreiter and J{\"{u}}rgen Schmidhuber. 1997.
\newblock \href{https://doi.org/10.1162/neco.1997.9.8.1735}{Long short-term
  memory}.
\newblock {\em Neural Computation\/} 9(8):1735--1780.
\newblock
  \href{https://doi.org/10.1162/neco.1997.9.8.1735}{https://doi.org/10.1162/neco.1997.9.8.1735}.

\bibitem[{Kingma and Ba(2014)}]{adam}
Diederik~P. Kingma and Jimmy Ba. 2014.
\newblock \href{http://arxiv.org/abs/1412.6980}{Adam: {A} method for stochastic
  optimization}.
\newblock {\em CoRR\/} abs/1412.6980.
\newblock
  \href{http://arxiv.org/abs/1412.6980}{http://arxiv.org/abs/1412.6980}.

\bibitem[{Kumar et~al.(2016)Kumar, Irsoy, Ondruska, Iyyer, Bradbury, Gulrajani,
  Zhong, Paulus, and Socher}]{ama}
Ankit Kumar, Ozan Irsoy, Peter Ondruska, Mohit Iyyer, James Bradbury, Ishaan
  Gulrajani, Victor Zhong, Romain Paulus, and Richard Socher. 2016.
\newblock \href{http://jmlr.org/proceedings/papers/v48/kumar16.html}{Ask me
  anything: Dynamic memory networks for natural language processing}.
\newblock In {\em Proceedings of the 33nd International Conference on Machine
  Learning, {ICML} 2016, New York City, NY, USA, June 19-24, 2016\/}. pages
  1378--1387.
\newblock
  \href{http://jmlr.org/proceedings/papers/v48/kumar16.html}{http://jmlr.org/proceedings/papers/v48/kumar16.html}.

\bibitem[{Liu et~al.(2016{\natexlab{a}})Liu, Qiu, Chen, and Huang}]{fusion2016}
Pengfei Liu, Xipeng Qiu, Jifan Chen, and Xuanjing Huang. 2016{\natexlab{a}}.
\newblock \href{http://aclweb.org/anthology/P/P16/P16-1098.pdf}{Deep fusion
  lstms for text semantic matching}.
\newblock In {\em Proceedings of the 54th Annual Meeting of the Association for
  Computational Linguistics, {ACL} 2016, August 7-12, 2016, Berlin, Germany,
  Volume 1: Long Papers\/}.
\newblock
  \href{http://aclweb.org/anthology/P/P16/P16-1098.pdf}{http://aclweb.org/anthology/P/P16/P16-1098.pdf}.

\bibitem[{Liu et~al.(2016{\natexlab{b}})Liu, Sun, Lin, and Wang}]{Liu2016}
Yang Liu, Chengjie Sun, Lei Lin, and Xiaolong Wang. 2016{\natexlab{b}}.
\newblock \href{http://arxiv.org/abs/1605.09090}{Learning natural language
  inference using bidirectional {LSTM} model and inner-attention}.
\newblock {\em CoRR\/} abs/1605.09090.
\newblock
  \href{http://arxiv.org/abs/1605.09090}{http://arxiv.org/abs/1605.09090}.

\bibitem[{MacCartney and Manning(2008)}]{nli}
Bill MacCartney and Christopher~D. Manning. 2008.
\newblock \href{http://www.aclweb.org/anthology/C08-1066}{Modeling semantic
  containment and exclusion in natural language inference}.
\newblock In {\em {COLING} 2008, 22nd International Conference on Computational
  Linguistics, Proceedings of the Conference, 18-22 August 2008, Manchester,
  {UK}\/}. pages 521--528.
\newblock
  \href{http://www.aclweb.org/anthology/C08-1066}{http://www.aclweb.org/anthology/C08-1066}.

\bibitem[{Mou et~al.(2016)Mou, Men, Li, Xu, Zhang, Yan, and Jin}]{Mou2016}
Lili Mou, Rui Men, Ge~Li, Yan Xu, Lu~Zhang, Rui Yan, and Zhi Jin. 2016.
\newblock \href{http://aclweb.org/anthology/P/P16/P16-2022.pdf}{Natural
  language inference by tree-based convolution and heuristic matching}.
\newblock In {\em Proceedings of the 54th Annual Meeting of the Association for
  Computational Linguistics, {ACL} 2016, August 7-12, 2016, Berlin, Germany,
  Volume 2: Short Papers\/}.
\newblock
  \href{http://aclweb.org/anthology/P/P16/P16-2022.pdf}{http://aclweb.org/anthology/P/P16/P16-2022.pdf}.

\bibitem[{Parikh et~al.(2016)Parikh, T{\"{a}}ckstr{\"{o}}m, Das, and
  Uszkoreit}]{google2016}
Ankur~P. Parikh, Oscar T{\"{a}}ckstr{\"{o}}m, Dipanjan Das, and Jakob
  Uszkoreit. 2016.
\newblock \href{http://aclweb.org/anthology/D/D16/D16-1244.pdf}{A decomposable
  attention model for natural language inference}.
\newblock In {\em Proceedings of the 2016 Conference on Empirical Methods in
  Natural Language Processing, {EMNLP} 2016, Austin, Texas, USA, November 1-4,
  2016\/}. pages 2249--2255.
\newblock
  \href{http://aclweb.org/anthology/D/D16/D16-1244.pdf}{http://aclweb.org/anthology/D/D16/D16-1244.pdf}.

\bibitem[{Pennington et~al.(2014)Pennington, Socher, and Manning}]{glove}
Jeffrey Pennington, Richard Socher, and Christopher~D. Manning. 2014.
\newblock \href{http://www.aclweb.org/anthology/D14-1162}{Glove: Global vectors
  for word representation}.
\newblock In {\em Empirical Methods in Natural Language Processing (EMNLP)\/}.
  pages 1532--1543.
\newblock
  \href{http://www.aclweb.org/anthology/D14-1162}{http://www.aclweb.org/anthology/D14-1162}.

\bibitem[{Plummer et~al.(2015)Plummer, Wang, Cervantes, Caicedo, Hockenmaier,
  and Lazebnik}]{flickr}
Bryan~A. Plummer, Liwei Wang, Chris~M. Cervantes, Juan~C. Caicedo, Julia
  Hockenmaier, and Svetlana Lazebnik. 2015.
\newblock \href{https://doi.org/10.1109/ICCV.2015.303}{Flickr30k entities:
  Collecting region-to-phrase correspondences for richer image-to-sentence
  models}.
\newblock In {\em 2015 {IEEE} International Conference on Computer Vision,
  {ICCV} 2015, Santiago, Chile, December 7-13, 2015\/}. pages 2641--2649.
\newblock
  \href{https://doi.org/10.1109/ICCV.2015.303}{https://doi.org/10.1109/ICCV.2015.303}.

\bibitem[{Rockt{\"{a}}schel et~al.(2015)Rockt{\"{a}}schel, Grefenstette,
  Hermann, Kocisk{\'{y}}, and Blunsom}]{Rocktaschel2015}
Tim Rockt{\"{a}}schel, Edward Grefenstette, Karl~Moritz Hermann, Tom{\'{a}}s
  Kocisk{\'{y}}, and Phil Blunsom. 2015.
\newblock \href{http://arxiv.org/abs/1509.06664}{Reasoning about entailment
  with neural attention}.
\newblock {\em CoRR\/} abs/1509.06664.
\newblock
  \href{http://arxiv.org/abs/1509.06664}{http://arxiv.org/abs/1509.06664}.

\bibitem[{Sha et~al.(2016)Sha, Chang, Sui, and Li}]{re-read}
Lei Sha, Baobao Chang, Zhifang Sui, and Sujian Li. 2016.
\newblock \href{http://aclweb.org/anthology/C/C16/C16-1270.pdf}{Reading and
  thinking: Re-read {LSTM} unit for textual entailment recognition}.
\newblock In {\em {COLING} 2016, 26th International Conference on Computational
  Linguistics, Proceedings of the Conference: Technical Papers, December 11-16,
  2016, Osaka, Japan\/}. pages 2870--2879.
\newblock
  \href{http://aclweb.org/anthology/C/C16/C16-1270.pdf}{http://aclweb.org/anthology/C/C16/C16-1270.pdf}.

\bibitem[{Srivastava et~al.(2014)Srivastava, Hinton, Krizhevsky, Sutskever, and
  Salakhutdinov}]{dropout}
Nitish Srivastava, Geoffrey~E. Hinton, Alex Krizhevsky, Ilya Sutskever, and
  Ruslan Salakhutdinov. 2014.
\newblock \href{http://dl.acm.org/citation.cfm?id=2670313}{Dropout: a simple
  way to prevent neural networks from overfitting}.
\newblock {\em Journal of Machine Learning Research\/} 15(1):1929--1958.
\newblock
  \href{http://dl.acm.org/citation.cfm?id=2670313}{http://dl.acm.org/citation.cfm?id=2670313}.

\bibitem[{Vendrov et~al.(2015)Vendrov, Kiros, Fidler, and
  Urtasun}]{Vendrov2015}
Ivan Vendrov, Ryan Kiros, Sanja Fidler, and Raquel Urtasun. 2015.
\newblock \href{http://arxiv.org/abs/1511.06361}{Order-embeddings of images and
  language}.
\newblock {\em CoRR\/} abs/1511.06361.
\newblock
  \href{http://arxiv.org/abs/1511.06361}{http://arxiv.org/abs/1511.06361}.

\bibitem[{Wang and Jiang(2016)}]{Wang2016}
Shuohang Wang and Jing Jiang. 2016.
\newblock \href{http://aclweb.org/anthology/N/N16/N16-1170.pdf}{Learning
  natural language inference with {LSTM}}.
\newblock In {\em {NAACL} {HLT} 2016, The 2016 Conference of the North American
  Chapter of the Association for Computational Linguistics: Human Language
  Technologies, San Diego California, USA, June 12-17, 2016\/}. pages
  1442--1451.
\newblock
  \href{http://aclweb.org/anthology/N/N16/N16-1170.pdf}{http://aclweb.org/anthology/N/N16/N16-1170.pdf}.

\bibitem[{Wang et~al.(2017)Wang, Hamza, and Florian}]{ibm2017}
Zhiguo Wang, Wael Hamza, and Radu Florian. 2017.
\newblock \href{https://doi.org/10.24963/ijcai.2017/579}{Bilateral
  multi-perspective matching for natural language sentences}.
\newblock In {\em Proceedings of the Twenty-Sixth International Joint
  Conference on Artificial Intelligence, {IJCAI} 2017, Melbourne, Australia,
  August 19-25, 2017\/}. pages 4144--4150.
\newblock
  \href{https://doi.org/10.24963/ijcai.2017/579}{https://doi.org/10.24963/ijcai.2017/579}.

\bibitem[{Williams et~al.(2017)Williams, Nangia, and Bowman}]{multinli}
Adina Williams, Nikita Nangia, and Samuel~R. Bowman. 2017.
\newblock \href{http://arxiv.org/abs/1704.05426}{\-coverage challenge corpus
  for sentence understanding through inference}.
\newblock {\em CoRR\/} abs/1704.05426.
\newblock
  \href{http://arxiv.org/abs/1704.05426}{http://arxiv.org/abs/1704.05426}.

\bibitem[{Yu and Munkhdalai(2017{\natexlab{a}})}]{nse2017}
Hong Yu and Tsendsuren Munkhdalai. 2017{\natexlab{a}}.
\newblock
  \href{http://aclanthology.info/papers/E17-1038/neural-semantic-encoders}{Neural
  semantic encoders}.
\newblock In {\em Proceedings of the 15th Conference of the European Chapter of
  the Association for Computational Linguistics, {EACL} 2017, Valencia, Spain,
  April 3-7, 2017, Volume 1: Long Papers\/}. pages 397--407.
\newblock
  \href{http://aclanthology.info/papers/E17-1038/neural-semantic-encoders}{http://aclanthology.info/papers/E17-1038/neural-semantic-encoders}.

\bibitem[{Yu and Munkhdalai(2017{\natexlab{b}})}]{nti2017}
Hong Yu and Tsendsuren Munkhdalai. 2017{\natexlab{b}}.
\newblock
  \href{http://aclanthology.info/papers/E17-1002/neural-tree-indexers-for-text-understanding}{Neural
  tree indexers for text understanding}.
\newblock In {\em Proceedings of the 15th Conference of the European Chapter of
  the Association for Computational Linguistics, {EACL} 2017, Valencia, Spain,
  April 3-7, 2017, Volume 1: Long Papers\/}. pages 11--21.
\newblock
  \href{http://aclanthology.info/papers/E17-1002/neural-tree-indexers-for-text-understanding}{http://aclanthology.info/papers/E17-1002/neural-tree-indexers-for-text-understanding}.

\bibitem[{Zhao et~al.(2016)Zhao, Huang, and Ma}]{kai_2016}
Kai Zhao, Liang Huang, and Mingbo Ma. 2016.
\newblock \href{http://aclweb.org/anthology/C/C16/C16-1212.pdf}{Textual
  entailment with structured attentions and composition}.
\newblock In {\em {COLING} 2016, 26th International Conference on Computational
  Linguistics, Proceedings of the Conference: Technical Papers, December 11-16,
  2016, Osaka, Japan\/}. pages 2248--2258.
\newblock
  \href{http://aclweb.org/anthology/C/C16/C16-1212.pdf}{http://aclweb.org/anthology/C/C16/C16-1212.pdf}.

\end{thebibliography}
	\bibliographystyle{acl_natbib}
	
	\appendix
	
	\section{Ensemble Strategy Study} 
	\label{app:ensemble:sec}
	We use the following configurations in our ensemble model study:
	
	\begin{itemize}
		\item DR-BiLSTM (with different initialization seeds): here, we consider 6 DR-BiLSTMs with different initialization seeds.
		
		\item \emph{tanh}-Projection:  same configuration as DR-BiLSTM, but we use \emph{tanh} instead of \emph{ReLU} as the activation function in Equations~\ref{eq:prj:p} and \ref{eq:prj:h} in the paper:
		\begin{equation}
		p_i =  \textit{tanh}(W_p a_i + b_p)
		\end{equation}
		\begin{equation}
		q_j =  \textit{tanh}(W_p b_j + b_p)
		\end{equation}
		
		\item DR-BiLSTM (with 1 round of dependent reading): same configuration as DR-BiLSTM, but we do not use dependent reading during the inference process. In other words, we use $\tilde{p} = \bar{p}$ and $\tilde{q} = \bar{q}$ instead of Equations~\ref{eq:mpool:p} and \ref{eq:mpool:h} in the paper respectively.
		
		\item DR-BiLSTM (with 3 rounds of dependent reading): same configuration as the above, but we use 3 rounds of dependent reading. Formally, we replace Equations~\ref{eq:enc:pout} and \ref{eq:enc:hout} in the paper with the following equations respectively:
		
		\begin{equation}
		\begin{split}
		-, s_v &= \textit{BiLSTM}(v, 0) \\
		-, s_{vu} &= \textit{BiLSTM}(u, s_v) \\
		-, s_{vuv} &= \textit{BiLSTM}(v, s_{vu}) \\
		\hat{u}, - &= \textit{BiLSTM}(u, s_{vuv})
		\end{split}
		\label{eq:enc:rpout}
		\end{equation}
		
		\begin{equation}
		\begin{split}
		-, s_u &= \textit{BiLSTM}(u, 0) \\
		-, s_{uv} &= \textit{BiLSTM}(v, s_u) \\
		-, s_{uvu} &= \textit{BiLSTM}(u, s_{uv}) \\
		\hat{v}, - &= \textit{BiLSTM}(v, s_{uvu})
		\end{split}
		\label{eq:enc:rhout}
		\end{equation}
		
	\end{itemize}
	
	Our final ensemble model, DR-BiLSTM (Ensemble) is the combination of the following 6 models: tanh-Projection, DR-BiLSTM (with 1 round of dependent reading), DR-BiLSTM (with 3 rounds of dependent reading), and 3 DR-BiLSTMs with different initialization seeds.
	
\begin{figure}[t]
		\centering
		\includegraphics[width=0.49\textwidth ]{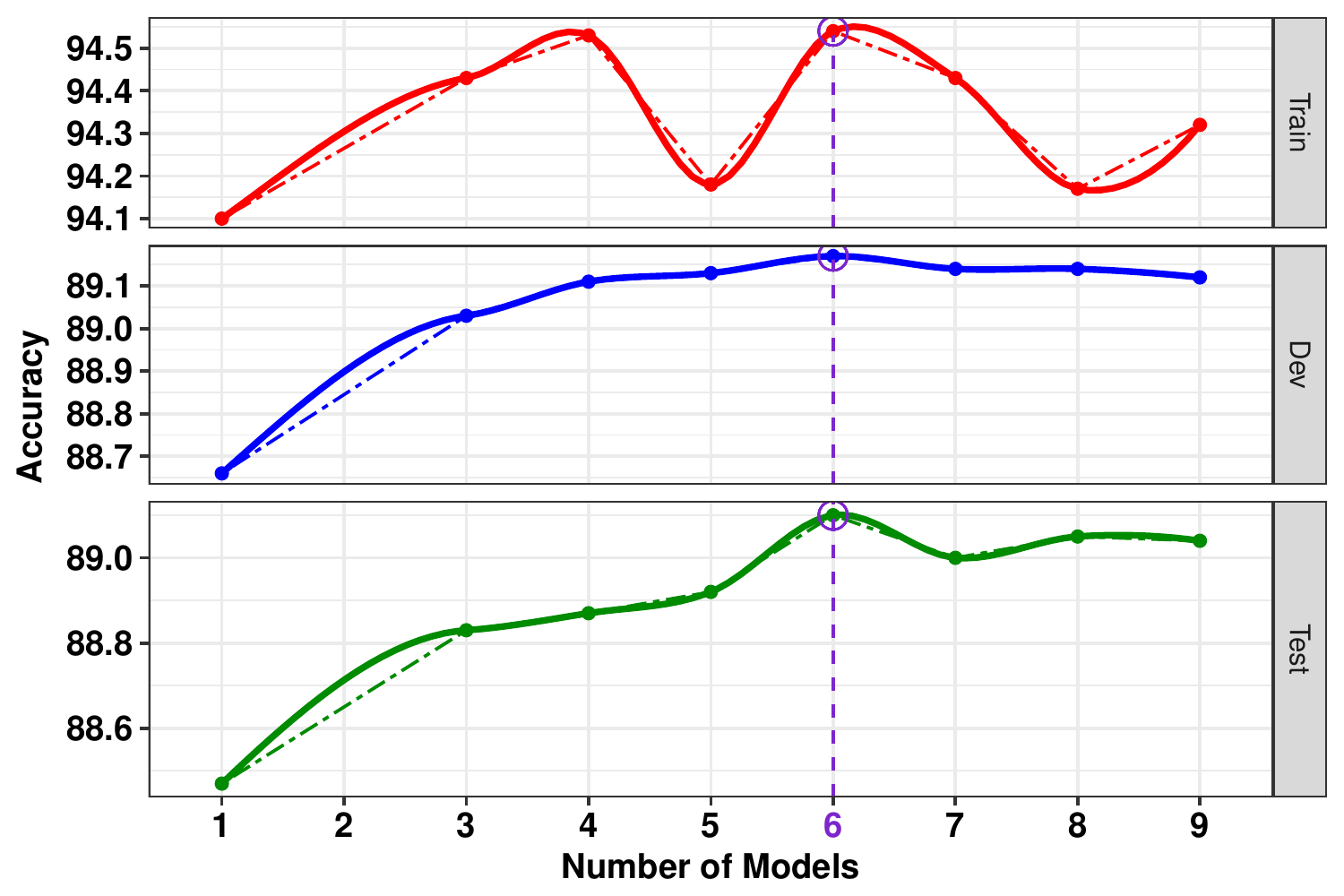}
		\caption{Performance of $n$ ensemble models using majority voting on natural language inference reported for training set (red, top), development set (blue, middle), and test set (green, bottom) of SNLI. The best performance on development set is used as the criteria to determine the final ensemble. The best performance on development set is observed using 6 models. \label{fig:mj:ensemble}}
	\end{figure}

	We also experiment with majority voting and averaging the probability distribution strategies for ensemble models using the same set of models as our weighted averaging ensemble method (as described above). Figure~\ref{fig:mj:ensemble} shows the behavior of the majority voting strategy with different number of models. Interestingly, the best development accuracy is also observed using 6 individual models including tanh-Projection, DR-BiLSTM (with 1 round of dependent reading), DR-BiLSTM (with 3 rounds of dependent reading), and 3 DR-BiLSTMs with varying initialization seeds that are different from our DR-BiLSTM (Ensemble) model.
	
	\begin{table*}[ht]
		\begin{center}
			\small
			\begin{tabular}{ll}
				\hline
				\textbf{Original Sentence} & \textbf{Corrected Sentence}  \\ \hline\hline
				\textbf{\textit{Froends}} ride in an open top vehicle together. & \textbf{\textit{Friends}} ride in an open top vehicle together. \\
				A middle \textbf{\textit{easten}} store. & A middle \textbf{\textit{eastern}} store. \\
				A woman is looking at a \textbf{\textit{phtographer}} & A woman is looking at a \textbf{\textit{photographer}} \\
				The mother and daughter are \textbf{\textit{fighitn}}. & The mother and daughter are \textbf{\textit{fighting}}. \\
				Two \textbf{\textit{kiled}} men hold bagpipes & Two \textbf{\textit{killed}} men hold bagpipes \\
				A woman escapes a from a hostile \textbf{\textit{enviroment}} & A woman escapes a from a hostile \textbf{\textit{environment}} \\
				Two \textbf{\textit{daschunds}} play with a red ball & Two \textbf{\textit{dachshunds}} play with a red ball \\
				A black dog is running through a \textbf{\textit{marsh-like}} area. & A black dog is running through a \textbf{\textit{marsh like}} area. \\
				a singer wearing a \textbf{\textit{jacker}} performs on stage & a singer wearing a \textbf{\textit{jacket}} performs on stage \\
				There is a \textbf{\textit{sculture}} & There is a \textbf{\textit{sculpture}} \\
				Taking a \textbf{\textit{neverending}} break & Taking a \textbf{\textit{never ending}} break \\
				The woman has sounds \textbf{\textit{emanting}} from her mouth. & The woman has sounds \textbf{\textit{emanating}} from her mouth. \\
				the lady is \textbf{\textit{shpping}} & the lady is \textbf{\textit{shopping}} \\
				A Bugatti and a \textbf{\textit{Lambourgini}} compete in a road race. & A Bugatti and a \textbf{\textit{Lamborghini}} compete in a road race.  \\ \hline
			\end{tabular}
		\end{center}
		\caption{\label{tab:mis:data} Examples of original sentences that contain erroneous words (misspelled) in the test set of SNLI along with their corrected counterparts. Erroneous words are shown in \textbf{\textit{bold and italic}}.}
	\end{table*}
	
	We should note that our weighted averaging ensemble strategy performs better than the majority voting method in both development set and test set of SNLI, which indicates the effectiveness of our approach. Furthermore, our method could show more consistent behavior for training and test sets when we increased the number of models (see Figure~\ref{fig:ensemble} in Section \ref{sec:es} of the paper). According to our observations, averaging the probability distributions fails to improve the development set accuracy using two and three models, so we did not study it further.
	
	\section{Preprocessing Study}
	\label{app:preproc:sec}
	
\begin{figure*}[ht]
		\begin{center}
			\subfigure[Erroneous sample (daschunds in premise).]{%
				\includegraphics[width=0.49\textwidth]{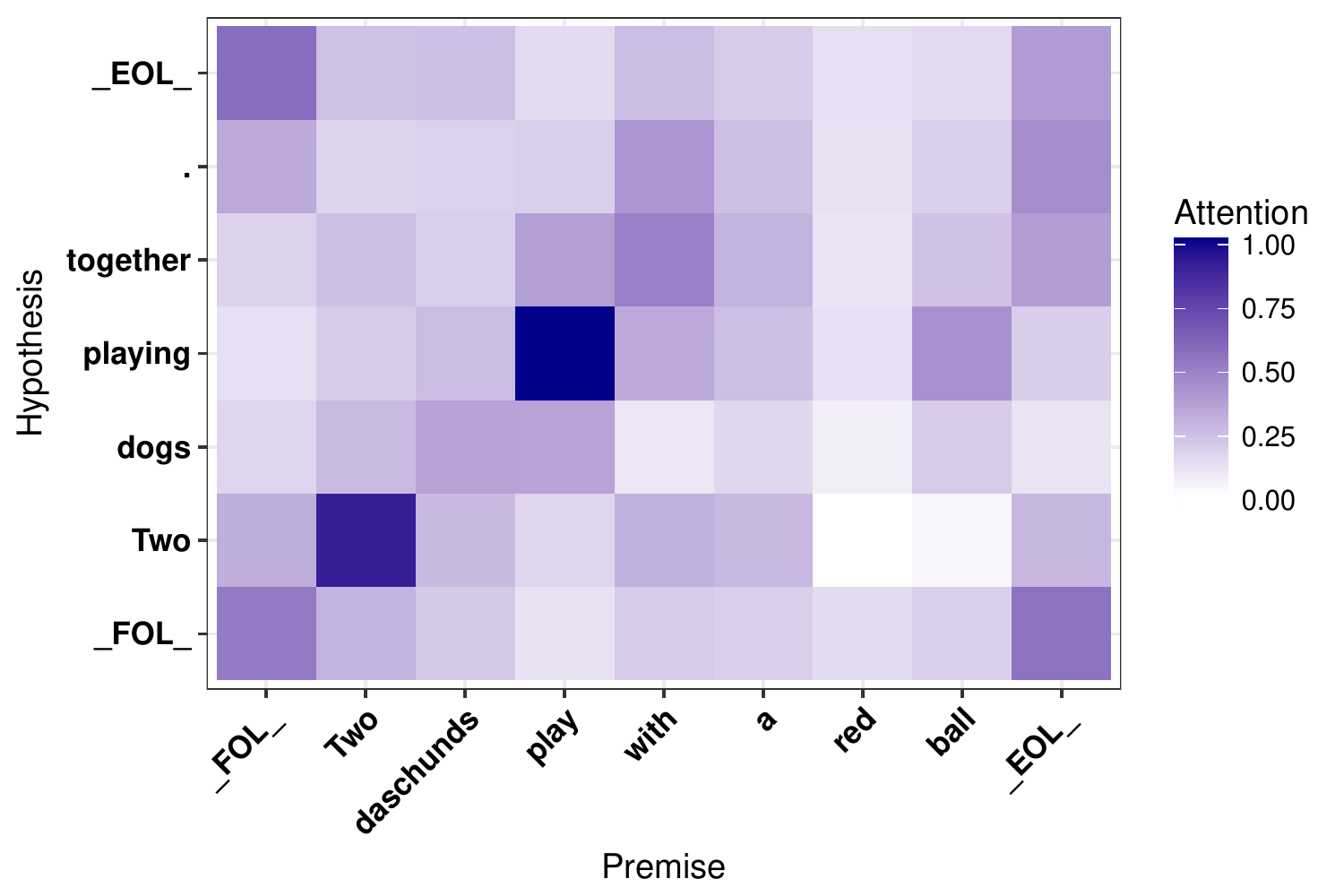}
			}%
			\subfigure[Fixed sample  (dachshunds in premise).]{%
				\includegraphics[width=0.49\textwidth]{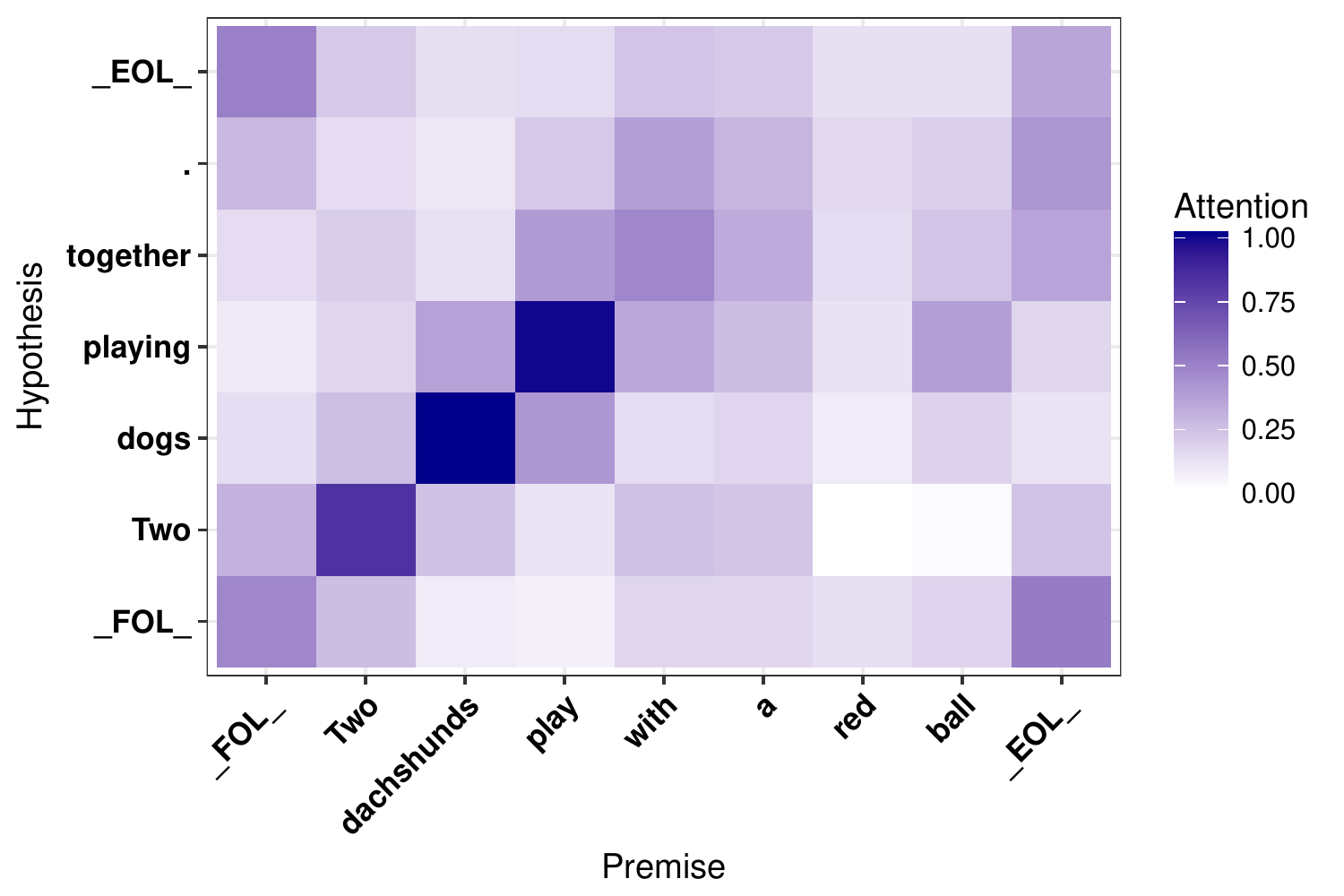}
			}
		\end{center}
		\caption{
			Visualization of the energy function for one erroneous sample (a) and the fixed sample (b). The gold label is \emph{Entailment}. Our model returns \emph{Contradiction} for the erroneous sample, but correctly classifies the fixed sample.
		}
		\label{fig:att:miss1}
	\end{figure*}

\begin{figure*}[ht]
		\begin{center}
			\subfigure[Erroneous sample (daschunds in premise).]{%
				\includegraphics[width=0.49\textwidth]{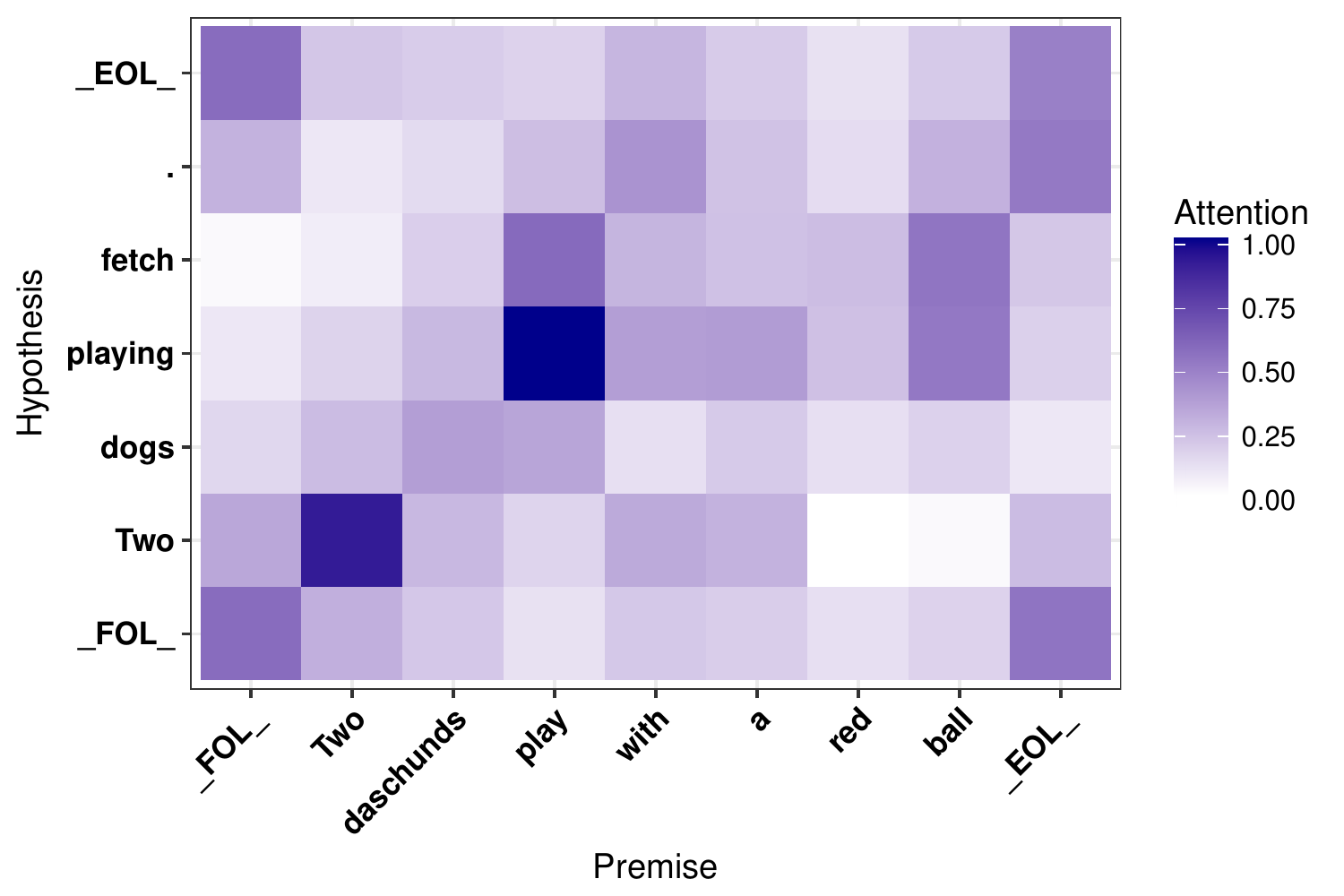}
			}%
			\subfigure[Fixed sample  (dachshunds in premise).]{%
				\includegraphics[width=0.49\textwidth]{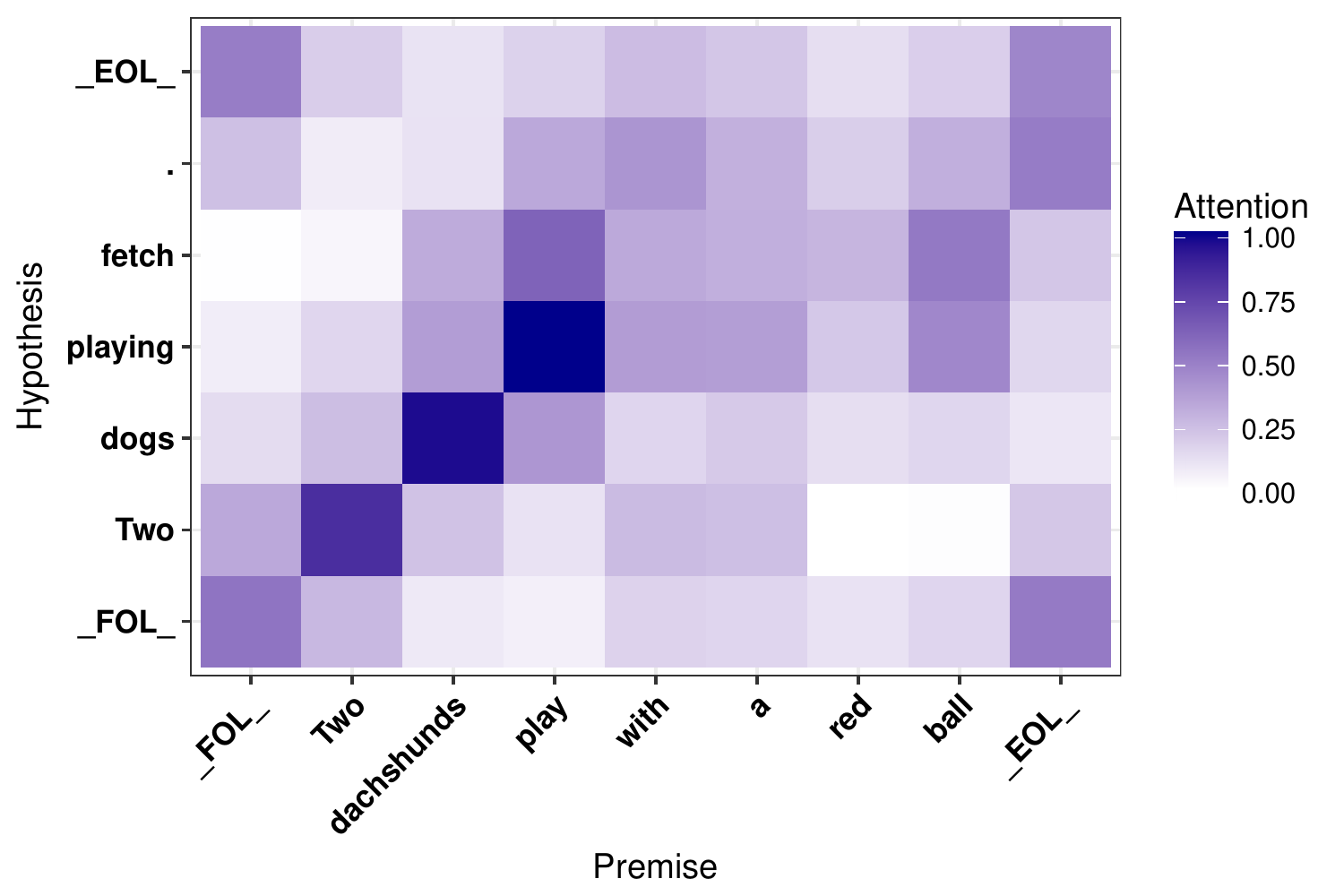}
			}
		\end{center}
		\caption{
			Visualization of the energy function for one erroneous sample (a) and the fixed sample (b). The gold label is \emph{Neutral}. Our model returns \emph{Contradiction} for the erroneous sample, but correctly classifies the fixed sample.
		}
		\label{fig:att:miss2}
	\end{figure*}

\begin{figure*}[ht]
		\begin{center}
			\subfigure[Erroneous sample (Froends in hypothesis).]{%
				\includegraphics[width=0.49\textwidth]{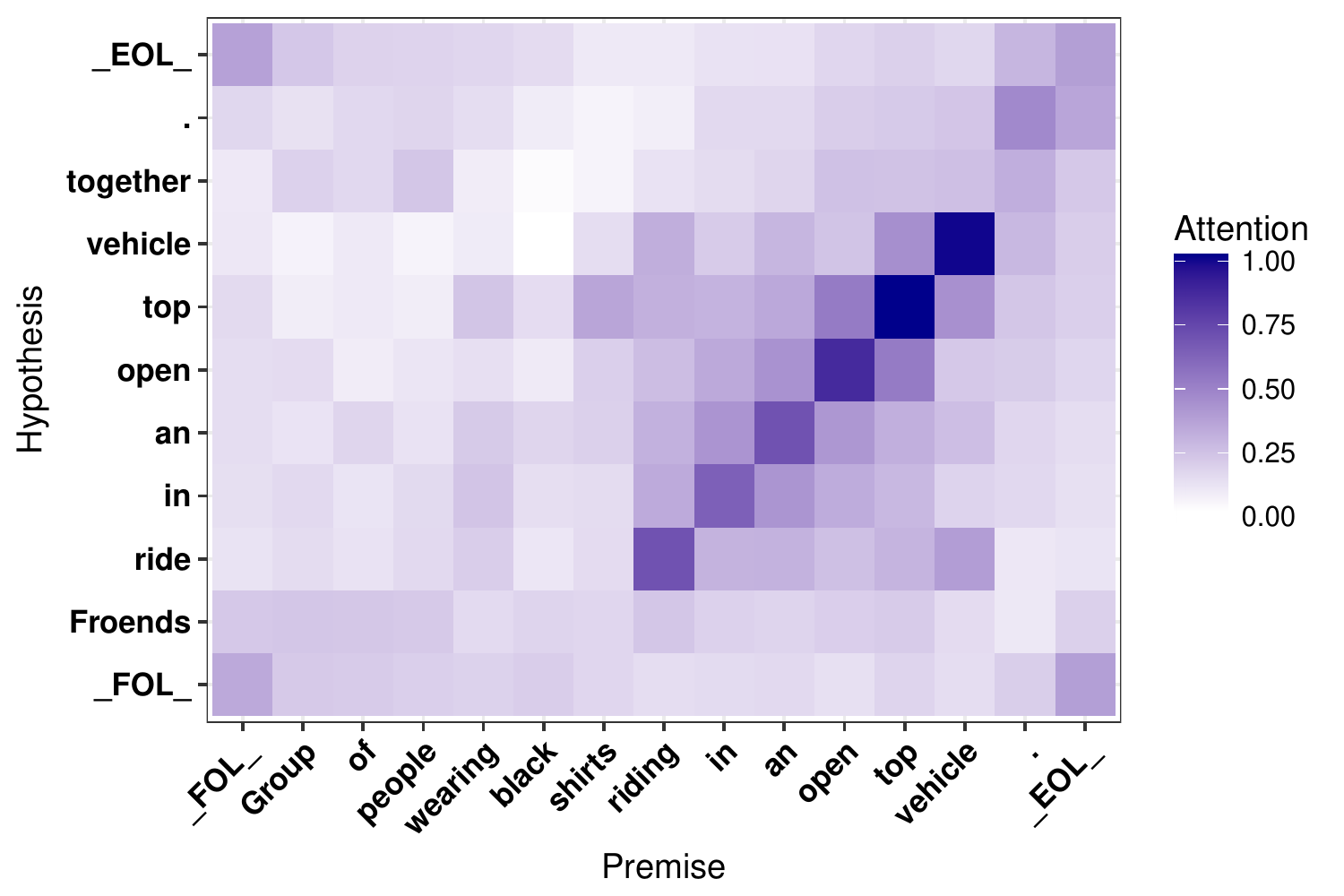}
			}%
			\subfigure[Fixed sample  (Friends in hypothesis).]{%
				\includegraphics[width=0.49\textwidth]{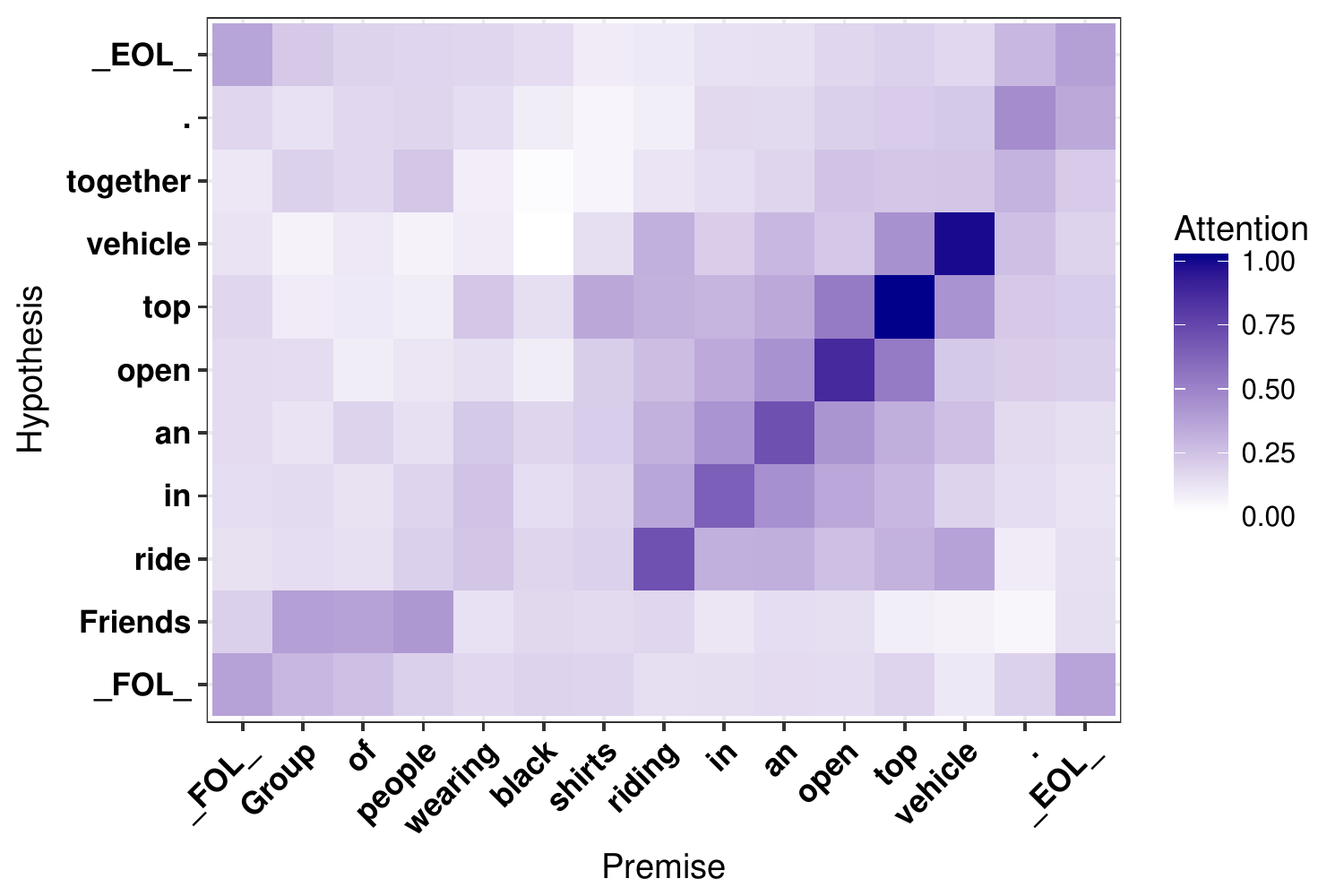}
			}
		\end{center}
		\caption{
			Visualization of the energy function for one erroneous sample (a) and the fixed sample (b). The gold label is \emph{Neutral}. Our model returns \emph{Entailment} for the erroneous sample, but correctly classifies the fixed sample.
		}
		\label{fig:att:miss3}
	\end{figure*}

\begin{figure*}[ht]
		\begin{center}
			\subfigure[Erroneous sample (easten in hypothesis).]{%
				\includegraphics[width=0.49\textwidth]{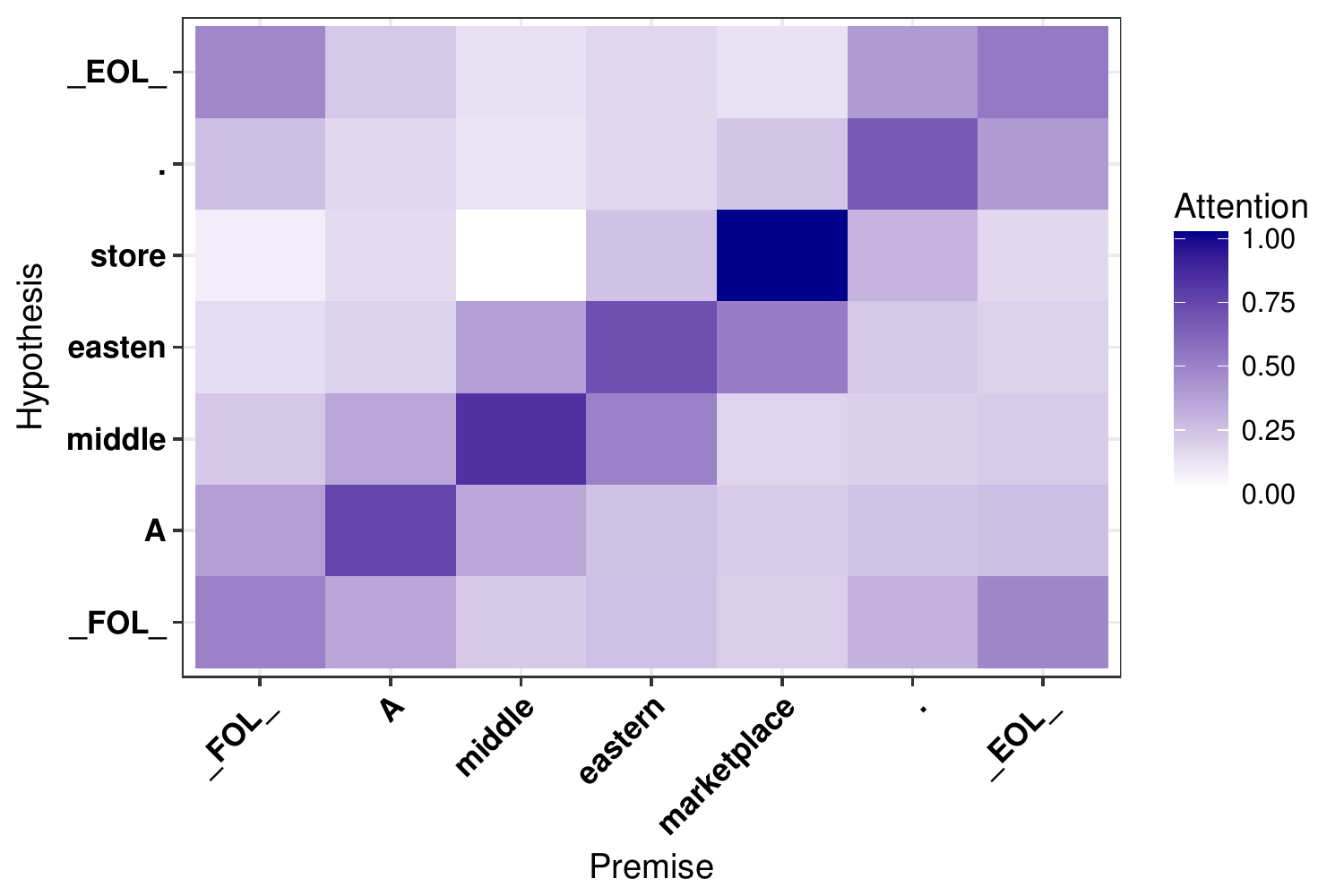}
			}%
			\subfigure[Fixed sample  (eastern in hypothesis).]{%
				\includegraphics[width=0.49\textwidth]{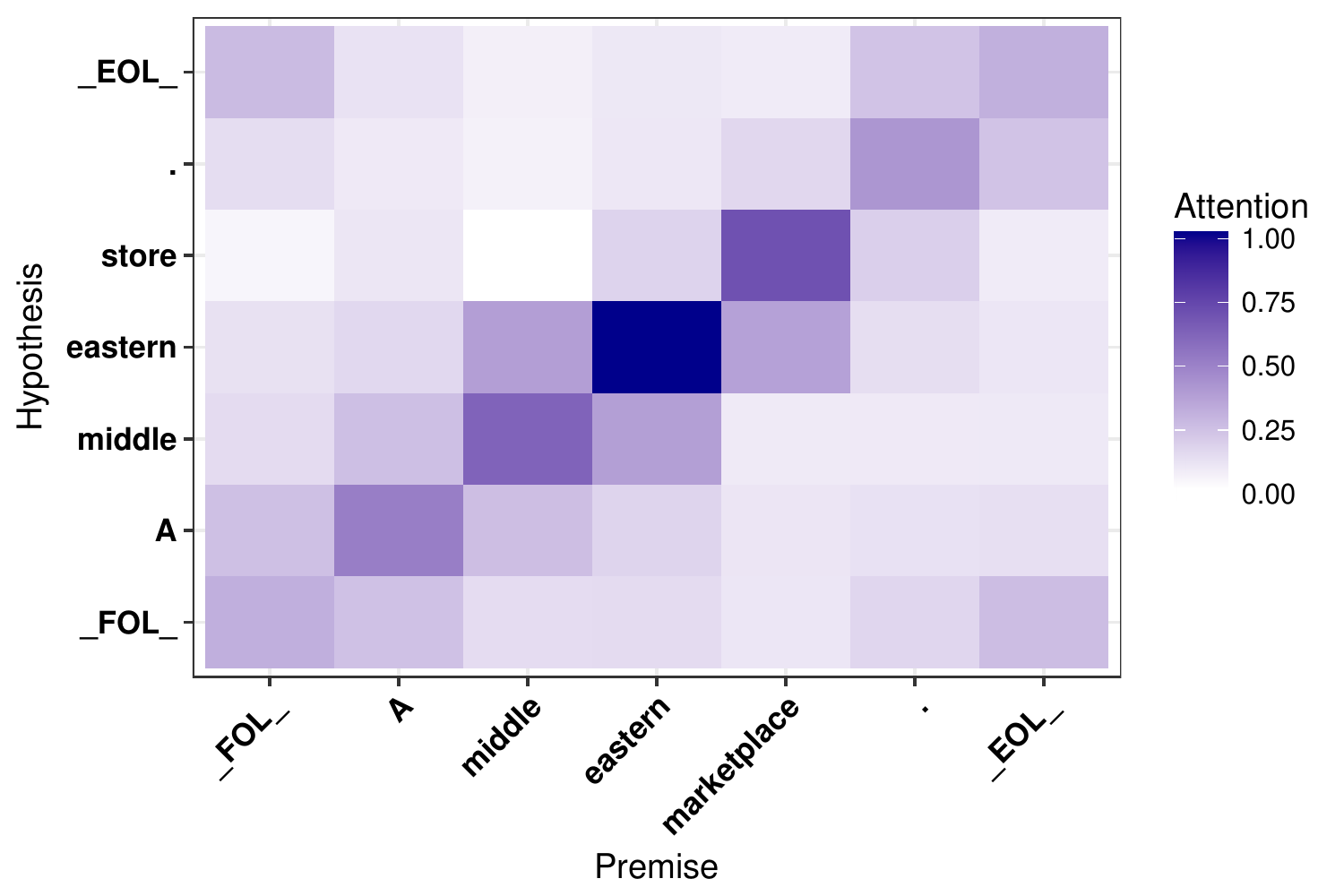}
			}
		\end{center}
		\caption{
			Visualization of the energy function for one erroneous sample (a) and the fixed sample (b). The gold label is \emph{Entailment}. Our model returns \emph{Contradiction} for the erroneous sample, but correctly classifies the fixed sample.
		}
		\label{fig:att:miss4}
	\end{figure*}

\begin{figure*}[ht]
		\begin{center}
			\subfigure[Erroneous sample (jacker in hypothesis).]{%
				\includegraphics[width=0.49\textwidth]{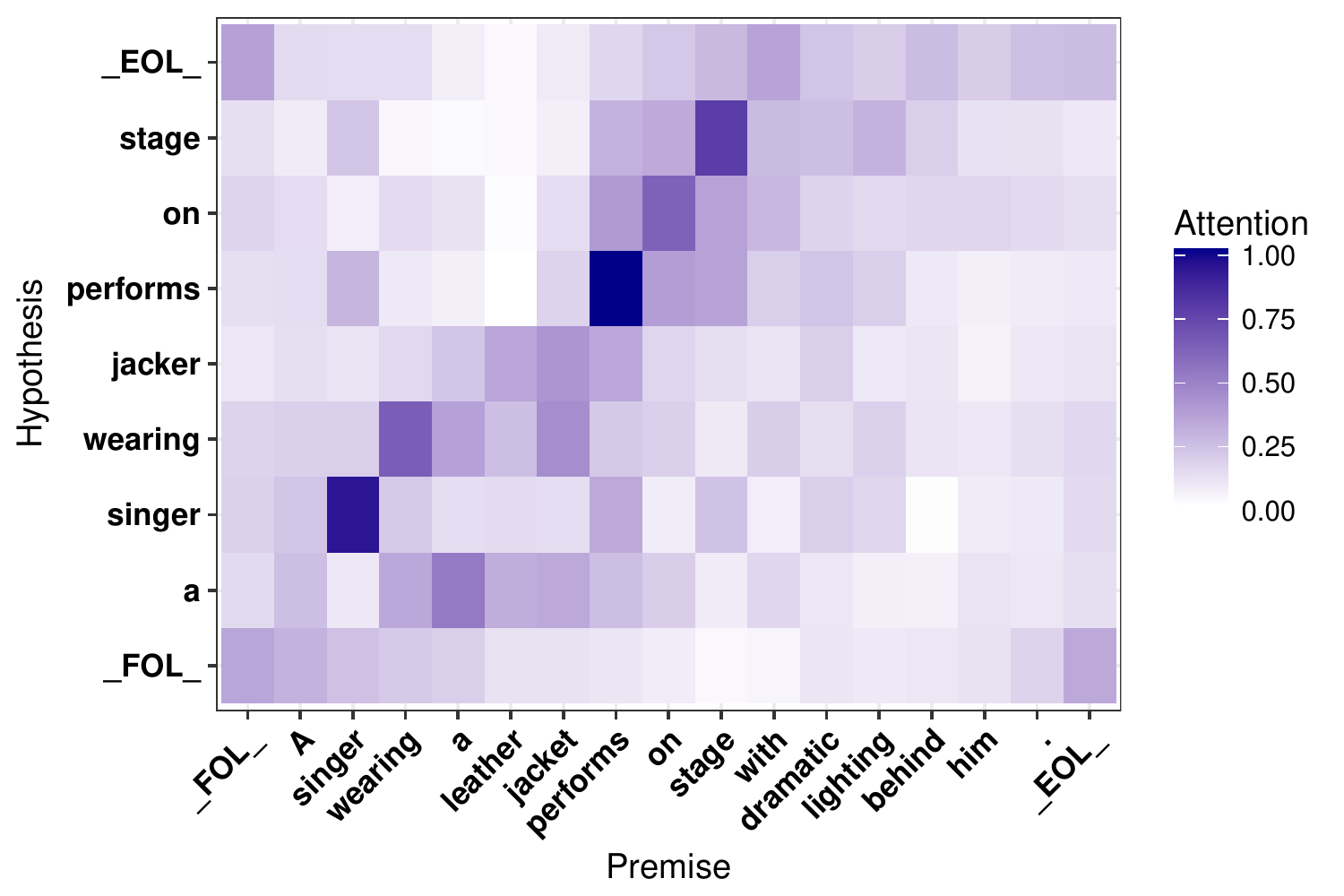}
			}%
			\subfigure[Fixed sample  (jacket in hypothesis).]{%
				\includegraphics[width=0.49\textwidth]{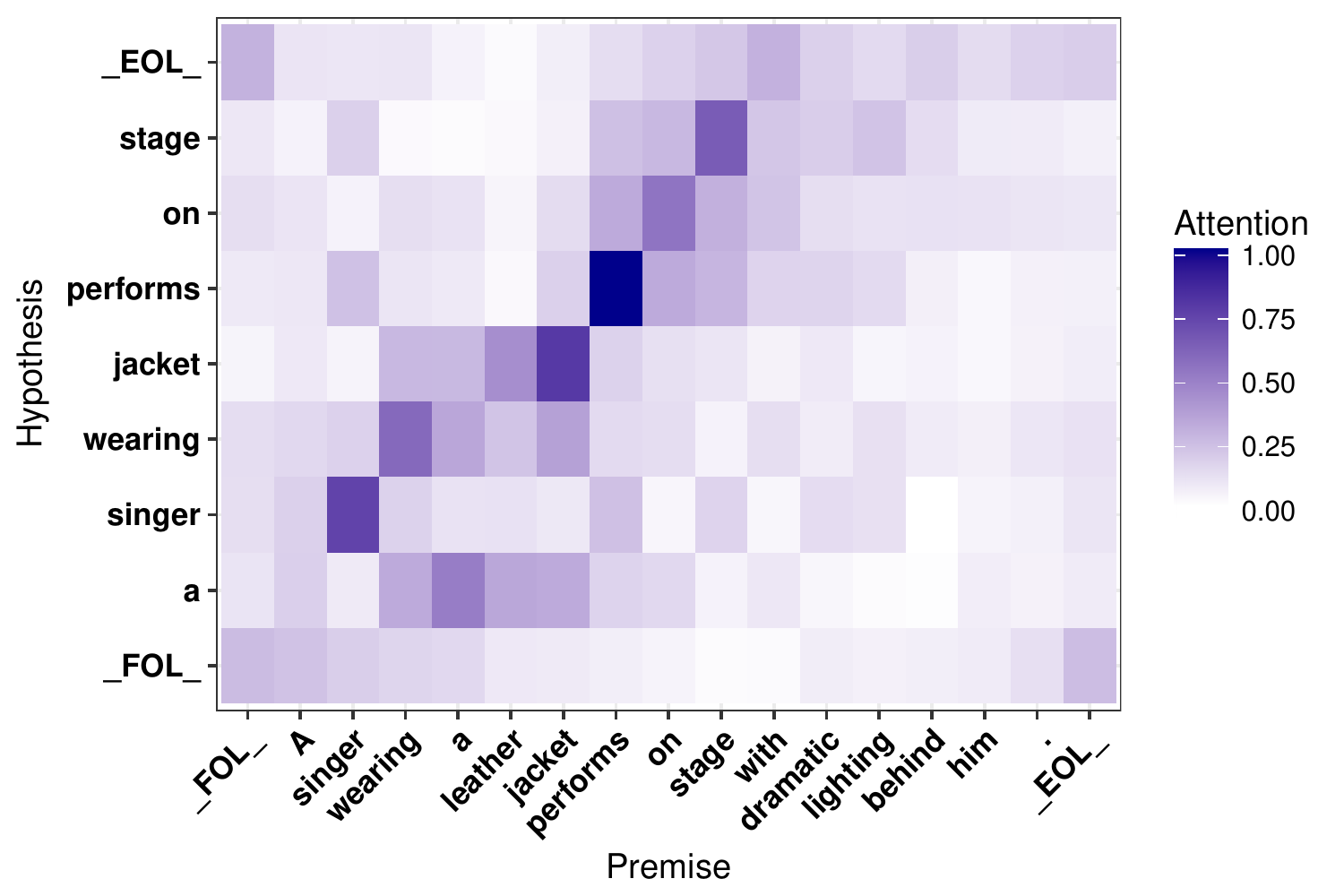}
			}
		\end{center}
		\caption{
			Visualization of the energy function for one erroneous sample (a) and the fixed sample (b). The gold label is \emph{Entailment}. Our model returns \emph{Neutral} for the erroneous sample, but correctly classifies the fixed sample.
		}
		\label{fig:att:miss5}
	\end{figure*}

\begin{figure*}[ht]
		\begin{center}
			\subfigure[Erroneous sample (sculture in hypothesis).]{%
				\includegraphics[width=0.49\textwidth]{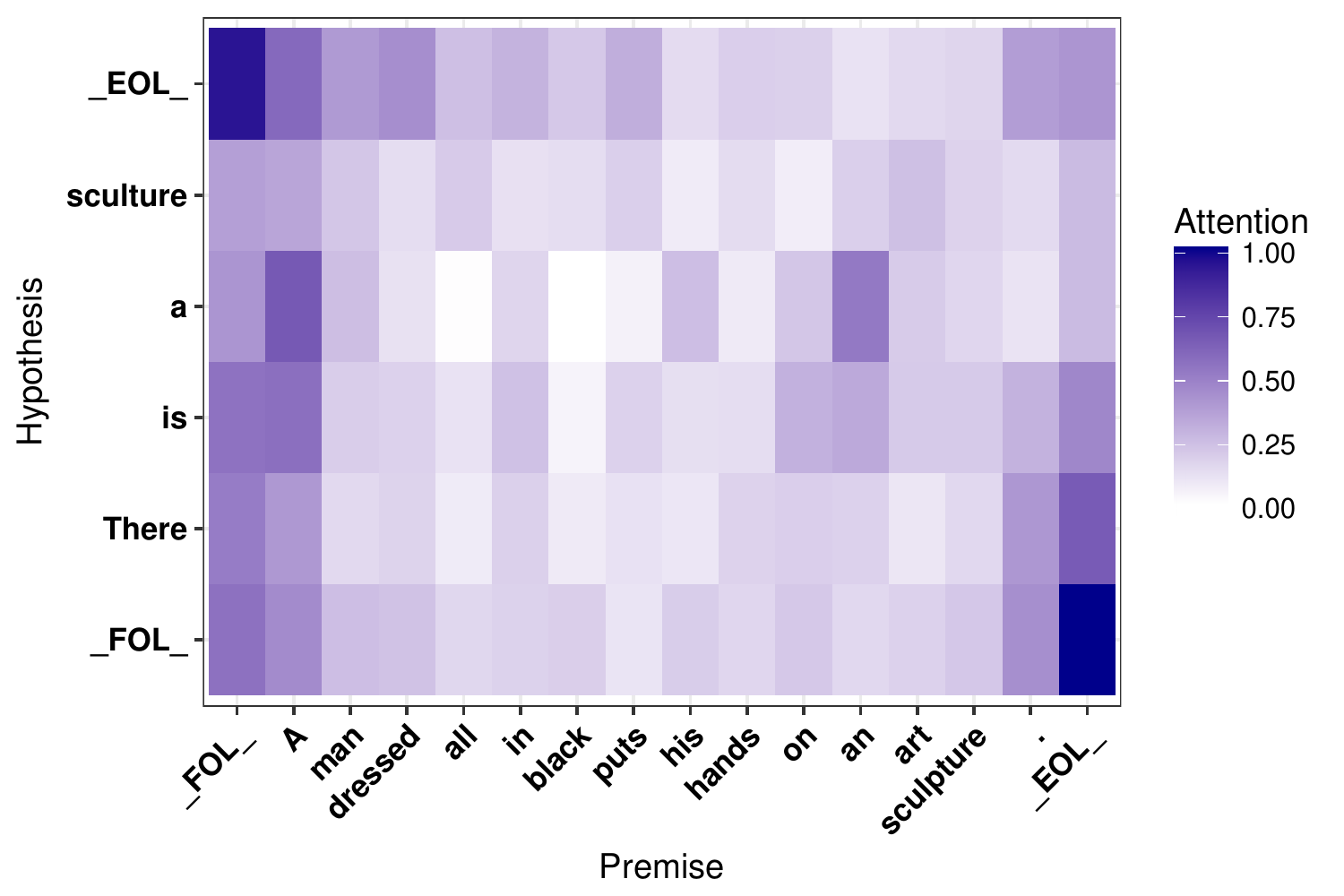}
			}%
			\subfigure[Fixed sample  (sculpture in hypothesis).]{%
				\includegraphics[width=0.49\textwidth]{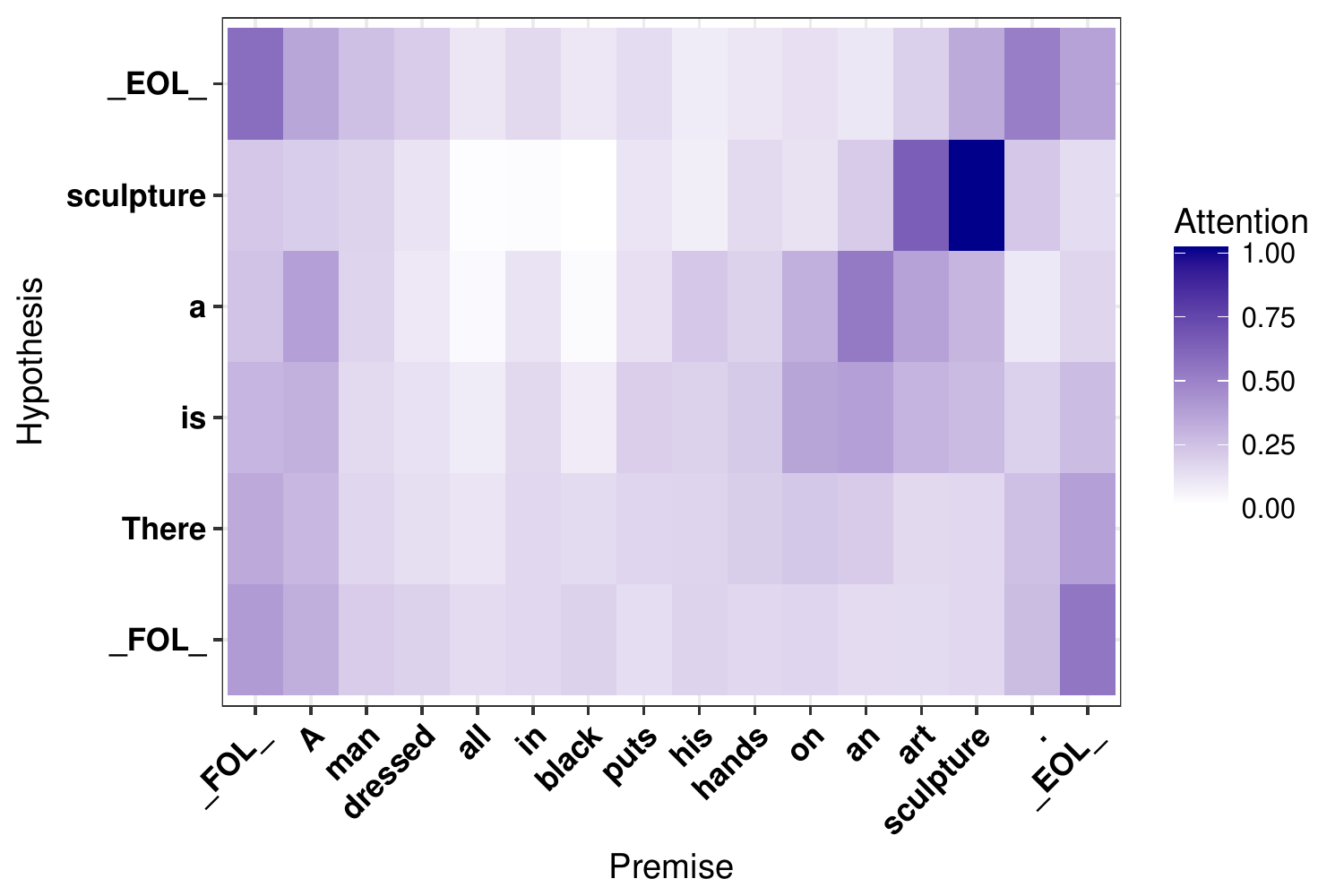}
			}
		\end{center}
		\caption{
			Visualization of the energy function for one erroneous sample (a) and the fixed sample (b). The gold label is \emph{Entailment}. Our model returns \emph{Neutral} for the erroneous sample, but correctly classifies the fixed sample.
		}
		\label{fig:att:miss6}
	\end{figure*}

	\noindent Table~\ref{tab:mis:data} shows some erroneous sentences from the SNLI test set along with their corrected equivalents (after preprocessing). Furthermore, we show the energy function (Equation~\ref{eq:energy} in the paper) visualizations of 6 examples from the aforementioned data samples in Figures~\ref{fig:att:miss1}, \ref{fig:att:miss2}, \ref{fig:att:miss3}, \ref{fig:att:miss4}, \ref{fig:att:miss5}, and \ref{fig:att:miss6}. Each figure presents the visualization of an original erroneous sample along its corrected version. These figures clearly illustrate that fixing the erroneous words leads to producing correct attentions over the sentences. This can be observed by comparing the attention for the erroneous words and corrected words, e.g. ``daschunds'' and ``dachshunds'' in the premise of Figures~\ref{fig:att:miss1} and \ref{fig:att:miss2}. Note that we add two dummy notations (i.e. \_FOL\_, and \_EOL\_) to all sentences which indicate their beginning and end.

	\section{Category Study}
	\label{app:attcat:sec}
	
\begin{figure*}[ht]
		\begin{center}
			\subfigure[Normalized attention of DR-BiLSTM.]{%
				\includegraphics[width=0.49\textwidth]{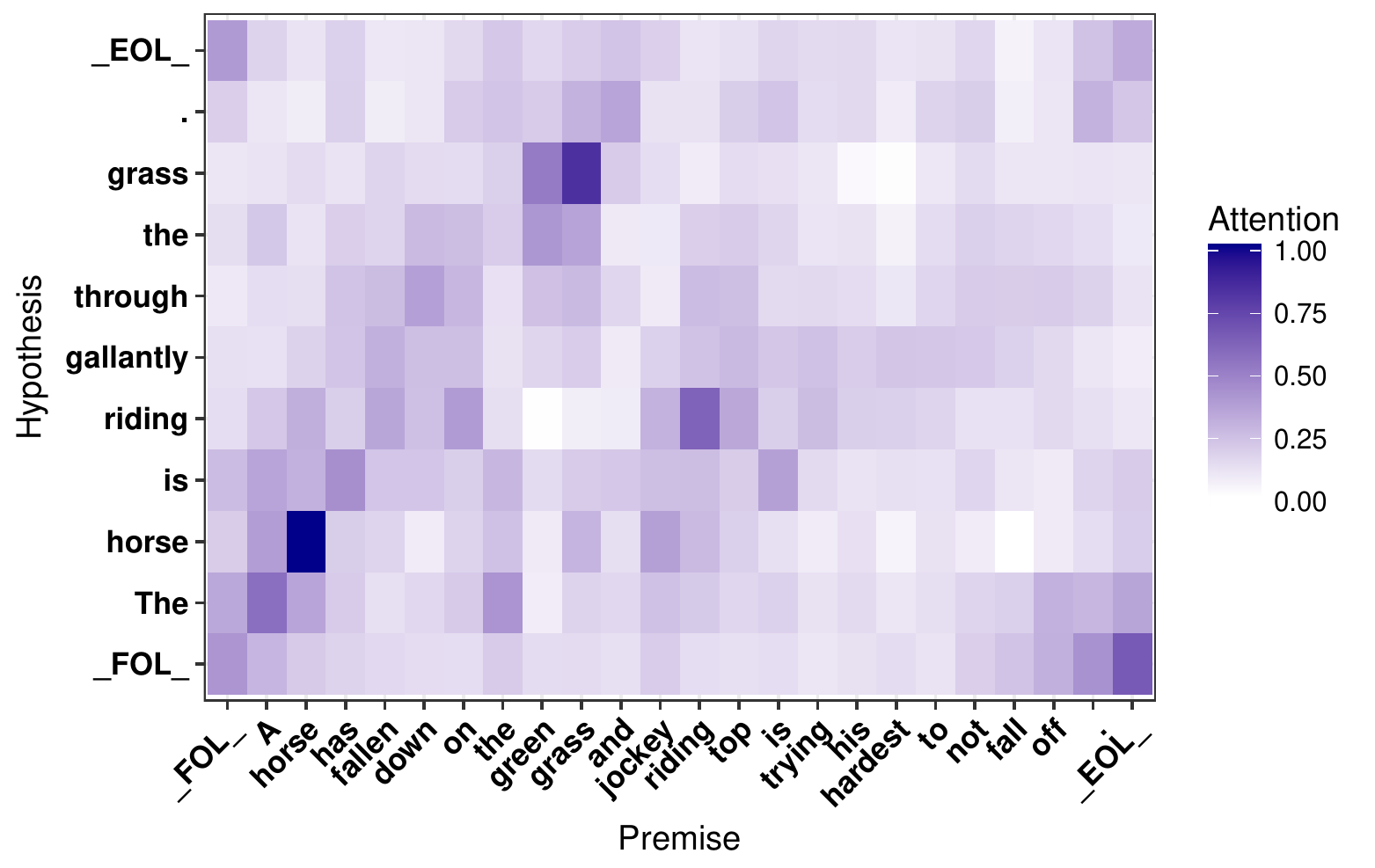}
			}%
			\subfigure[Normalized attention of ESIM]{%
				\includegraphics[width=0.49\textwidth]{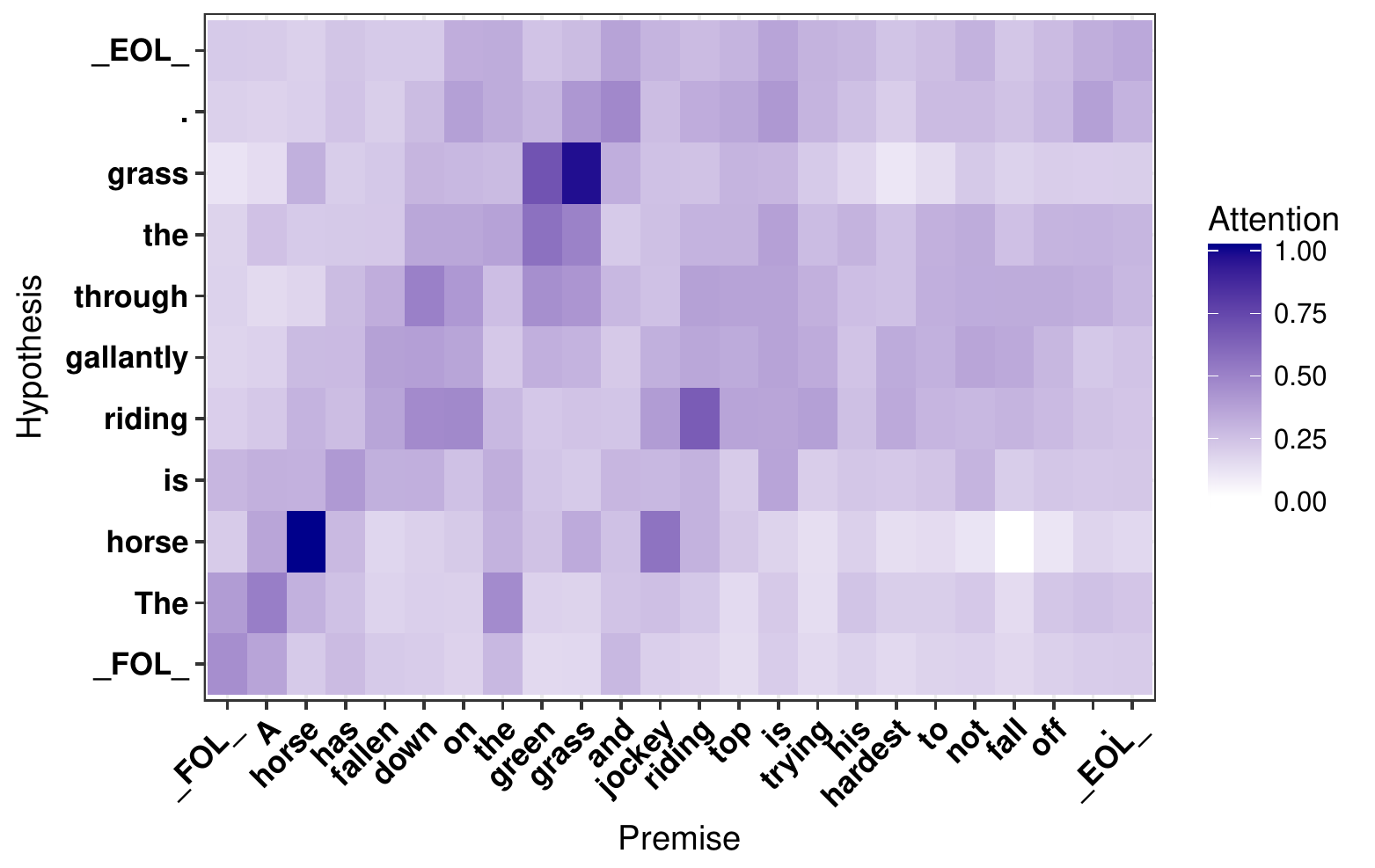}
			}
		\end{center}
		\caption{
			Visualization of the normalized attention weights of DR-BiLSTM (a) and ESIM (b) models for one sample from the SNLI test set. This sample belongs to the Negation category. The gold label is \emph{Contradiction}. Our model returns \emph{Contradiction} while ESIM returns \emph{Entailment}.
		}
		\label{fig:att:ana:cat1}
	\end{figure*}

\begin{figure*}[ht]
		\begin{center}
			\subfigure[Normalized attention of DR-BiLSTM.]{%
				\includegraphics[width=0.49\textwidth]{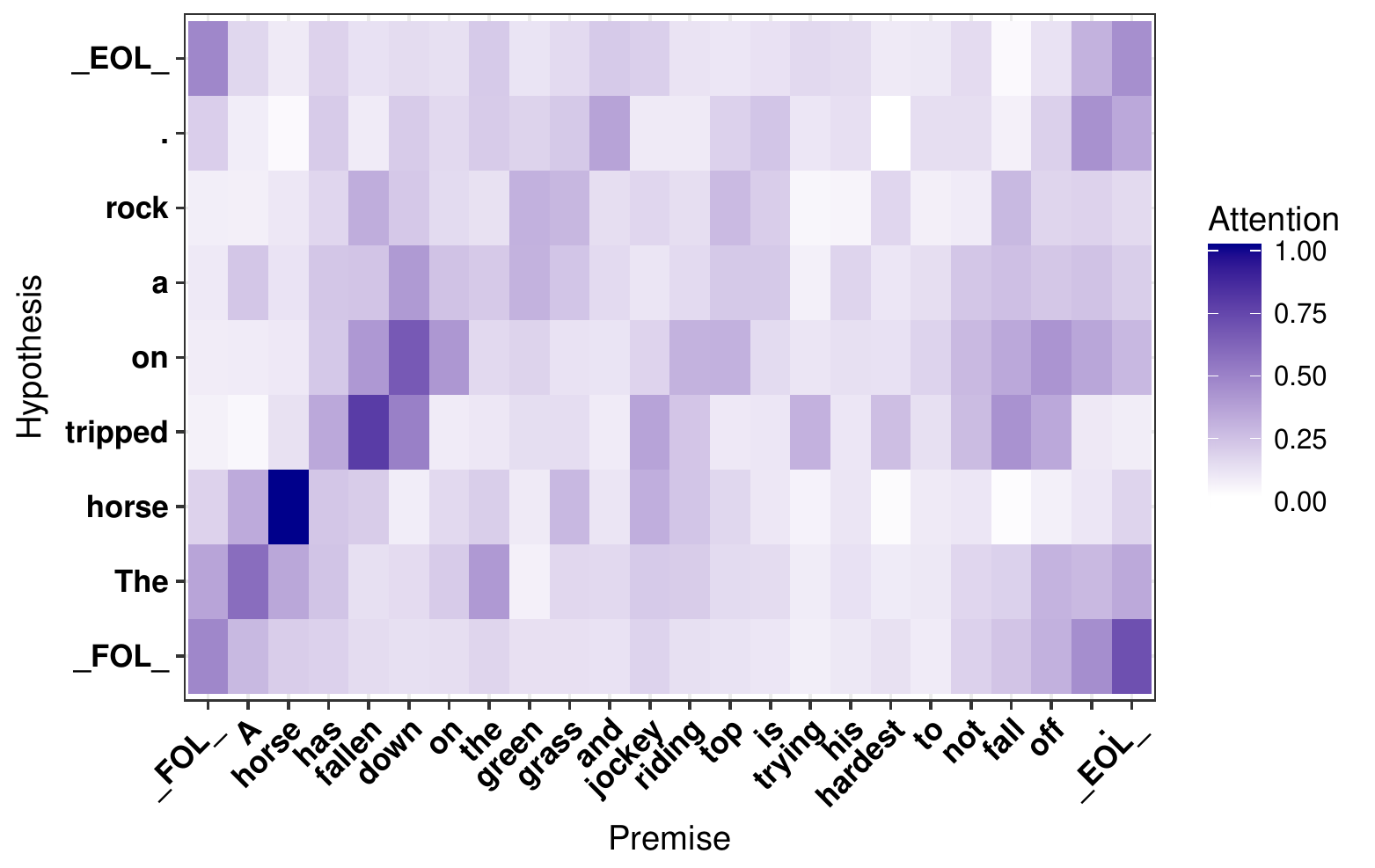}
			}%
			\subfigure[Normalized attention of ESIM]{%
				\includegraphics[width=0.49\textwidth]{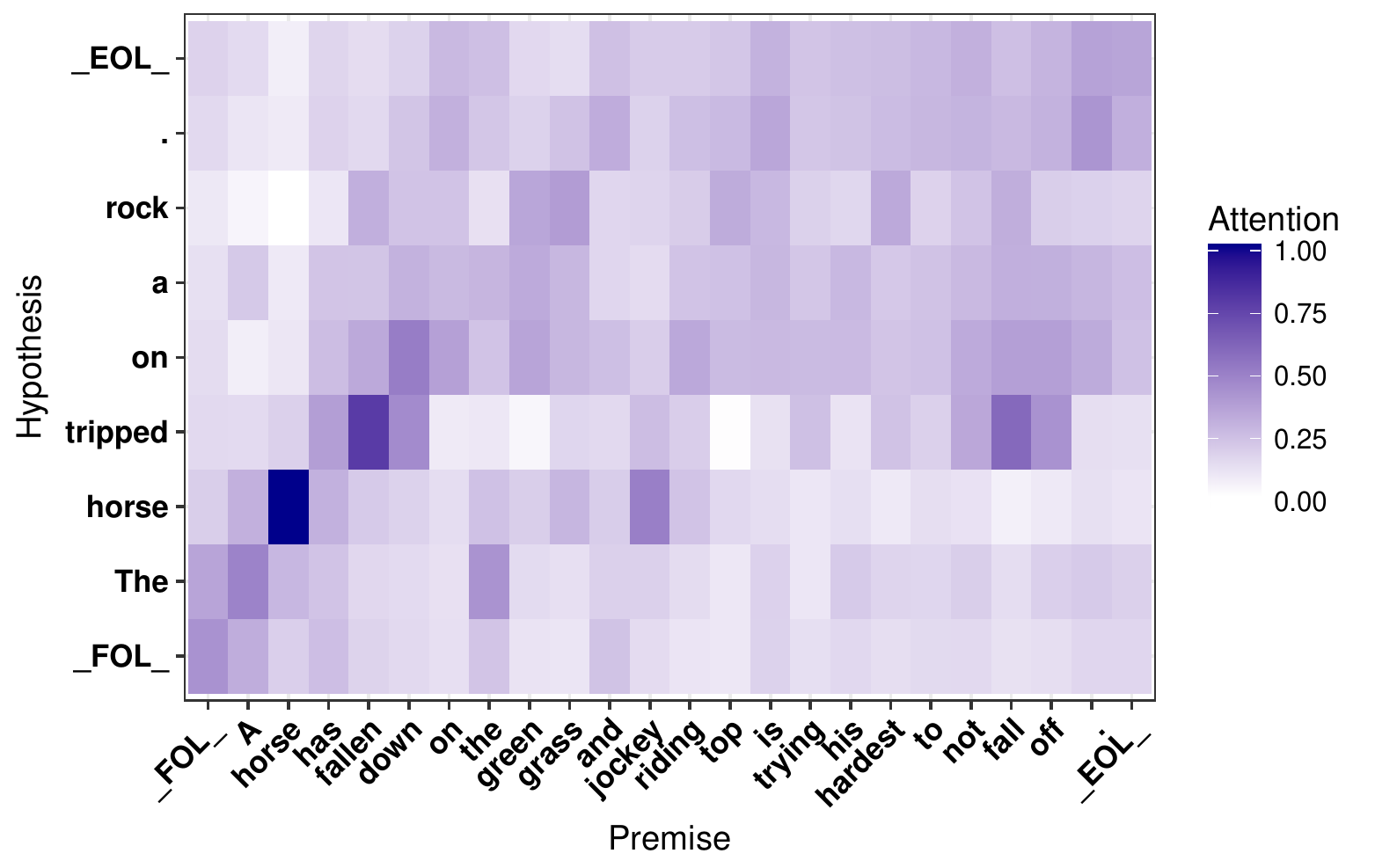}
			}
		\end{center}
		\caption{
			Visualization of the normalized attention weights of DR-BiLSTM (a) and ESIM (b) models for one sample from the SNLI test set. This sample belongs to the Negation category. The gold label is \emph{Contradiction}. Our model returns \emph{Contradiction} while ESIM returns \emph{Entailment}.
		}
		\label{fig:att:ana:cat2}
	\end{figure*}

\begin{figure*}[ht]
		\begin{center}
			\subfigure[Normalized attention of DR-BiLSTM.]{%
				\includegraphics[width=0.49\textwidth]{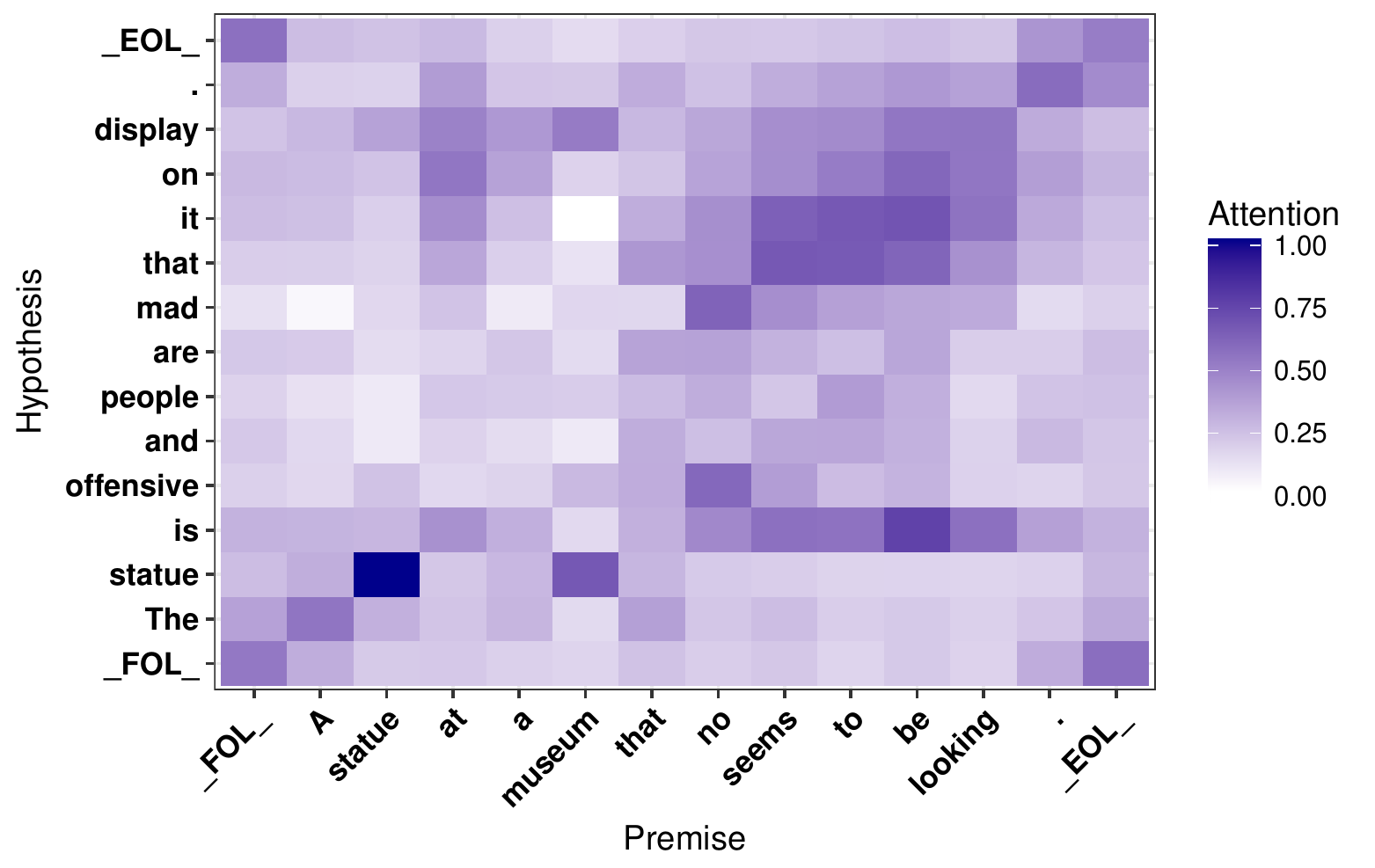}
			}%
			\subfigure[Normalized attention of ESIM]{%
				\includegraphics[width=0.49\textwidth]{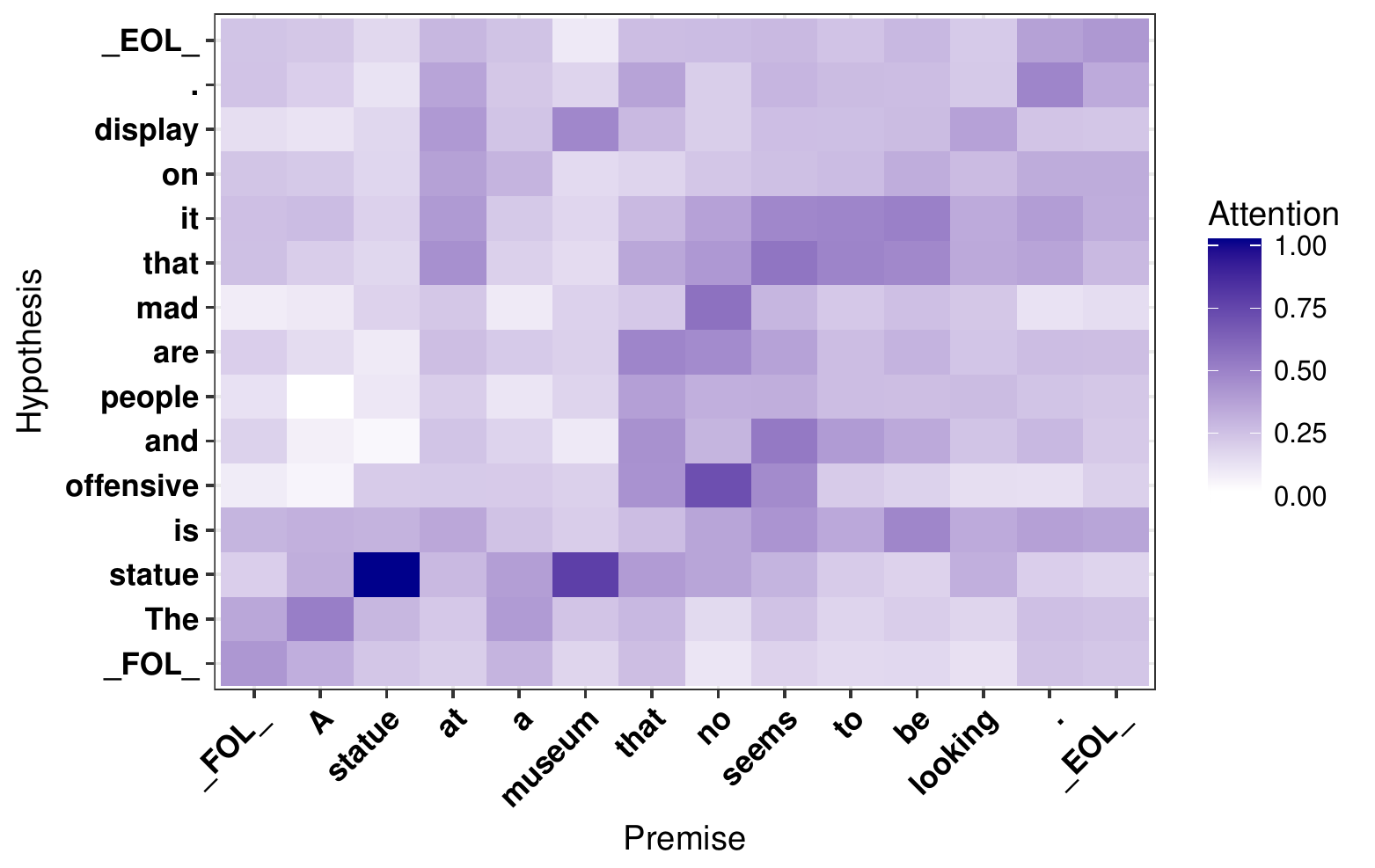}
			}
		\end{center}
		\caption{
			Visualization of the normalized attention weights of DR-BiLSTM (a) and ESIM (b) models for one sample from the SNLI test set. This sample belongs to both Negation and Quantifier categories. The gold label is \emph{Neutral}. Our model returns \emph{Neutral} while ESIM returns \emph{Contradiction}.
		}
		\label{fig:att:ana:cat3}
	\end{figure*}

\begin{figure*}[ht]
		\begin{center}
			\subfigure[Normalized attention of DR-BiLSTM.]{%
				\includegraphics[width=0.49\textwidth]{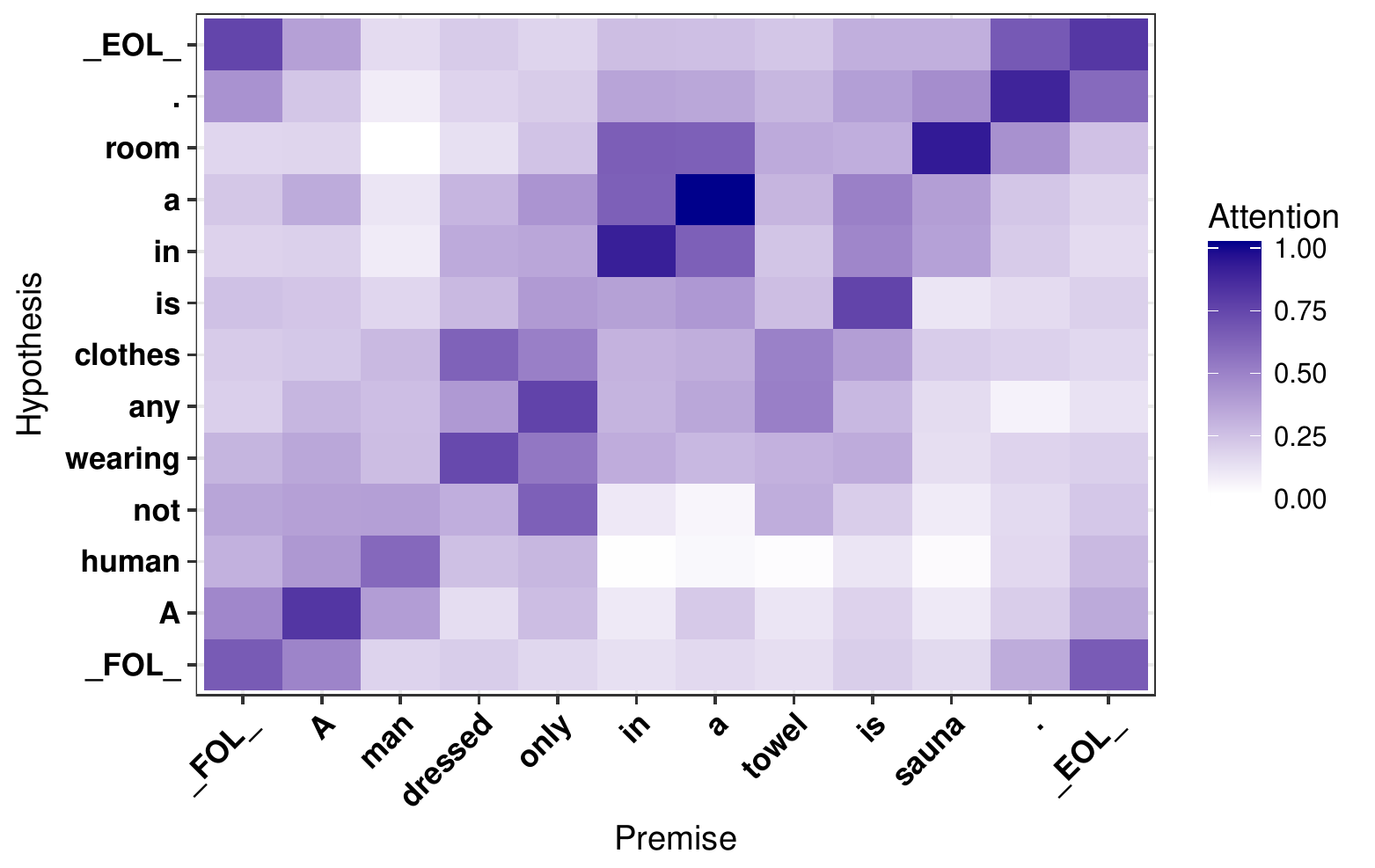}
			}%
			\subfigure[Normalized attention of ESIM]{%
				\includegraphics[width=0.49\textwidth]{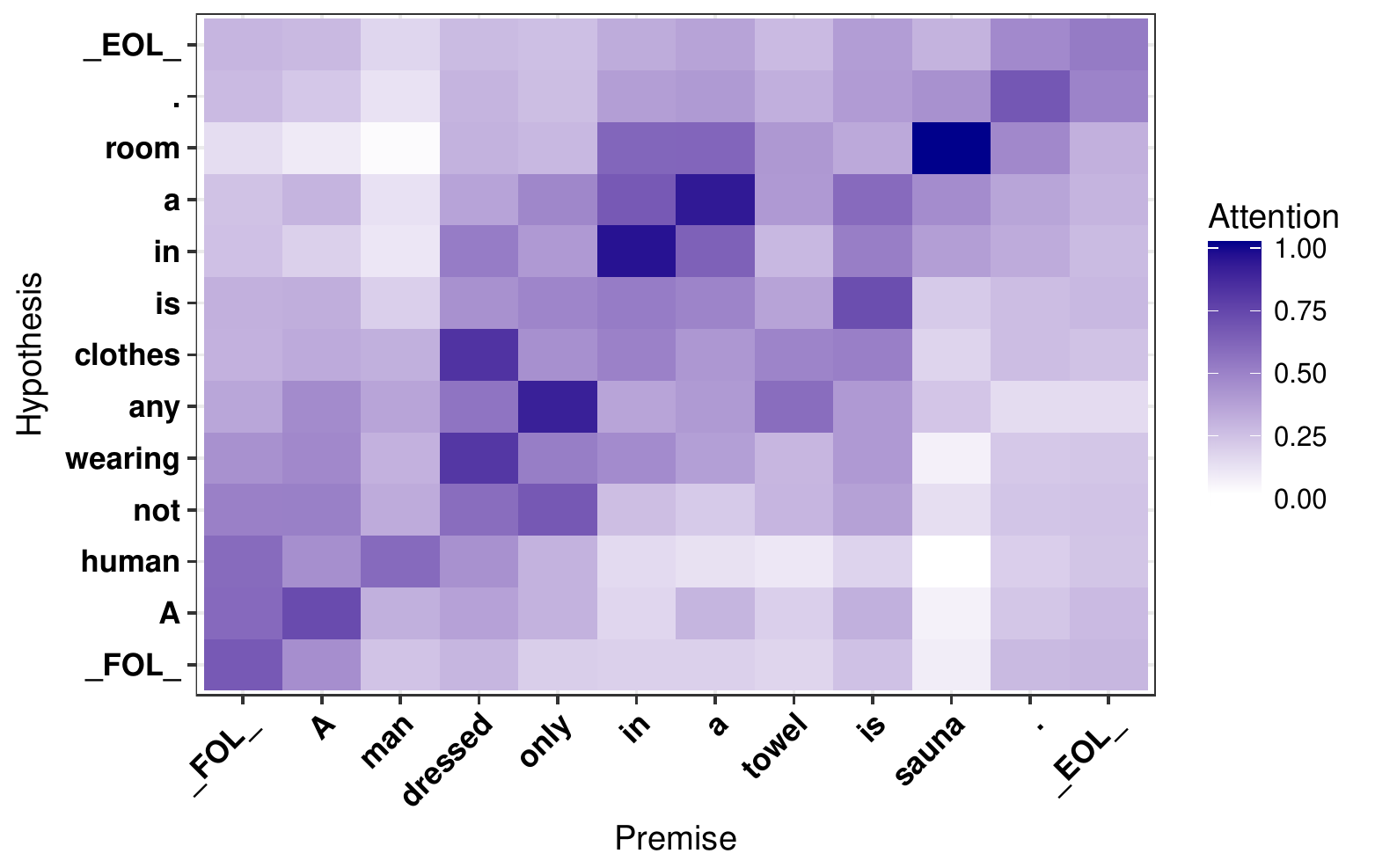}
			}
		\end{center}
		\caption{
			Visualization of the normalized attention weights of DR-BiLSTM (a) and ESIM (b) models for one sample from the SNLI test set. This sample belongs to both Negation and Quantifier categories. The gold label is \emph{Entailment}. Our model returns \emph{Entailment} while ESIM returns \emph{Contradiction}.
		}
		\label{fig:att:ana:cat4}
	\end{figure*}

	Here we investigate the normalized attention weights of DR-BiLSTM and ESIM for four samples that belong to Negation and/or Quantifier categories (Figures~\ref{fig:att:ana:cat1} - \ref{fig:att:ana:cat4}). Each figure illustrates the normalized energy function of DR-BiLSTM (left diagram) and ESIM (right diagram) respectively. Provided figures indicate that ESIM assigns somewhat similar attention to most of the pairs while DR-BiLSTM focuses on specific parts of the given premise and hypothesis.
	
	\section{Attention Study}
	\label{app:att:sec}
	
\begin{figure*}[ht]
		\begin{center}
			\subfigure[Instance 1 - Entailment relationship.]{%
				\label{fig:sub:1:ent}
				\includegraphics[width=0.49\textwidth]{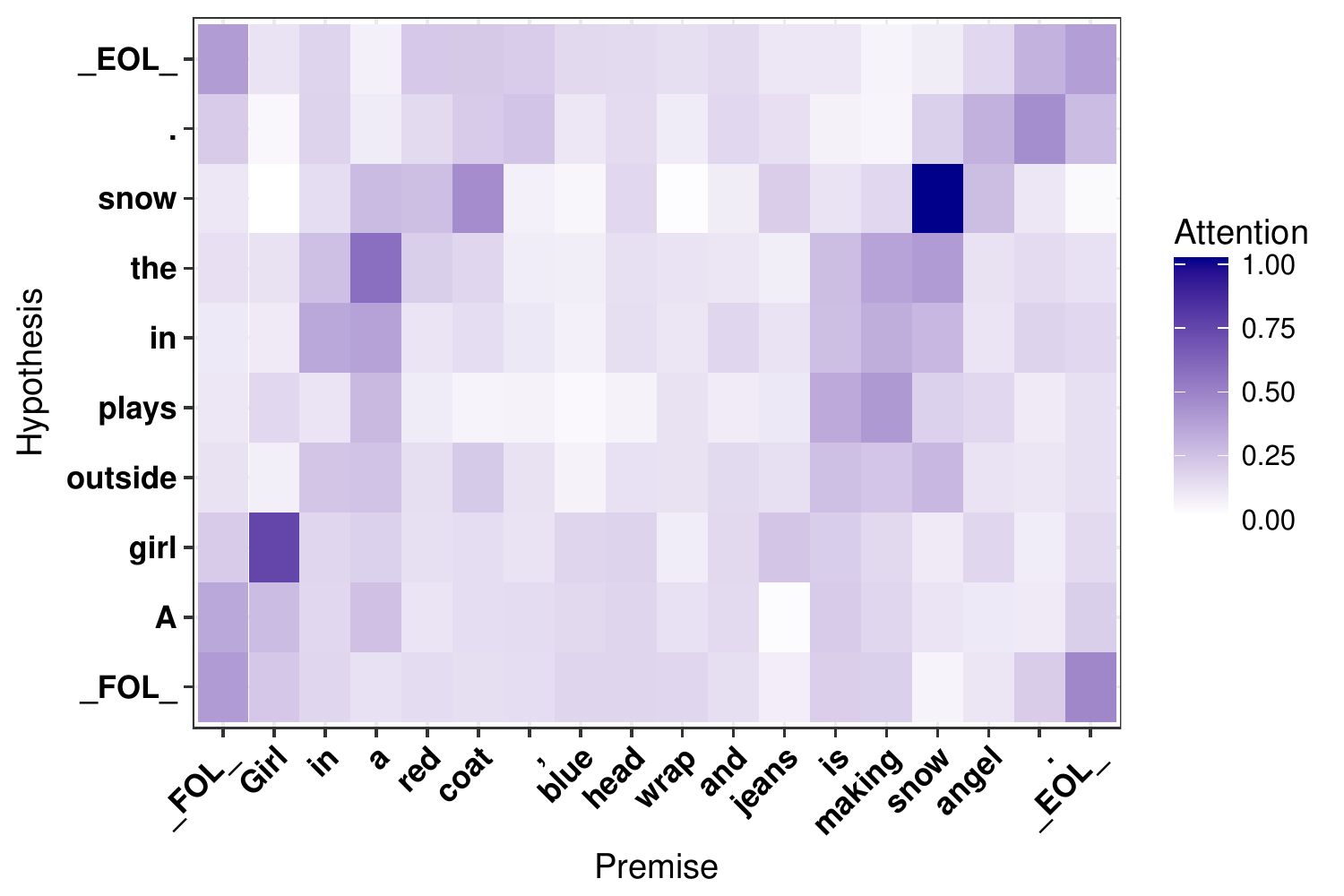}
			}%
			\subfigure[Instance 2 - Entailment relationship.]{%
				\label{fig:sub:2:ent}
				\includegraphics[width=0.49\textwidth]{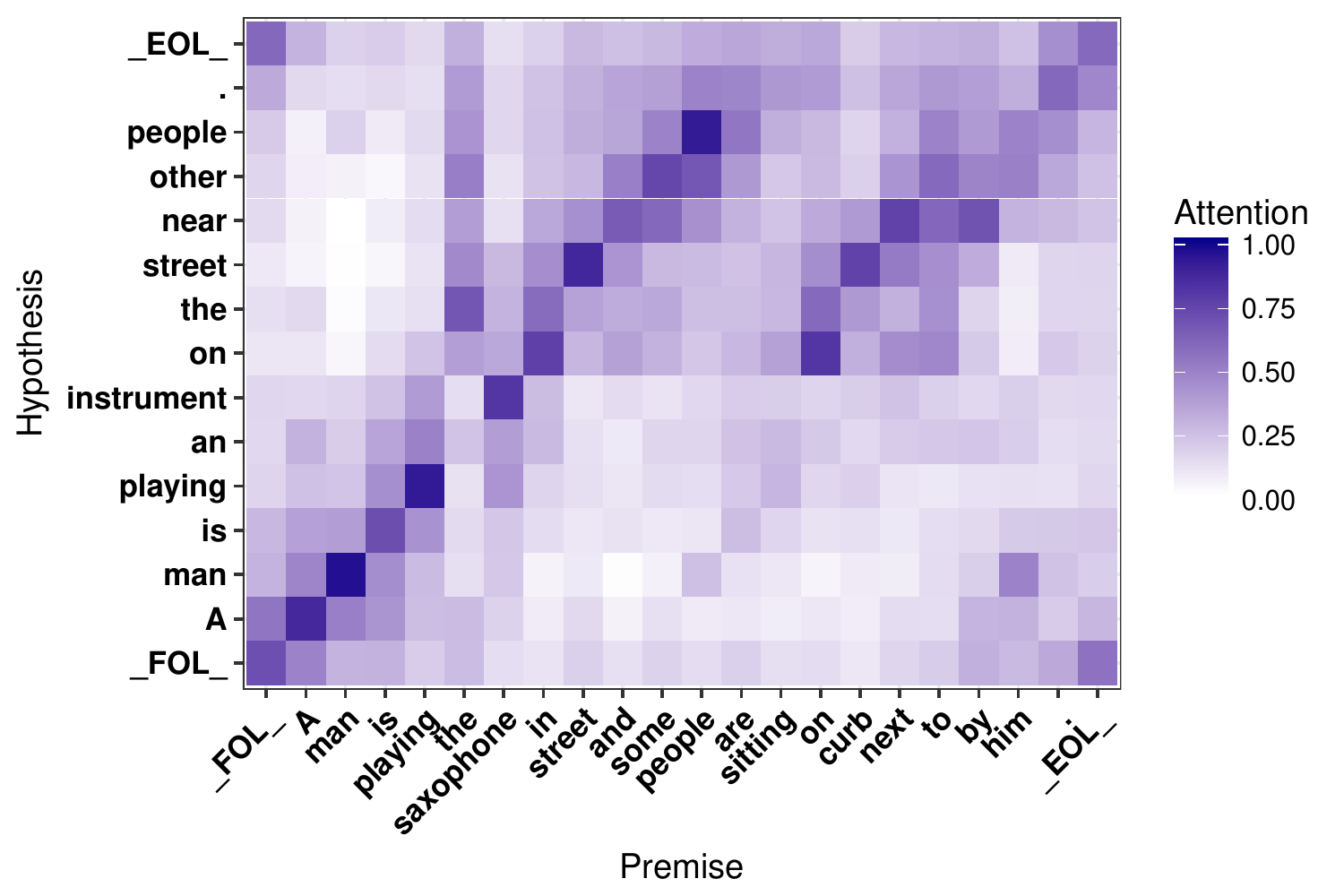}
			}\\ 
			\subfigure[Instance 1 - Neutral relationship.]{%
				\label{fig:sub:1:net}
				\includegraphics[width=0.49\textwidth]{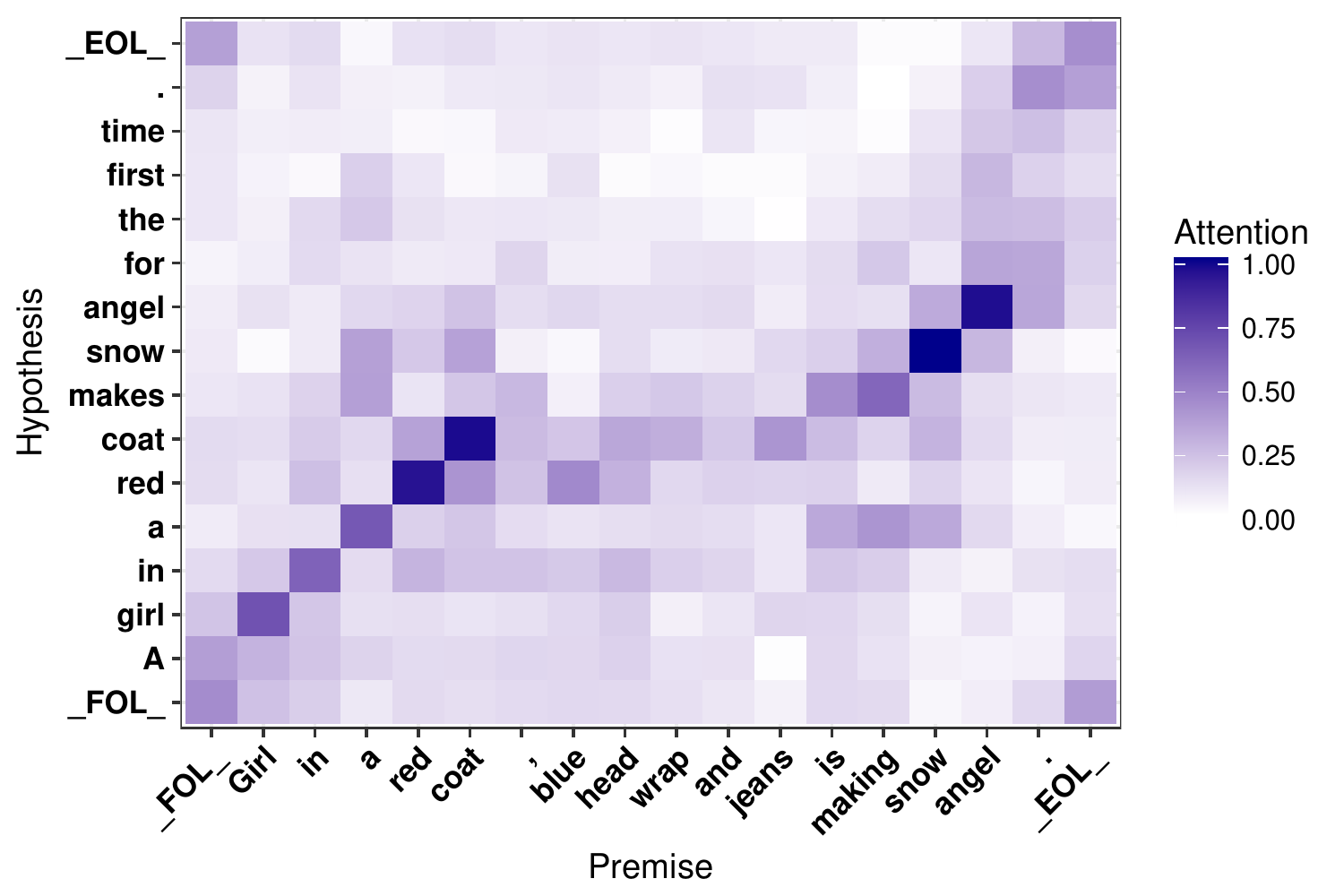}
			}%
			\subfigure[Instance 2 - Neutral relationship.]{%
				\label{fig:sub:2:net}
				\includegraphics[width=0.49\textwidth]{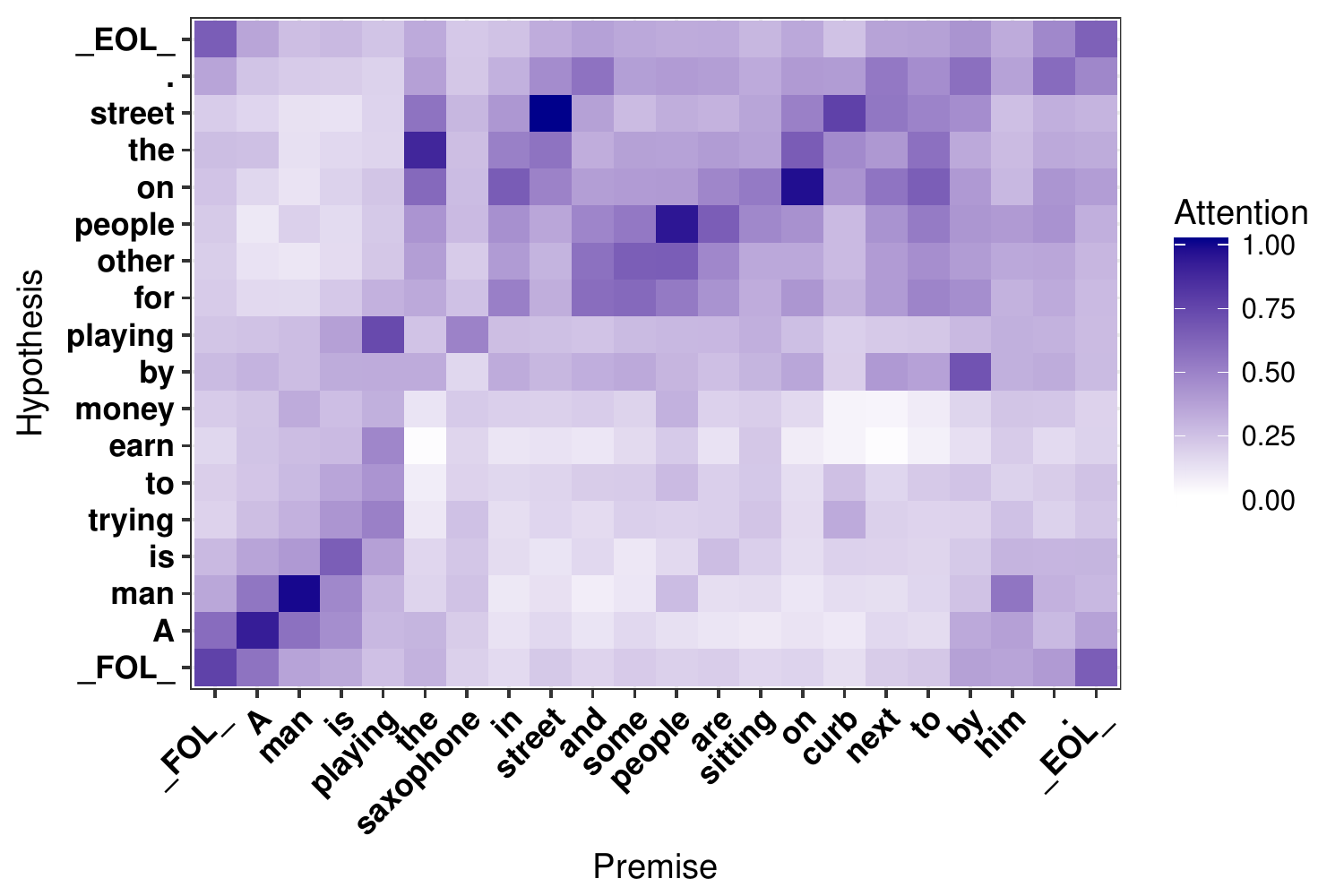}
			}\\%
			\subfigure[Instance 1 - Contradiction relationship.]{%
				\label{fig:sub:1:cnt}
				\includegraphics[width=0.49\textwidth]{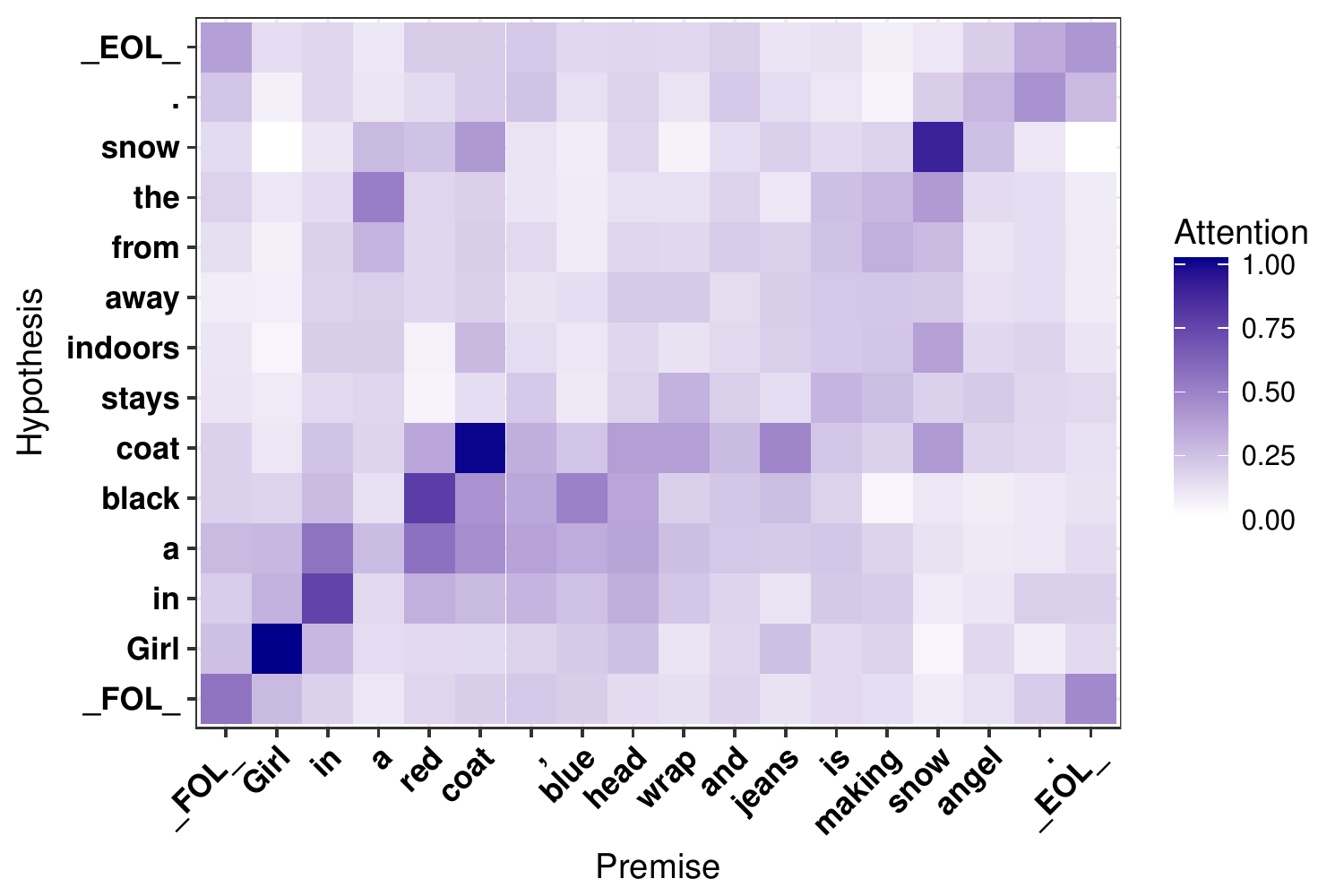}
			}%
			\subfigure[Instance 2 - Contradiction relationship.]{%
				\label{fig:sub:2:cnt}
				\includegraphics[width=0.49\textwidth]{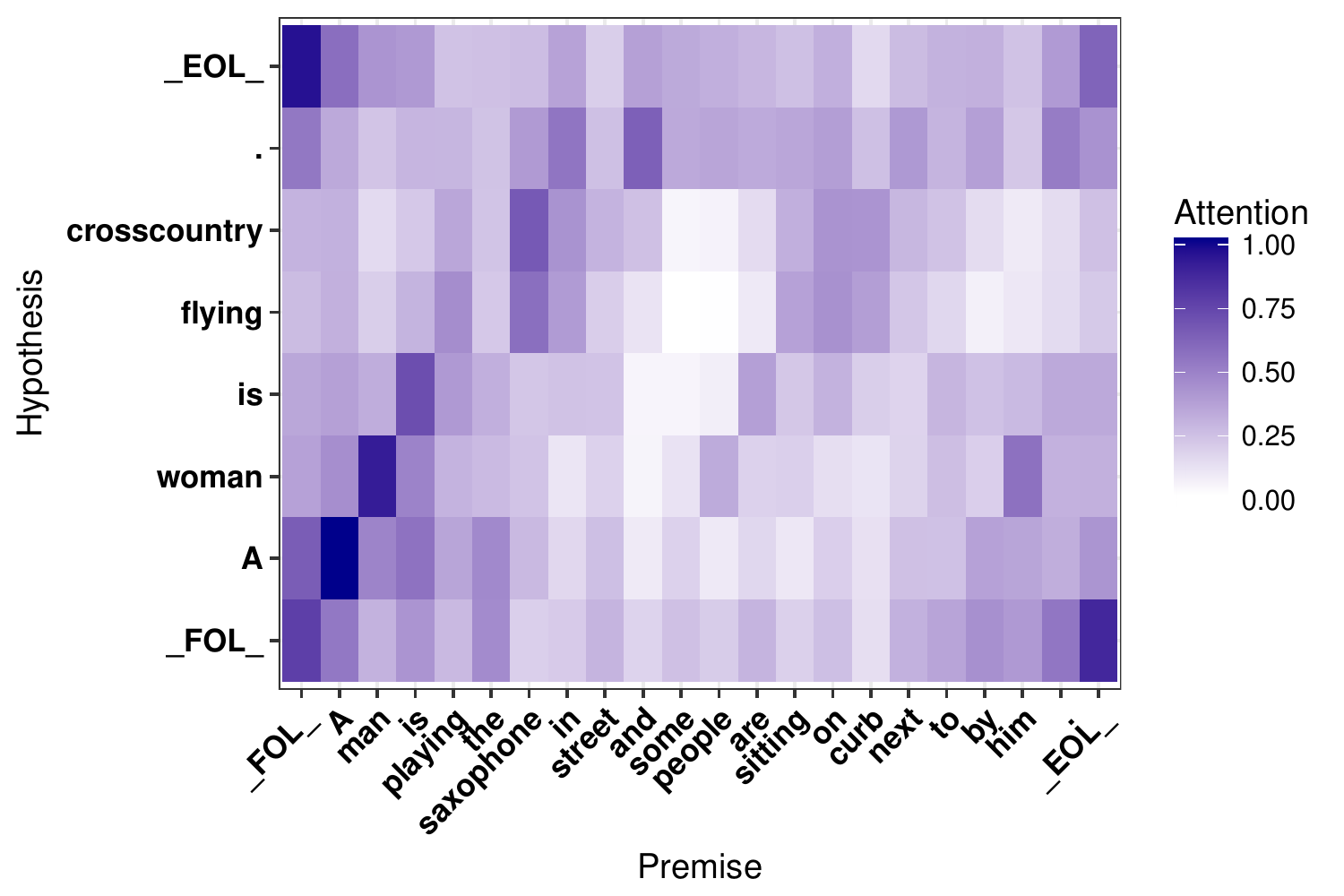}
			}%
		\end{center}
		\caption{
			Normalized attention weights for 6 data samples from the test set of SNLI dataset. (a,c,e) and (b,d,f) represent the normalized attention weights for \emph{Entailment}, \emph{Neutral}, and \emph{Contradiction} logical relationships of two premises (Instance 1 and 2) respectively. Darker color illustrates higher attention. 
		}
		\label{fig:att:sample:1}
	\end{figure*}

\begin{figure*}[ht]
		\begin{center}
			\subfigure[Instance 3 - Entailment relationship.]{%
				\label{fig:sub:3:ent}
				\includegraphics[width=0.49\textwidth]{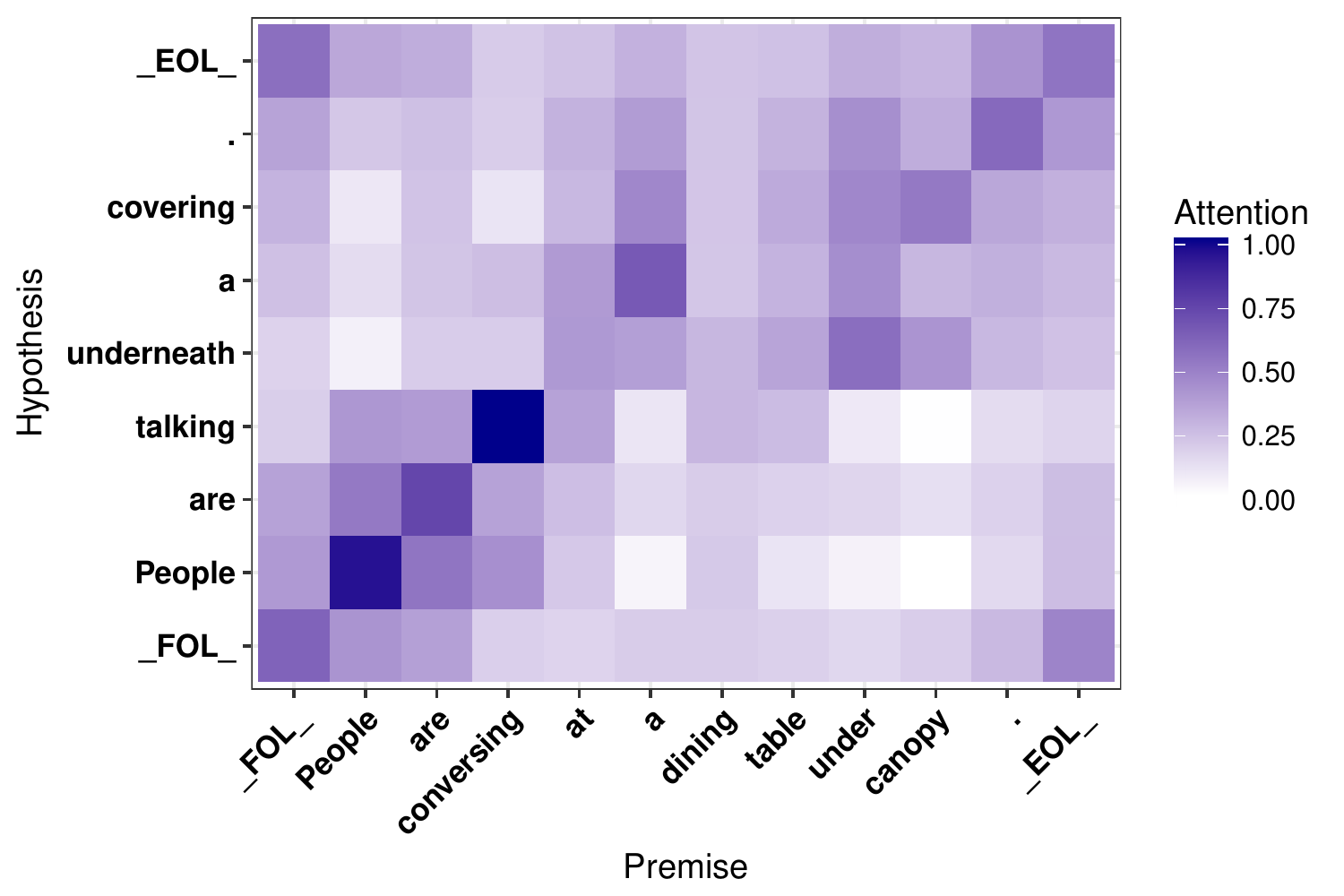}
			}%
			\subfigure[Instance 4 - Entailment relationship.]{%
				\label{fig:sub:4:ent}
				\includegraphics[width=0.49\textwidth]{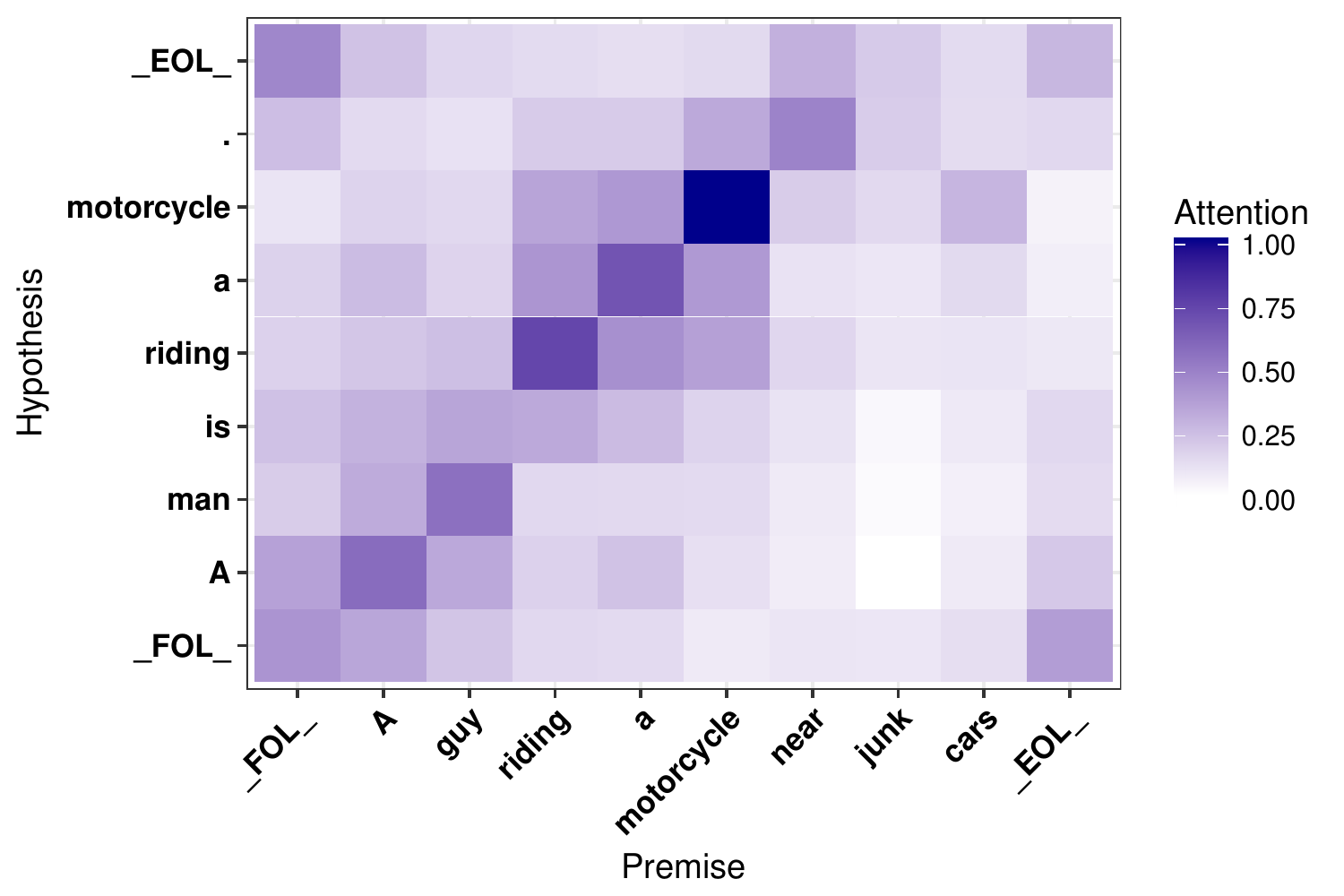}
			}\\ 
			\subfigure[Instance 3 - Neutral relationship.]{%
				\label{fig:sub:3:net}
				\includegraphics[width=0.49\textwidth]{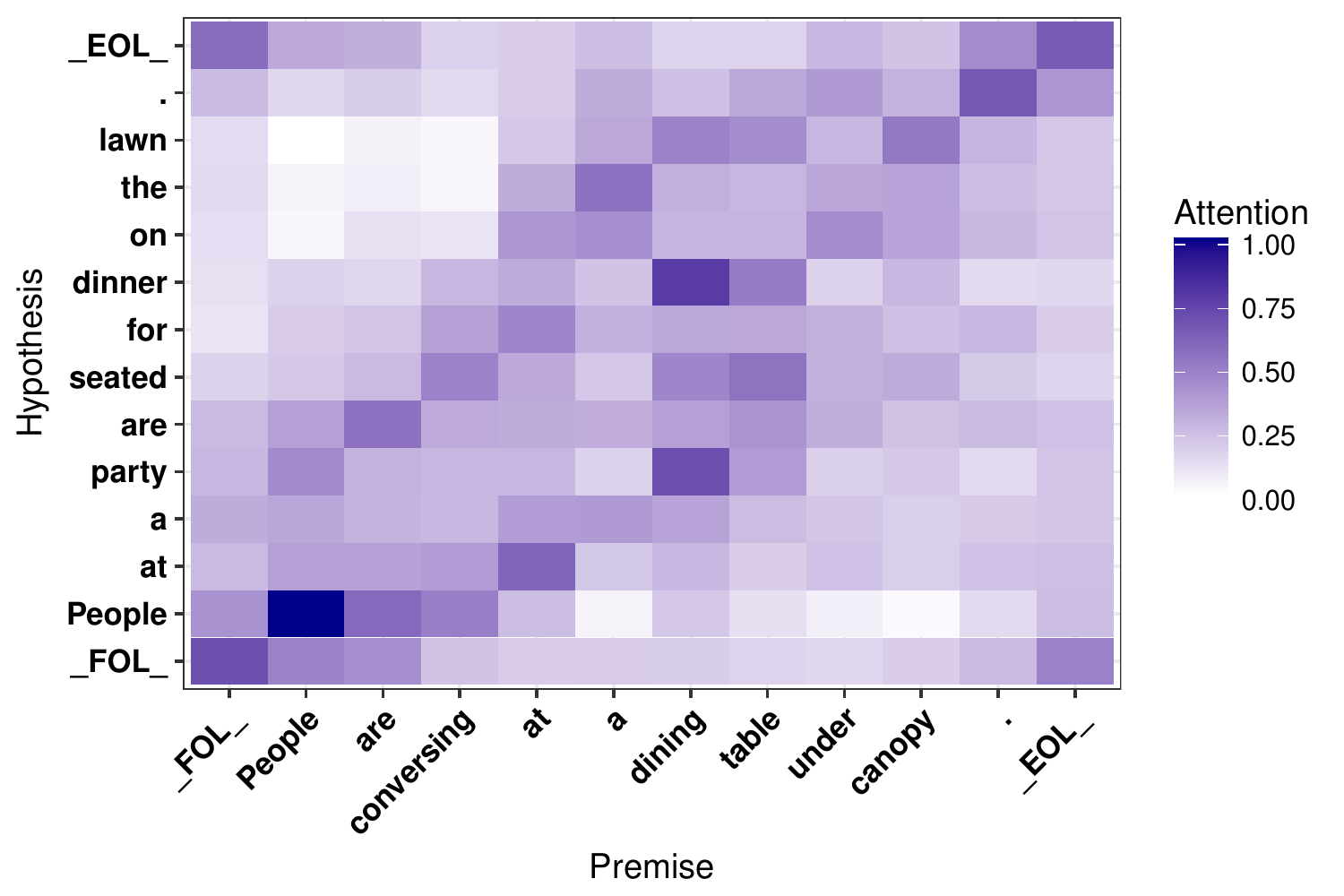}
			}%
			\subfigure[Instance 4 - Neutral relationship.]{%
				\label{fig:sub:4:net}
				\includegraphics[width=0.49\textwidth]{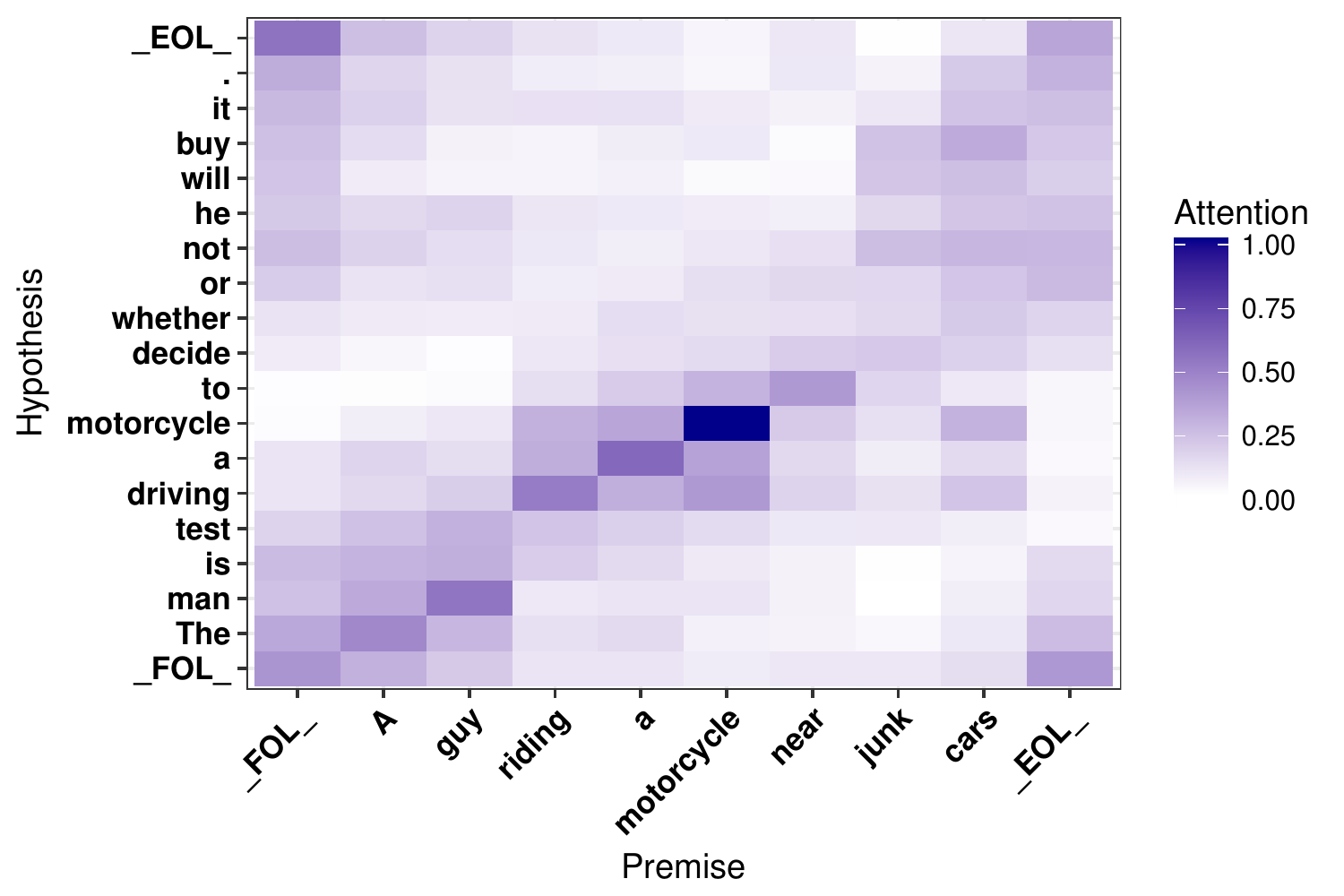}
			}\\%
			\subfigure[Instance 3 - Contradiction relationship.]{%
				\label{fig:sub:3:cnt}
				\includegraphics[width=0.49\textwidth]{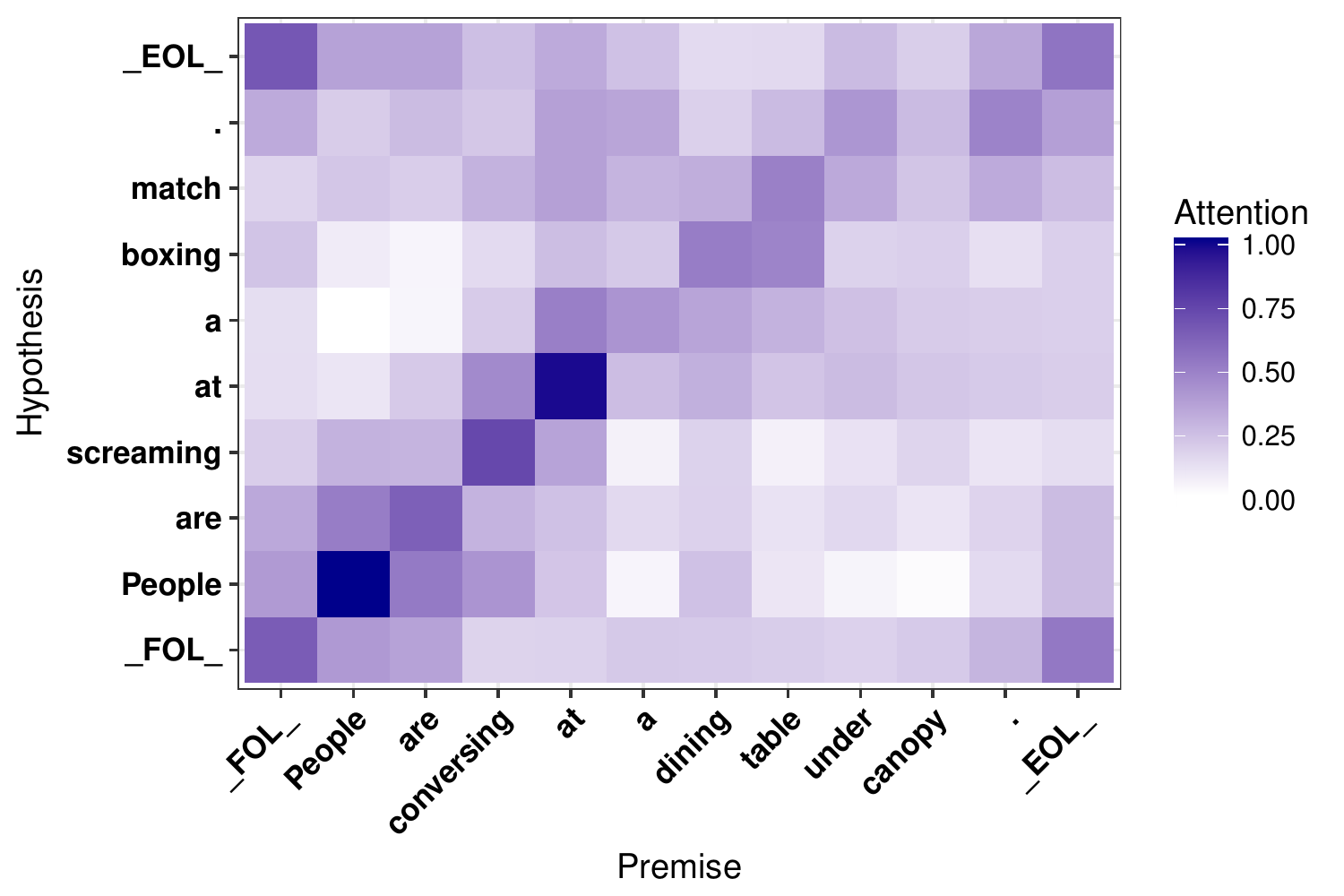}
			}%
			\subfigure[Instance 4 - Contradiction relationship.]{%
				\label{fig:sub:4:cnt}
				\includegraphics[width=0.49\textwidth]{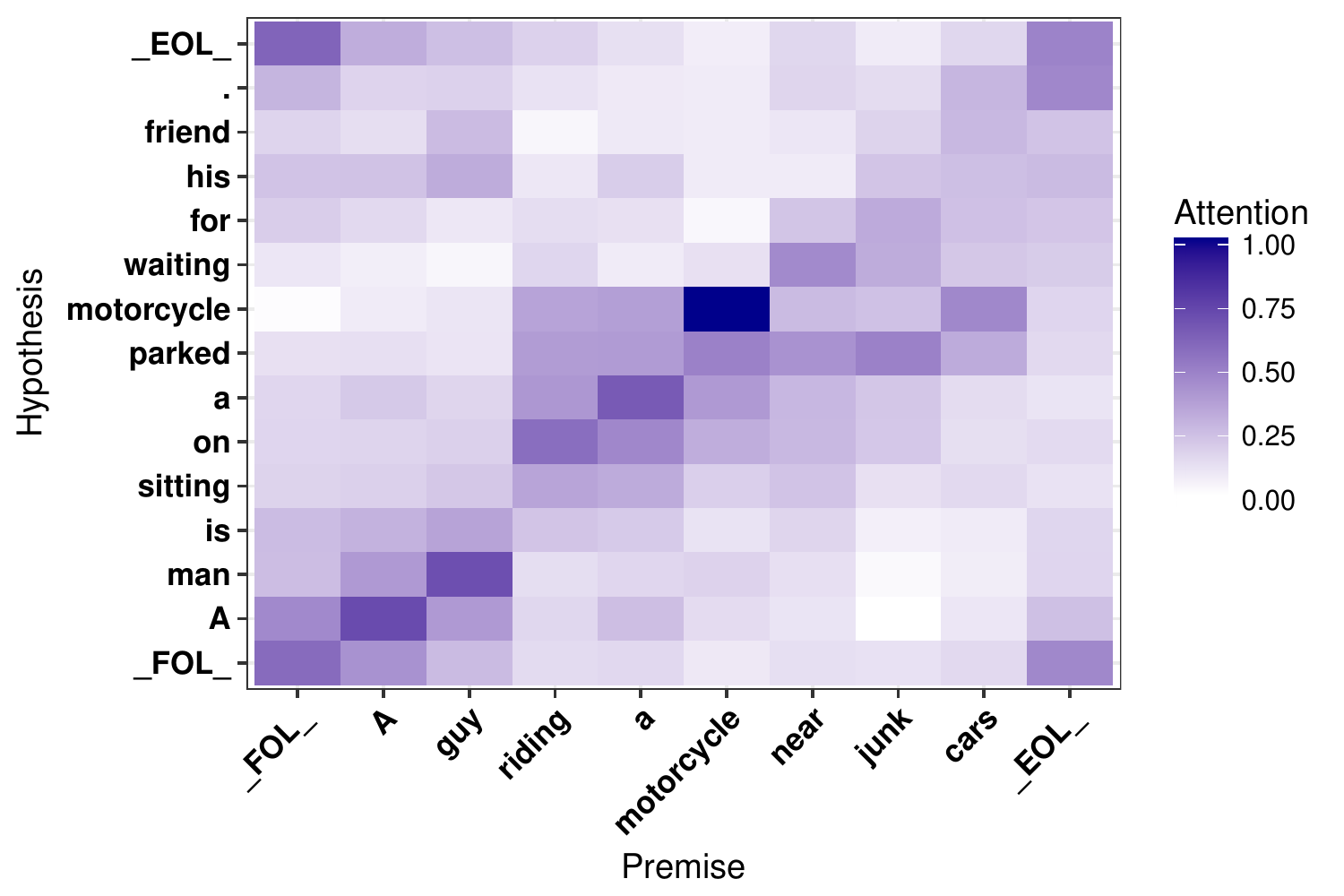}
			}%
		\end{center}
		\caption{
			Normalized attention weights for 6 data samples from the test set of SNLI dataset. (a,c,e) and (b,d,f) represent the normalized attention weights for \emph{Entailment}, \emph{Neutral}, and \emph{Contradiction} logical relationships of two premises (Instance 3 and 4) respectively. Darker color illustrates higher attention. 
		}
		\label{fig:att:sample:2}
	\end{figure*}

\begin{figure*}[ht]
		\begin{center}
			\subfigure[Instance 5 - Entailment relationship.]{%
				\label{fig:sub:5:ent}
				\includegraphics[width=0.49\textwidth]{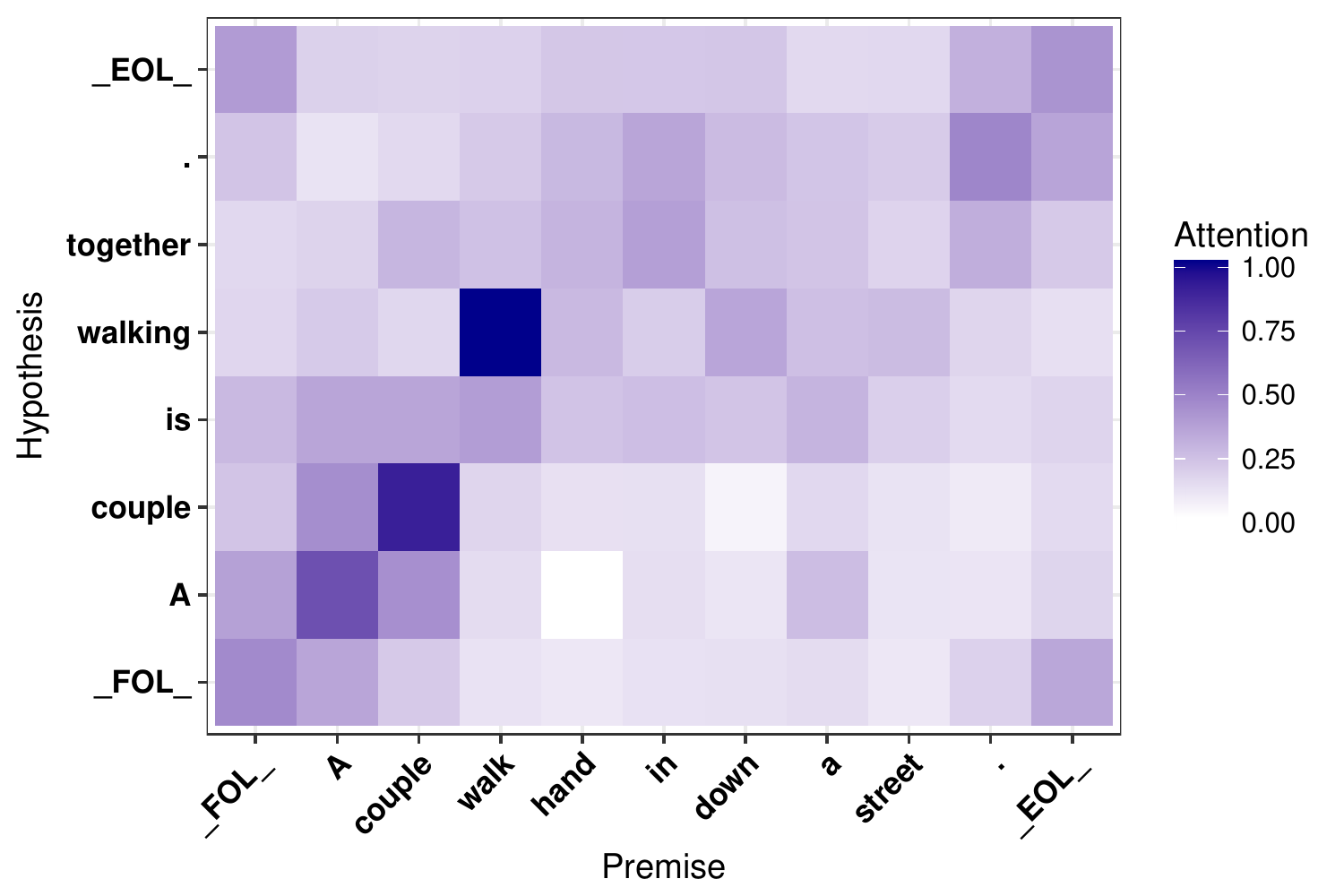}
			}%
			\subfigure[Instance 6 - Entailment relationship.]{%
				\label{fig:sub:6:ent}
				\includegraphics[width=0.49\textwidth]{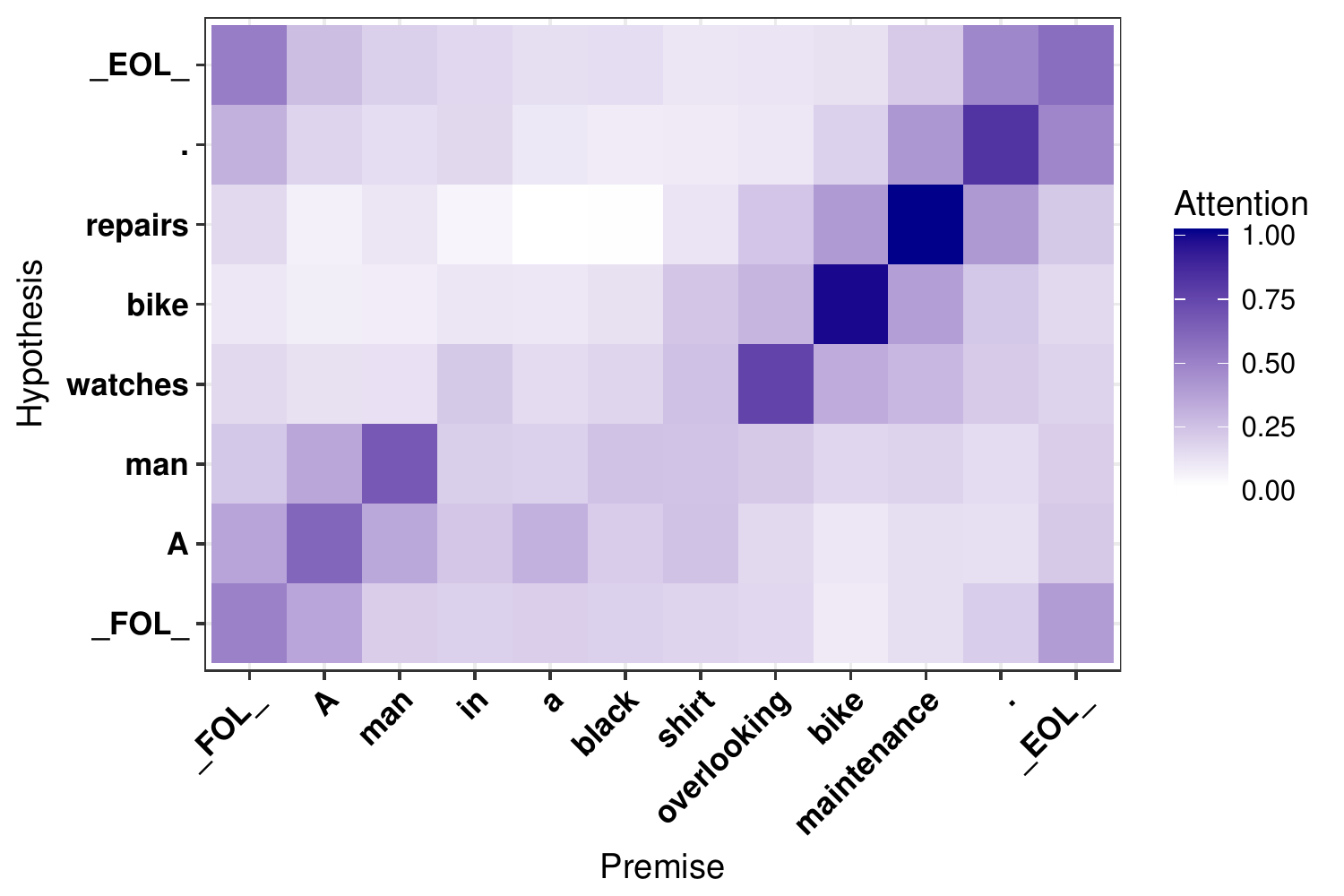}
			}\\ 
			\subfigure[Instance 5 - Neutral relationship.]{%
				\label{fig:sub:5:net}
				\includegraphics[width=0.49\textwidth]{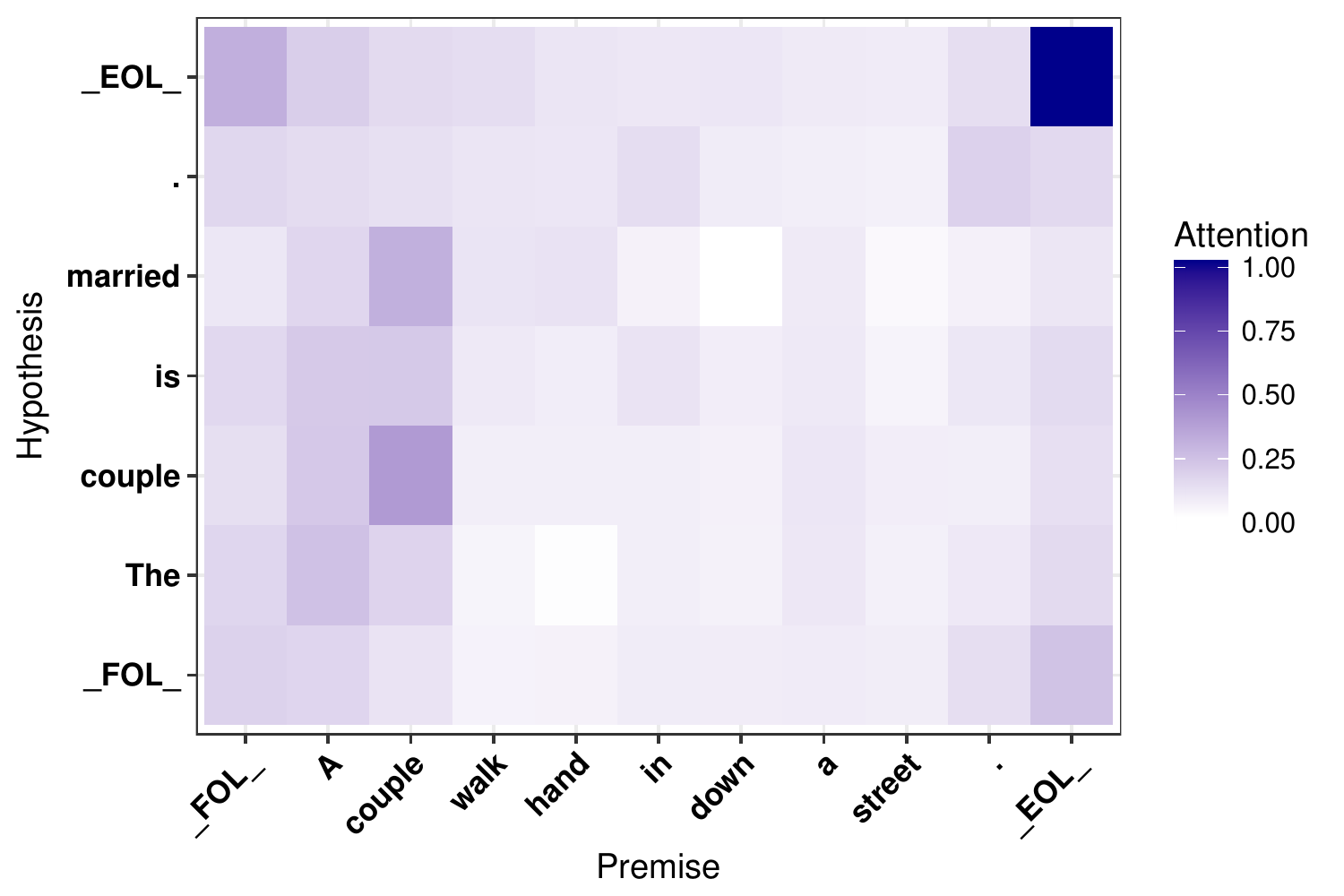}
			}%
			\subfigure[Instance 6 - Neutral relationship.]{%
				\label{fig:sub:6:net}
				\includegraphics[width=0.49\textwidth]{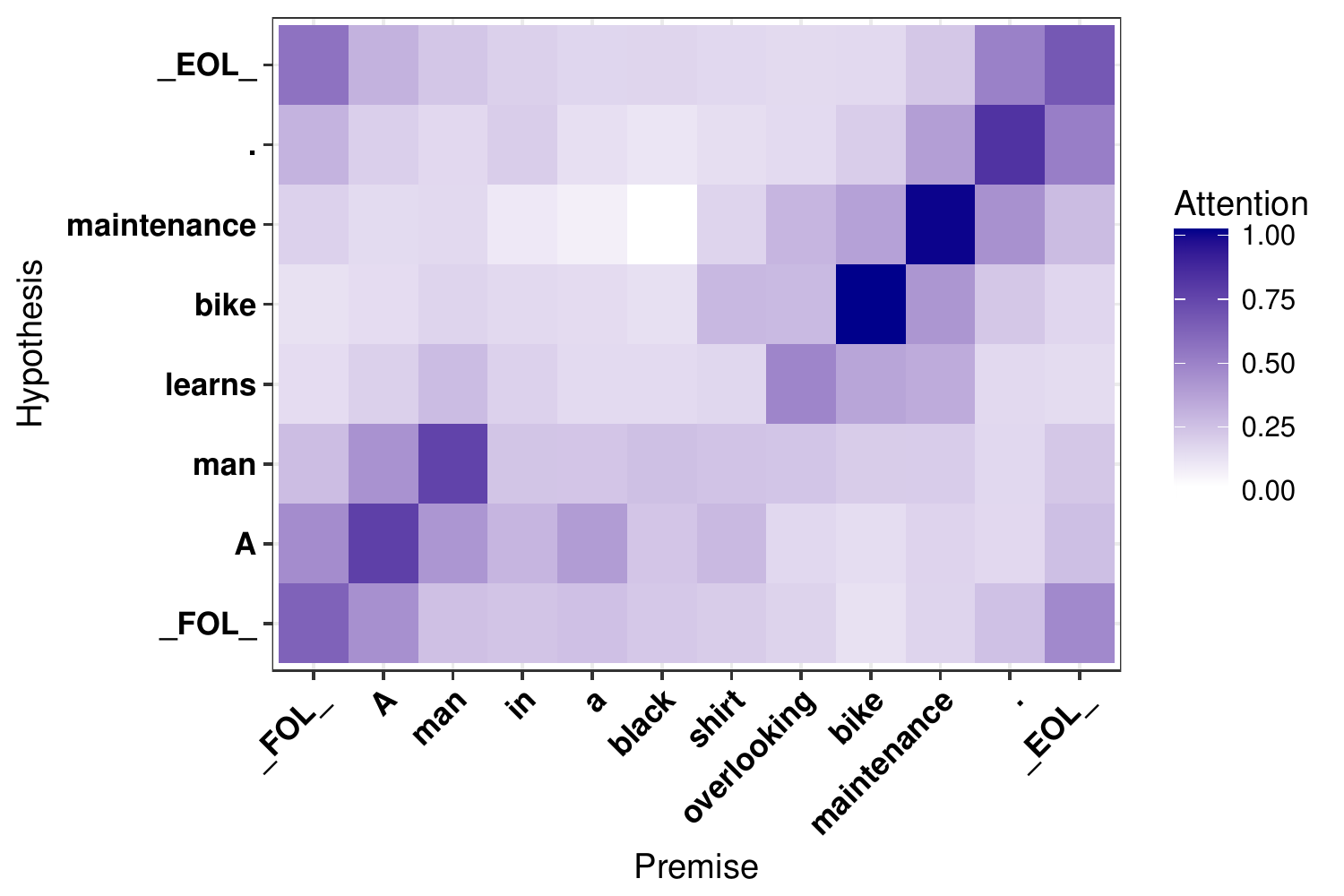}
			}\\%
			\subfigure[Instance 5 - Contradiction relationship.]{%
				\label{fig:sub:5:cnt}
				\includegraphics[width=0.49\textwidth]{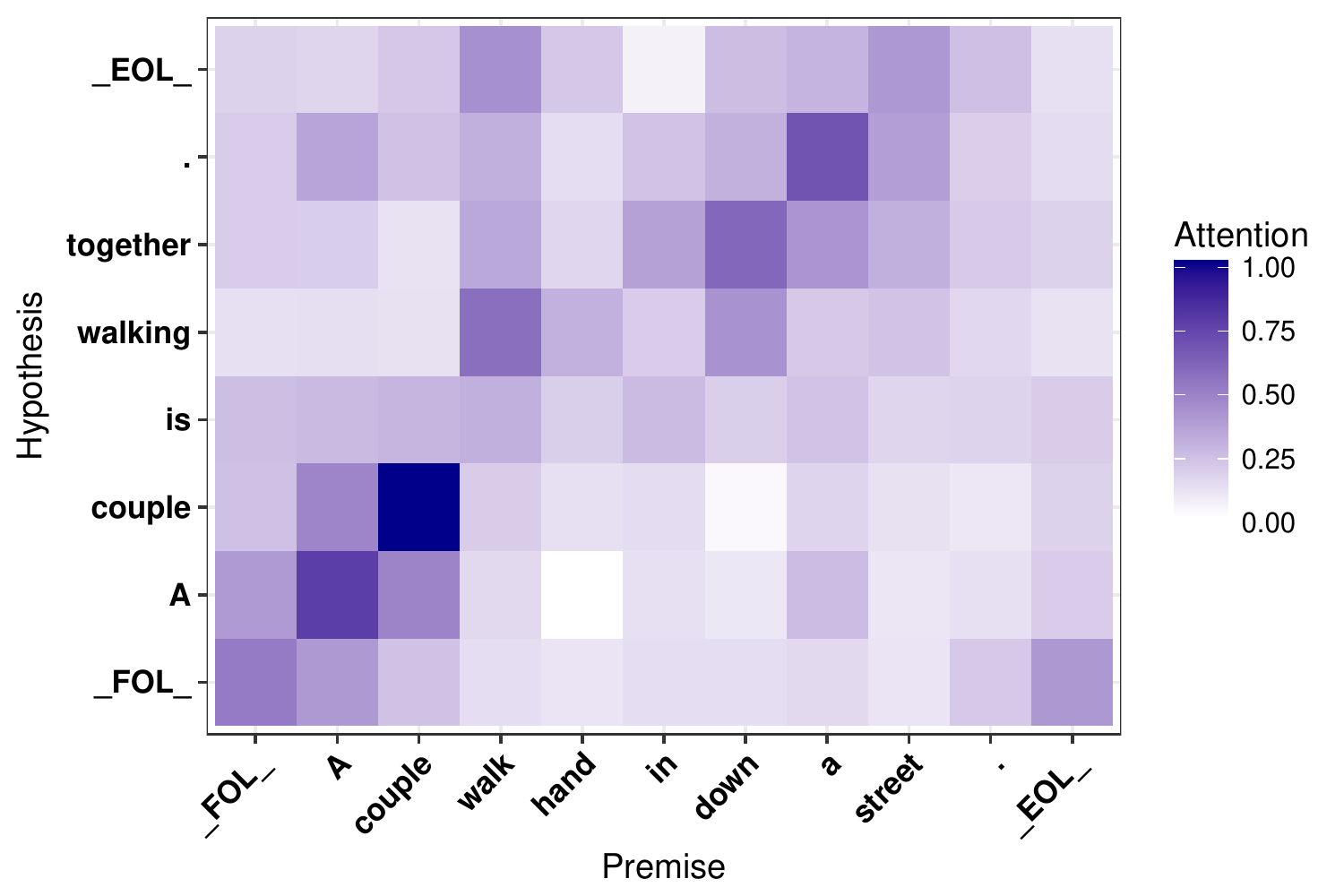}
			}%
			\subfigure[Instance 6 - Contradiction relationship.]{%
				\label{fig:sub:6:cnt}
				\includegraphics[width=0.49\textwidth]{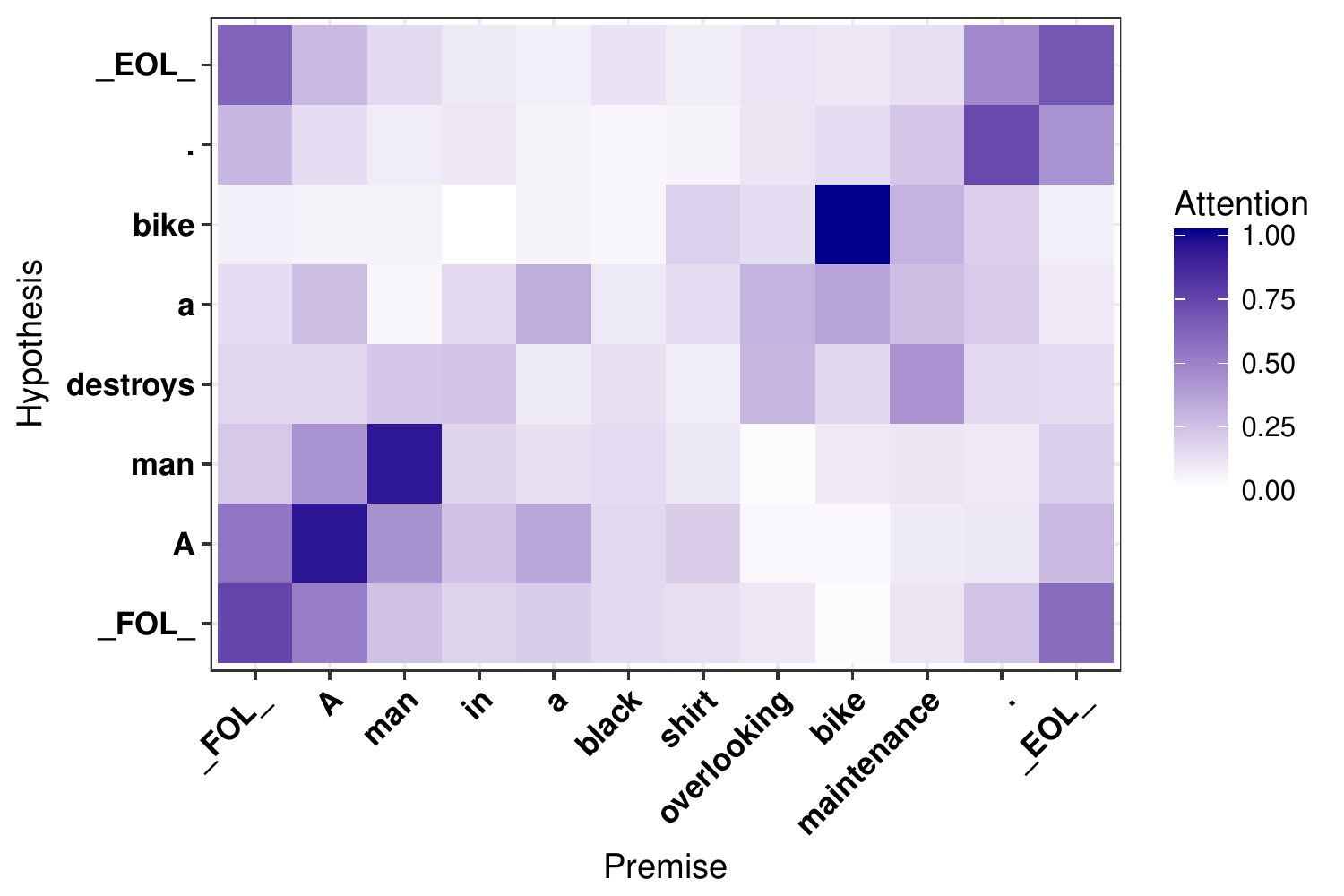}
			}%
		\end{center}
		\caption{
			Normalized attention weights for 6 data samples from the test set of SNLI dataset. (a,c,e) and (b,d,f) represent the normalized attention weights for \emph{Entailment}, \emph{Neutral}, and \emph{Contradiction} logical relationships of two premises (Instance 5 and 6) respectively. Darker color illustrates higher attention. 
		}
		\label{fig:att:sample:3}
	\end{figure*}

	In this section, we show visualizations of 18 samples of normalized attention weights (energy function, see Equation~\ref{eq:energy} in the paper). Each column in Figures~\ref{fig:att:sample:1}, \ref{fig:att:sample:2}, and \ref{fig:att:sample:3}, represents three data samples that share the same premise but differ in hypothesis. Also, each row is allocated to a specific logical relationship (Top: Entailment, Middle: Neutral, and Bottom: Contradiction). DR-BiLSTM classifies all data samples reported in these figures correctly.
	
\end{document}